\documentclass[]{elsarticle}

\usepackage[T1]{fontenc}
\usepackage{lmodern}

\usepackage{amsmath}
\usepackage{amsfonts,amsbsy,amscd}
\usepackage{mathtools}
\usepackage{amssymb}
\usepackage{mathrsfs}
\usepackage{array} 
\usepackage{mathdots}

\usepackage[exponent-product = \cdot]{siunitx}
\DeclareSIUnit\nounit{\relax}

\usepackage[dvipsnames]{xcolor}
\usepackage{pgfplots}
\pgfplotsset{compat=1.18}
\usepackage{graphics}
\usepackage{graphicx}
\usepackage{tikz}

\usetikzlibrary{
shapes.geometric, 
arrows.meta,                        
backgrounds,                        
ext.paths.ortho,                    
ext.positioning-plus,               
ext.node-families.shapes.geometric, 
arrows
}

\usepackage{hyperref}
\usepackage[nameinlink, capitalize]{cleveref}
\crefname{equation}{}{}
\crefname{appendix}{}{}

\usepackage[nolist]{acronym}

\usepackage[labelfont=bf, figurename=Fig.]{caption}
\usepackage{subcaption}
\usepackage{multirow}
\usepackage{booktabs}
\usepackage{enumitem}
\usepackage{tabularx}

\usepackage{color}
\usepackage{tcolorbox}

\usepackage{url}

\begin{acronym}
\acro{DIC}[DIC]{digital image correlation}
\acro{PDE}[PDE]{partial differential equation}
\acro{BC}[BC]{boundary condition}
\acro{MCMC}[MCMC]{Markov chain Monte Carlo}
\acro{HMC}[HMC]{Hamiltonian Monte Carlo}
\acro{NUTS}[NUTS]{No-U-Turn sampler}
\acro{NN}[NN]{neural network}
\acro{FFNN}[FFNN]{feed-forward neural network}
\acro{BNN}[BNN]{Bayesian neural network}
\acro{PINN}[PINN]{physics-informed neural network}
\acro{MAE}[MAE]{mean absolute error}
\acro{GPU}[GPU]{graphics processing unit}
\acro{UQ}[UQ]{uncertainty quantification}
\acro{NF}[NF]{normalizing flow}
\acro{GP}[GP]{Gaussian process}
\acro{PDF}[PDF]{probability density function}
\acro{VI}[VI]{variational inference}
\acro{BI}[BI]{Bayesian inference}
\acro{KLD}[KLD]{Kulback-Leibler divergence}
\acro{ELBO}[ELBO]{evidence lower bound}
\acro{SEF}[SEF]{strain energy density function}
\acro{CANN}[CANN]{constitutive artificial neural network}
\acro{EUCLID}[EUCLID]{efficient unsupervised constitutive law identification and discovery}
\acro{UT}[UT]{uniaxial tension}
\acro{EBT}[EBT]{equibiaxial tension}
\acro{BT}[BT]{biaxial tension}
\acro{SS}[SS]{simple shear}
\acro{PS}[PS]{pure shear}
\acro{MAF}[MAF]{masked autoregressive flow}
\acro{IAF}[IAF]{inverse autoregressive flow}
\acro{realNVP}[real NVP]{real-valued non-volume preserving}
\acro{MADE}[MADE]{masked autoencoder for distribution estimation}
\acro{RMSE}[RMSE]{root mean squared error}
\acro{R2}[R\textsuperscript{2}]{coefficient of determination}
\acro{MR}[MR]{Mooney-Rivlin}
\acro{LLM}[LLM]{large language model}
\acro{CKAN}[CKAN]{constitutive Kolmogorov-Arnold network}
\end{acronym}

\newcommand{\bm}[1]{\boldsymbol{#1}}

\newcommand{\cauchyTensor}{\ten{C}}

\newcommand{\indStructuralTensors}{i}
\newcommand{\structuralTensorGeneral}{\ten{M}^{(\indStructuralTensors)}}
\newcommand{\numStructuralTensors}{n_{\mathrm{M}}}
\newcommand{\structuralTensorFiber}{\ten{M}_{\mathrm{f}}}
\newcommand{\structuralTensorSheet}{\ten{M}_{\mathrm{s}}}
\newcommand{\structuralTensorNormal}{\ten{M}_{\mathrm{n}}}
\newcommand{\normalDirectionVectorFiber}{\vec{f}_{0}}
\newcommand{\normalDirectionVectorSheet}{\vec{s}_{0}}
\newcommand{\normalDirectionVectorNormal}{\vec{n}_{0}}
\newcommand{\principalStretch}[1]{\lambda_{#1}}
\newcommand{\indInvariants}{i}
\newcommand{\invariantOne}{\sca{I}_{1}}
\newcommand{\invariantTwo}{\sca{I}_{2}}
\newcommand{\invariantFourF}{\sca{I}_{\mathrm{4f}}}
\newcommand{\invariantFourS}{\sca{I}_{\mathrm{4s}}}
\newcommand{\invariantFourN}{\sca{I}_{\mathrm{4n}}}
\newcommand{\invariantFourFBar}{\barsca{I}_{\mathrm{4f}}}
\newcommand{\invariantFourSBar}{\barsca{I}_{\mathrm{4s}}}
\newcommand{\invariantFourNBar}{\barsca{I}_{\mathrm{4n}}}
\newcommand{\invariantEightFS}{\sca{I}_{\mathrm{8fs}}}
\newcommand{\invariantEightFN}{\sca{I}_{\mathrm{8fn}}}
\newcommand{\invariantEightSN}{\sca{I}_{\mathrm{8sn}}}

\newcommand{\rSquare}{\mathrm{R}^{2}}
\newcommand{\rmse}{\mathrm{RMSE}}
\newcommand{\estimatedCoverage}{\mathrm{EC}_{95\%}}
\newcommand{\coverage}{\mathrm{C}_{95\%}}

\newcommand{\materialParams}{\bm{\kappa}}
\newcommand{\materialParamsRealization}{\hat{\bm{\kappa}}}
\newcommand{\materialParamLinear}{\sca{c}}
\newcommand{\materialParamNonlinear}{\vec{w}}
\newcommand{\materialParamNonlinearSingle}{\sca{w}}
\newcommand{\indMaterialParam}{i}
\newcommand{\materialParamOne}{\sca{\kappa}_{\indMaterialParam}}
\newcommand{\numMaterialParams}{n_{\kappa}}
\newcommand{\reducedMaterialParams}{\bm{\tilde{\kappa}}}
\newcommand{\numReducedMaterialParams}{\tilde{n}_{\kappa}}
\newcommand{\gpParams}{\bm{\zeta}}
\newcommand{\gpParamsObservedPDataset}{\bm{\zeta}_{\setObservedPDataset_{\indObservedP}}}
\newcommand{\gpParamsOptimal}{\bm{\zeta}^{*}}
\newcommand{\gpParamsOptimalObservedPSpecific}[2]{\bm{\zeta}_{\setObservedPTestSpecific{#1}_{\mathrm{#2}}}^{*}}
\newcommand{\gpParamsOutputScale}{\sigma}
\newcommand{\gpParamsOutputScaleSquared}{\sigma^{2}}
\newcommand{\gpParamsLengthScales}{\vec{l}}
\newcommand{\gpParamsLengthScalesComponent}{\sca{l}}
\newcommand{\nfParams}{\bm{\Phi}}
\newcommand{\nfParamsOptimal}{\bm{\Phi}^{*}}
\newcommand{\nfParamsConstant}{\bm{\Phi}_{\mathrm{c}}}

\newcommand{\lipschitzParams}{\bm{\theta}}
\newcommand{\lipschitzParamsOptimal}{\bm{\theta}^{*}}
\newcommand{\lipschitzParamsConstant}{\bm{\theta}_{\mathrm{c}}}

\newcommand{\modelTerm}{\phi}
\newcommand{\numModelTerms}{n_{\mathrm{\phi}}}
\newcommand{\numReducedModelTerms}{\tilde{n}_{\mathrm{\phi}}}
\newcommand{\indModelTerms}{j}
\newcommand{\modelOutput}{\sca{P}}
\newcommand{\modelOutputGeneral}{\sca{P}^{(\indTests, \indObservedP, \indFuncPoints)}}
\newcommand{\modelTermOgden}{\modelTerm_{\mathrm{O}}}
\newcommand{\numModelTermsOgden}{\numModelTerms^{\mathrm{O}}}
\newcommand{\indModelTermsOgden}{l}
\newcommand{\exponentOgden}{\alpha}
\newcommand{\exponentOgdenGeneral}{\exponentOgden^{(\indModelTermsOgden)}}
\newcommand{\modelTermMR}{\modelTerm_{\mathrm{MR}}}
\newcommand{\numModelTermsMR}{\numModelTerms^{\mathrm{MR}}}
\newcommand{\indModelTermsMROne}{m}
\newcommand{\indModelTermsMRTwo}{k}

\newcommand{\dataSet}{\mathcal{D}}
\newcommand{\numData}{n_{\mathrm{d}}}
\newcommand{\numDataTesti}{\numData^{(\indTests)}}
\newcommand{\indData}{d}
\newcommand{\numTests}{n_{\mathrm{t}}}
\newcommand{\indTests}{t}
\newcommand{\setObservedP}{\mathcal{P}}
\newcommand{\setObservedPTesti}{\setObservedP^{(\indTests)}}
\newcommand{\setObservedPTestSpecific}[1]{\setObservedP^{(\mathrm{#1})}}
\newcommand{\setObservedPDataset}{\setObservedP}
\newcommand{\numObservedP}{n_{\mathrm{q}}}
\newcommand{\numObservedPTesti}{\numObservedP^{(\indTests)}}
\newcommand{\numObservedPTestSpecific}[1]{\numObservedP^{(\mathrm{#1})}}
\newcommand{\indObservedP}{q}
\newcommand{\observedPGeneral}{\vec{P}}
\newcommand{\observedPGeneralUnseen}{\barvec{P}}
\newcommand{\observedPGeneralTrueSquared}{\vec{P}^{*^{2}}}
\newcommand{\observedPGeneralTransposed}{\vec{P}\transpose}
\newcommand{\observedPGeneralTrue}{\vec{P}^{*}}

\newcommand{\observationMapGeneral}{\operatorname{O}^{(\indTests,\indObservedP)}}
\newcommand{\observationMapSpecific}[2]{\operatorname{O}^{(\mathrm{#1},\mathrm{#2})}}

\newcommand{\deformationFilterGeneral}{\operatorname{F}_{\setObservedPTesti_{\indObservedP}}}
\newcommand{\deformationFilterSpecific}[1]{\operatorname{F}_{\mathrm{#1}}}
\newcommand{\reducedFGeneral}{\vec{\Lambda}^{(\indTests,\indObservedP)}}
\newcommand{\reducedF}{\vec{\Lambda}}
\newcommand{\reducedFComponent}{\sca{\Lambda}}
\newcommand{\reducedFComponentDifferent}{\sca{\Lambda}'}
\newcommand{\reducedFUnseen}{\barvec{\Lambda}}
\newcommand{\reducedFDifferent}{\vec{\Lambda}'}
\newcommand{\reducedFAll}{\ten{\Lambda}}
\newcommand{\numReducedF}{n_{\Lambda}}
\newcommand{\numReducedFObservedP}{n_{\Lambda}^{(\indTests,\indObservedP)}}

\newcommand{\gpMeanFunction}{\sca{m}}

\newcommand{\gpMeanVectorPosterior}{\barvec{m}}
\newcommand{\gpCovarianceFunction}{\sca{k}}
\newcommand{\gpCovarianceMatrix}{\ten{K}}
\newcommand{\gpCovarianceMatrixNoise}{\tildeten{K}}
\newcommand{\gpCovarianceMatrixPosterior}{\barten{K}}
\newcommand{\gpCovarianceMatrixElement}{\ten{K}_{i,j}}
\newcommand{\gp}{\mathcal{N}}
\newcommand{\gpGeneral}{\gp_{\setObservedPTesti_{\indObservedP}}}
\newcommand{\gpSpecific}[2]{\gp_{\setObservedPTestSpecific{#1}_{\mathrm{#2}}}}
\newcommand{\sampleSpaceFuncGP}{\Omega_{\funcGP}}
\newcommand{\sampleSpaceObservedPGeneral}{\Omega_{\observedPGeneral}}

\newcommand{\noiseTerm}{\bm{\varepsilon}}
\newcommand{\samplesVariabilityTerm}{\bm{\eta}}
\newcommand{\errorTerm}{\bm{\epsilon}}
\newcommand{\errorCovariance}{\bm{\Sigma}_{\epsilon}}
\newcommand{\relativeErrorStddev}{\sigma_{\mathrm{r}}}
\newcommand{\minErrorStddev}{\sigma_{\mathrm{min}}}

\newcommand{\func}{\vec{f}}
\newcommand{\funcValue}{\sca{f}}
\newcommand{\numFuncs}{n_{\mathrm{f}}}
\newcommand{\numFuncPoints}{n_{\mathrm{s}}}
\newcommand{\indFuncPoints}{s}
\newcommand{\numFuncPointsTesti}{\numFuncPoints^{(\indTests)}}
\newcommand{\numFuncPointsTestSpecific}[1]{\numFuncPoints^{(#1)}}
\newcommand{\funcPointsDeformationGradients}{\ten{F}_{\mathrm{f}}}
\newcommand{\funcPointsDeformationGradientsTesti}{\funcPointsDeformationGradients^{(\indTests)}}
\newcommand{\funcGP}{\vec{f}_{\mathrm{GP}}}
\newcommand{\funcDistGP}{p_{\mathrm{GP}}(\vec{f})}
\newcommand{\funcDistGPMeasure}{p_{\mathrm{GP}}}
\newcommand{\funcModel}{\vec{f}_{\mathrm{M}}}
\newcommand{\funcDistModel}{p_{\mathrm{M}}(\vec{f})}
\newcommand{\funcDistModelMeasure}{p_{\mathrm{M}}}
\newcommand{\funcDistModelNF}{p_{\mathrm{M}}(\vec{f}; \nfParams)}
\newcommand{\lossWasserstein}{\mathcal{L}_{\mathrm{W}}}
\newcommand{\numItersWasserstein}{n_{\mathrm{iters}}^{\mathrm{W}}}

\newcommand{\wassersteinLipschitzNetwork}{\sca{f}_{\mathrm{LN}}}
\newcommand{\lossLipschitz}{\mathcal{L}_{\mathrm{L}}}
\newcommand{\numItersLipschitz}{n_{\mathrm{iters}}^{\mathrm{L}}}
\newcommand{\lipschitzPenaltyCoefficient}{\lambda_{\mathrm{L}}}
\newcommand{\lipschitzCombinedFunc}{\hatvec{f}}
\newcommand{\lipschitzCombinedFuncDist}{p_{\mathrm{\hatsca{f}}}}

\newcommand{\materialParamsDist}{p_{\kappa}(\materialParams)}
\newcommand{\materialParamsDistOptimal}{p^{*}_{\kappa}(\materialParams)}
\newcommand{\reducedMaterialParamsDist}{p_{\tilde{\kappa}}(\reducedMaterialParams)}
\newcommand{\materialParamsDistNF}{p_{\kappa}(\materialParams; \nfParams)}
\newcommand{\reducedMaterialParamsDistNF}{p_{\tilde{\kappa}}(\reducedMaterialParams; \nfParams)}
\newcommand{\reducedMaterialParamsDistNFOptimized}{p_{\tilde{\kappa}}(\reducedMaterialParams; \nfParamsOptimal)}

\newcommand{\parametersToStateMap}{\vec{T}_{\mathrm{M}}}
\newcommand{\parametersToStateMapPush}{\vec{T}_{\mathrm{M}_{\#}}}
\newcommand{\materialParamsDistPush}{p_{\kappa}(\func)}

\newcommand{\nfTransform}{\vec{T}}
\newcommand{\nfNumTransforms}{n_{\mathrm{T}}}
\newcommand{\nfIndTransforms}{k}
\newcommand{\nfIntermediateSamples}{\vec{z}}
\newcommand{\nfBaseSamples}{\vec{u}}
\newcommand{\nfBaseSamplesDist}{p_{\mathrm{u}}(\nfBaseSamples)}
\newcommand{\nfIAFLocation}{\vec{l}}
\newcommand{\nfIAFScaleUnconstrained}{\vec{s}}
\newcommand{\nfIAFScaleConstrained}{\vec{s}_{\mathrm{c}}}
\newcommand{\nfMADENumLayers}{L}
\newcommand{\nfMADEIndLayer}{l}
\newcommand{\nfMADEHiddenState}{\vec{h}}
\newcommand{\nfMADEHiddenStateSize}{n_{\mathrm{h}}^{(\nfMADEIndLayer)}}
\newcommand{\nfMADEActivation}{\vec{a}}
\newcommand{\nfMADEWeights}{\ten{W}}
\newcommand{\nfMADEBiases}{\vec{b}}
\newcommand{\nfMADEMask}{\ten{M}}
\newcommand{\nfMADEMaskFunc}{\sca{m}}

\newcommand{\totalSobolIndex}{\sca{S}_{\mathrm{T}}}
\newcommand{\totalSobolIndexGeneral}{\totalSobolIndex^{(\indTests, \indObservedP, \indFuncPoints)}}
\newcommand{\totalSobolIndexAveraged}{\barsca{S}_{\mathrm{T}}}

\newcommand{\uqInterval}{\mathcal{C}_{\func}}
\newcommand{\uqIntervalGeneral}{\uqInterval^{(\indTests, \indObservedP, \indFuncPoints)}}
\newcommand{\uqIntervalLowGeneral}{\sca{L}_{\func}^{(\indTests, \indObservedP, \indFuncPoints)}}
\newcommand{\uqIntervalUpGeneral}{\sca{U}_{\func}^{(\indTests, \indObservedP, \indFuncPoints)}}
\newcommand{\funcValueGeneral}{\funcValue^{(\indTests, \indObservedP, \indFuncPoints)}}
\newcommand{\funcValueGeneralData}{\funcValue^{(\indTests, \indObservedP, \indData)}}

\DeclareMathOperator*{\argmin}{arg\,min}
\DeclareMathOperator*{\argmax}{arg\,max}

\newcommand{\supop}[1]{\underset{\operatorname{#1}}{\operatorname{sup}}}

\newcommand{\norm}[1]{\left\lVert#1\right\rVert}  

\newcommand{\expectation}[1]{\mathbb{E}_{#1}}
\newcommand{\variance}[1]{\mathbb{V}_{#1}}

\newcommand{\sca}[1]{ \ensuremath{#1} }
\newcommand{\barsca}[1]{ \ensuremath{ \bar{ \sca{#1} } } }
\newcommand{\widebarsca}[1]{ \ensuremath{ \overline{ \sca{#1} } } }
\newcommand{\hatsca}[1]{ \ensuremath{ \hat{ \sca{#1} } } }

\renewcommand{\vec}[1]{ \ensuremath{ \mathbf{#1} } }
\newcommand{\barvec}[1]{\ensuremath{ \bar{ \vec{#1} } } }
\newcommand{\hatvec}[1]{\ensuremath{ \hat{ \vec{#1} } } }

\newcommand{\ten}[1]{\ensuremath{ \boldsymbol{ \mathsf{#1} } } }
\newcommand{\barten}[1]{\ensuremath{ \bar{ \ten{#1} } } }

\newcommand{\tildeten}[1]{\ensuremath{ \tilde{ \ten{#1} } } }

\newcommand{\elm}[1]{{\, \in \mathbb{R}}^{#1}}
\newcommand{\elmm}[2]{{\, \in \mathbb{R}}^{#1 \times #2}}

\newcommand{\half}{\frac{1}{2}}

\newcommand{\transpose}{^{\top}}
\newcommand{\minusTranspose}{^{-\top}}
\newcommand{\tr}[1]{\operatorname{tr}\left(#1\right)}

\begin{document}
\begin{frontmatter}
\title{Uncertainty quantification in model discovery by distilling interpretable material constitutive models from Gaussian process posteriors}

\author[1]{David Anton\corref{cor1}} \ead{d.anton@tu-braunschweig.de}
\author[1]{Henning Wessels} \ead{h.wessels@tu-braunschweig.de}
\author[2]{Ulrich Römer} \ead{u.roemer@tu-braunschweig.de}
\author[3]{Alexander Henkes} \ead{ahenke@ethz.ch}
\author[1]{Jorge-Humberto Urrea-Quintero} \ead{jorge.urrea-quintero@tu-braunschweig.de}
\cortext[cor1]{Corresponding author}
    
\affiliation[1]{organization={Technische Universität Braunschweig, Institute of Applied Mechanics}, addressline={Pockelsstraße 3}, postcode={38106}, city={Braunschweig}, country={Germany}}
\affiliation[2]{organization={Technische Universität Braunschweig, Institute for Acoustics and Dynamics}, addressline={Langer Kamp 19}, postcode={38106}, city={Braunschweig}, country={Germany}}
\affiliation[3]{organization={Eidgenössische Technische Hochschule Zürich, Computational Mechanics Group}, addressline={Tannenstrasse 3}, postcode={8092}, city={Zürich}, country={Switzerland}}

\begin{abstract}

\noindent Constitutive model discovery refers to the task of identifying an appropriate model structure, usually from a predefined model library, while simultaneously inferring its material parameters. The data used for model discovery are measured in mechanical tests and are thus inevitably affected by noise which, in turn, induces uncertainties. Previously proposed methods for uncertainty quantification in model discovery either require the selection of a prior for the material parameters, are restricted to linear coefficients of the model library or are limited in the flexibility of the inferred parameter probability distribution. We therefore propose a partially Bayesian framework for uncertainty quantification in model discovery that does not require prior selection for the material parameters and also allows for the discovery of constitutive models with inner-non-linear parameters: First, we augment the available stress-deformation data with a Gaussian process. Second, we approximate the parameter distribution by a normalizing flow, which allows for modeling complex joint distributions. Third, we distill the parameter distribution by matching the distribution of stress-deformation functions induced by the parameters with the Gaussian process posterior. Fourth, we perform a Sobol' sensitivity analysis to obtain a sparse and interpretable model. We demonstrate the capability of our framework for both isotropic and experimental anisotropic data.

\end{abstract}


\end{frontmatter}


\section{Introduction}\label{sec:introduction}

In order to unlock the predictive capabilities of continuum mechanics, it is essential to discover constitutive models for the material of interest that relate stress to strain and possibly other physical quantities. The conventional approach to material modeling involves a two-step process. First, a constitutive model with fixed structure is formulated, typically expressed in terms of invariant or tensor-basis representations, and constrained to satisfy thermodynamic requirements \cite{ogden_nonLinearElasticDeformations_1984,holzapfel_nonlinearSolidMechanics_2000}. Second, the degrees of freedom of the constitutive model, also known as material parameters, are calibrated using measurement data \cite{steinmann_hyperelasticModelsTreloar_2012,ricker_systematicFittingHyperelasticity_2023}. For an overview of calibration methods, the reader is referred to, e.g., \cite{roemer_modelCalibrationInSolidMechanics_2025}. However, the predictive capability of a calibrated model is decisively dependent on the suitability of the chosen constitutive model \cite{holzapfel_nonlinearSolidMechanics_2000,ricker_systematicFittingHyperelasticity_2023,gurtin_mechanicsThermodynamicsOfContinua_2010}.\\

\noindent \textbf{Constitutive model discovery:} 
The idea of constitutive model discovery is to find a suitable structure for the constitutive model and to infer the material parameters at the same time \cite{flaschel_unsupervisedDiscoveryEUCLID_2021}. In this process, the constitutive model is usually selected as a reduced subset of a previously defined model library of candidate terms such as, e.g., a combination of generalized Mooney-Rivlin \cite{rivlin_MooneyRivlinModel_1947} and generalized Ogden \cite{ogden_OgdenModel_1972} models. The \ac{EUCLID} framework pioneered this approach in the discovery of interpretable hyperelastic constitutive models in an unsupervised setting \cite{flaschel_unsupervisedDiscoveryEUCLID_2021,thakolkaran_NNEUCLID_2022,flaschel_brainEUCLID_2023}. \Ac{EUCLID} has also been extended to inelastic materials \cite{flaschel_discoveringPlasticityModels_2022,flaschel_generalizedStandardMaterialModelsEUCLID_2023,marino_linearViscoelasticityEUCLID_2023}. Further attempts utilize artificial \acp{NN} as constitutive model. Although purely data-driven approaches are generally flexible, they lack interpretability and can show non-physical and unstable material behavior \cite{fuhg_reviewDataDrivenConstitutiveLaws_2025}. To prevent non-physical behavior, recent developments incorporate physical constraints, such as thermodynamic consistency, polyconvexity, objectivity and material symmetry, directly into the \ac{NN} architecture \cite{linka_CANNs_2021,klein_polyconvexAnisotropicHyperelasticityWithNN_2022,linden_NNHyperelasticityEnforcingPhysics_2023,fuhg_hyperelasticAnisotropyTensorBasedNN_2022,kalina_NNMeetsAnisotropicHyperelasticity_2025,tepole_polyconvexPANNs_2025}. 

In addition to satisfying fundamental physical constraints, \acp{CANN} have been proposed to enable interpretability of the \ac{NN}-based constitutive model by assigning the weights of the \ac{NN} a physical meaning \cite{linka_newCANNs_2023,linka_brainCANNs_2023,linka_skinCANNs_2023,peirlinck_universalMaterialSubroutineHyperelasticity_2024,martonova_modelDiscoveryCardiacTissue_2025}. \Acp{CANN} can also be interpreted as model libraries including linear and non-linear candidate terms. Recently, \acp{CANN} have also been combined with \acp{LLM} \cite{tacke_constitutiveScientificGenerativeAgent_2025}. There are also hybrid approaches, such as \acp{CKAN} \cite{thakolkaran_inputConvexKANs_2025,abdolazizi_CKANs_2025}, that aim to combine the accuracy of purely data-driven methods with the interpretability of symbolic expressions.\\

\noindent \textbf{Uncertainty quantification in model discovery:} The data used to find a suitable model are measured in mechanical tests and thus are corrupted by noise, which directly introduces uncertainties. In addition, stress-deformation measurements, in particular, may be sparse. Data sparsity, in turn, also increases uncertainty \cite{linden_BayesianParameterEstimation_2022}. When applying deterministic methods to the experimental data, uncertainty in the discovered model terms or corresponding material parameters is not quantified. Instead, we obtain information reduced to deterministic parameter values, which may lead to a false sense of confidence.

In a Bayesian statistical setting, the material parameters are treated as random variables with a prior distribution. This prior distribution is then updated to the posterior distribution according to Bayes' law by conditioning it on the data \cite{gelman_bayesianDataAnalysis_2013}. A central element and, at the same time, a prerequisite for using Bayes' law is the formulation of the prior. In \cite{joshi_BayesianEUCLID_2022}, the Bayesian-\ac{EUCLID} framework was proposed to discover hyperelastic constitutive models with uncertainties. The authors used a hierarchical Bayesian model with sparsity-promoting priors and a \ac{MCMC} sampling strategy. Similarly to \ac{EUCLID}, there is also a Bayesian statistical variant for \acp{CANN} \cite{linka_BayesianCANNs_2025}, known as Bayesian \acp{CANN}. Instead of optimizing for a deterministic value of the network weights, which correspond to the material parameters, the authors used variational Bayesian inference to learn the probability density of the weights in the output layer. However, in this approach, the parameters considered uncertain lack physical interpretability and are assumed to be independently distributed.

Bayesian approaches to \ac{UQ} for model discovery are complicated by the need to formulate informative priors. Both the large number of material parameters and the fact that the relevance of parameters is unknown before the model discovery process pose significant challenges. In a related context, \ac{UQ} for learning dynamical systems faces similar difficulties and priors are mainly used to enforce sparsity, see, e.g., \cite{hirsh_sparsifyingPriorsModelDiscovery_2022}. Similarly, for \acp{BNN}, the large number of parameters makes it practically impossible to formulate a well-informed prior for individual parameters \cite{melev_noFreeBayesianUQ_2025}.

Compared to Bayesian \acp{CANN}, Gaussian \acp{CANN} \cite{mcculloch_GaussianCANNsCorrelatedWeights_2025} are more interpretable and do not require the selection of a prior for the random material parameters. In addition, Gaussian \acp{CANN} allow the weights to be correlated. However, the random weights that again correspond to the material parameters are restricted to be Gaussian distributed. Furthermore, only the linear material parameters of the model library are assumed to be random variables, while the non-linear parameters are considered deterministic.

To the best of the authors' knowledge, the above-mentioned contributions are the only ones in the literature to date that have proposed methods for the statistical discovery of interpretable material constitutive models. For the sake of completeness, we would like to point out that other statistical methods have also been developed in recent years. These include, among others, a \ac{GP}-based constitutive modeling framework \cite{aggarwal_strainEnergyDensityAsGP_2023} or generative models for hyperelastic \acp{SEF} based on physics-informed probabilistic diffusion fields \cite{tac_physicsInformedProbabilisticDiffusionFields_2025}. However, these methods do not yield sparse and interpretable constitutive models.\\

\noindent \textbf{Our framework:} Motivated by the limitations of the methods mentioned above, we propose a new four-step framework for \ac{UQ} in the discovery of interpretable constitutive models that is partially Bayesian:
\begin{itemize}
    \item First, we augment the available stress-deformation data collected in mechanical tests with a \ac{GP} as one of the authors has proposed in \cite{roemer_GPBasedCalibrationBiomechanicalModels_2022} in the context of surrogate modeling. 
    \item Second, we approximate the probability distribution of the material parameters in the model library by a \ac{NF} \cite{tabak_nonparametricDensityEstimationAlgorithms_2013,rezende_normalizingFlows_2015}, which allows modeling complex and high-dimensional joint distributions \cite{papamakarios_normalizingFlows_2021, wang_normalizingFlowAdaptiveSurrogate_2022}. The distribution of material parameters, in turn, induces a distribution over stress-deformation functions through the structure of the model library for the \ac{SEF}. 
    \item Third, we distill the distribution of the material parameters by matching the distribution over stress-deformation functions induced by them with the target distribution given by the \ac{GP} posterior. Therefore, we minimize the Wasserstein-1 distance between the two distributions with respect to the \ac{NF} parameters. 
    \item Fourth, we perform a Sobol' sensitivity analysis \cite{sobol_globalSensitivityIndices_2001}, which finally yields sparse and interpretable constitutive models.
\end{itemize}

\noindent The framework we propose contributes to the current state of \ac{UQ} in model discovery as follows:
\begin{itemize}
    \item The framework is only partially Bayesian and does not require the selection of an informative prior probability distribution for the material parameters. Instead, we only need a \ac{GP} prior which can be learned from the available stress-deformation data. This is particularly appealing because model discovery is motivated by the fact that the relevance of the material parameters and even the structure of the constitutive model are not known a priori.
    \item \Acp{NF} offer a high degree of flexibility for modeling the joint probability distribution of material parameters compared to simpler distributions, such as multivariate Gaussian distributions.
    \item The selection of the relevant model terms is based on their total-order Sobol' sensitivity. The sensitivity analysis provides us with an interpretable selection criterion and further insights into the selection process.
\end{itemize}

The framework is inspired by the work in \cite{tran_functionalPriorForBNN_2022} which deals with determining suitable priors for the parameters of \acp{BNN}. Their investigations start with the observation that the functional priors of \acp{BNN} are much easier to interpret and control than a prior defined directly for the \ac{NN} parameters. Therefore, they propose to match the functional prior of the \ac{BNN} with a target \ac{GP} prior by minimizing their Wasserstein-1 distance with respect to the distributional parameters of the prior ansatz for the \ac{NN} weights. Two key differences between our framework and the one proposed in \cite{tran_functionalPriorForBNN_2022} are that we use \ac{GP} posteriors as target distributions while they used priors and that we minimize the Wasserstein-1 distance to distill the distribution over parameters of a model library instead of the weights of a \ac{BNN} that lack interpretability.

Finally, we refer to the inference process as distillation, as we distill the distribution over the material parameters from the distribution over the stress-deformation functions encoded by the \ac{GP} posterior. The distilled distribution is easier to interpret, but preserves uncertainties. The concept of distilling knowledge from data has also been coined in the context of model discovery of physical laws in \cite{schmidt_distillingNaturalLaws_2009}. Furthermore, we note that our proposed framework can be associated with generative modeling. From the generative modeling point of view, the objective of training the \ac{NF} would also be to generate stress-deformation functions that follow the distribution given by the target \ac{GP} posterior. However, the motivation is different. In our case, the primary motivation is again to infer the distribution of material parameters and not the generation of new stress-deformation functions.\\

We demonstrate the capability of our approach for the isotropic Treloar dataset \cite{treloar_vulcanisedRubberData_1944} and an anisotropic dataset of human cardiac tissue \cite{sommer_humanVentricularMyocardium_2015}.
The research code for our numerical tests is implemented in the Python programming language and published on GitHub and Zenodo \cite{anton_codeUQInModelDiscovery_2025}. Our code is mainly based on \texttt{PyTorch} \cite{paszke_PyTorch_2019}. For the implementation of the \acp{GP}, we used \texttt{GPyTorch} \cite{gardner_gpytorch_2018}. In addition, we implemented the \ac{NF} and the Sobol' sensitivity analysis using the \texttt{normflows} \cite{stimper_normflows_2023} and \texttt{SALib} \cite{herman_SALib_2017, iwanaga_SALib2_2022} frameworks, respectively.


\section{Methodology}\label{sec:methodology}

In this section, we present a framework for the quantification of uncertainties in the discovery of material constitutive models. The method is based on the distillation of a joint distribution over the parameters of a sparse constitutive model from \ac{GP} posteriors of stress-deformation functions. First, we state the problem to be solved and present our solution approach. Second, we recapitulate the basics of hyperelastic constitutive modeling, define a general hyperelastic model library for \acp{SEF} that covers both isotropic and anisotropic materials, and introduce the notation used in this paper. Finally, we present the four-step framework for distilling a joint distribution of material parameters from the \ac{GP} posteriors and further elaborate on the individual steps in more detail.

\subsection{Problem statement and solution approach}\label{subsec:problem_and_solution}

For a given dataset $\dataSet$ containing stress-deformation measurements from mechanical tests, the objective is to identify an appropriate and interpretable constitutive model from a predefined constitutive model library and simultaneously infer the probability density of the material parameters. The interpretability of the discovered model is improved by reducing the initial vector of material parameters $\materialParams \elm{\numMaterialParams}$ of the model library to the relevant parameters $\reducedMaterialParams \elm{\numReducedMaterialParams}$, so that $\numReducedMaterialParams \ll \numMaterialParams$.

For the sake of brevity, the following motivating discussion is kept short. Details on the derivations can be found in~\ref{sec:details_variational}. The starting point for a Bayesian approach to model discovery is usually the joint \ac{PDF} $p(\materialParams, \dataSet)$ of the material parameters $\materialParams$ and the stress-deformation data $\dataSet$. Using Bayes' law yields the posterior density
\begin{equation}\label{eq:bayes_without_functions}
    p(\materialParams \mid \dataSet) 
    = \frac{p(\dataSet \mid \materialParams) \, p(\materialParams)}{p(\dataSet)},
\end{equation}
where $p(\dataSet \mid \materialParams)$ denotes the likelihood function, $p(\materialParams)$ the prior density, and $p(\dataSet)$ the evidence which can be considered constant for fixed data $\dataSet$ \cite{gelman_bayesianDataAnalysis_2013}. The sparsity of the constitutive model can generally be promoted, e.g., by using sparsifying-priors for $p(\materialParams)$ as done in \cite{joshi_BayesianEUCLID_2022}. However, the typical use case for model discovery assumes that there is a large number of material parameters, and a priori very little is known about the relevance of the individual material parameters and also the structure of the constitutive model. Therefore, it is challenging to formulate a suitable prior $p(\materialParams)$, even if the prior is mainly used for sparsification. When using non-informative, i.e., very vague or diffuse priors, the posterior density $p(\materialParams \mid \dataSet)$ in \cref{eq:bayes_without_functions} is almost entirely determined by the likelihood function $p(\dataSet \mid \materialParams)$. Consequently, the benefits of Bayesian regularization are lost, and the posterior reduces to frequentist estimates \cite{lemoine_beyondNoninformativePriors_2019}.

Instead of targeting the posterior \eqref{eq:bayes_without_functions} directly, we introduce an intermediate \ac{GP} with stress values $\func$. This latent \ac{GP} is conditioned on $\dataSet$ and subsequently serves as a data surrogate for inferring $\materialParams$. As shown in~\ref{sec:details_variational}, this two-step approach can be expressed as 
\begin{equation}\label{eq:bayes_marginal_posterior}
    p(\materialParams \mid \dataSet) 
    = \int p(\materialParams, \func \mid \dataSet) \, \mathrm{d}\func
    = \int p(\materialParams \mid \func) \, p(\func \mid \dataSet) \, \mathrm{d}\func.
\end{equation}
Since the integral in \cref{eq:bayes_marginal_posterior} is cumbersome to compute, we instead solve a variational problem to approximate $p(\materialParams \mid \dataSet)$ as follows
\begin{equation}\label{eq:bayes_variational_approximation}
    \materialParamsDistOptimal
    = \argmin_{\materialParamsDist} D 
        \bigl(
            \parametersToStateMapPush\materialParamsDistPush, p(\func \mid \dataSet)
        \bigr).
\end{equation}
Here, $\parametersToStateMap: \materialParams \mapsto \vec{g}$ represents the model library, taking parameter values $\materialParams$ to the discrete \ac{GP} space and $\parametersToStateMapPush\materialParamsDistPush$ is the \ac{PDF} of the push-forward measure of $\materialParamsDist$ implied by the deterministic model library. Furthermore, $D$ denotes a distance between \acp{PDF}. In~\ref{sec:details_variational}, we show that for a perfect model library, covering the entire set of \ac{GP} paths, $\parametersToStateMap$ being bijective, a suitable choice of $D$ and a sufficiently rich variational set for \ac{PDF} $\materialParamsDist$, $\materialParamsDistOptimal = p(\materialParams \mid \dataSet)$. Hence, the variational approach targets the posterior \eqref{eq:bayes_marginal_posterior}. However, we will apply \eqref{eq:bayes_variational_approximation} in much more general and realistic settings as a variational substitute for the full Bayesian approach, effectively removing the need to formulate a prior for the material parameters. Note that for an imperfect model library or a lack of expressiveness of $\materialParamsDist$, the distance in \cref{eq:bayes_variational_approximation} does not fully vanish, and there is a mismatch between the densities $\materialParamsDistOptimal$ and $p(\materialParams \mid \dataSet)$. In the variational approach, the sparsity of $\reducedMaterialParamsDist$ can then be achieved, e.g., in a downstream step in which a sensitivity analysis is performed and non-sensitive parameters are removed. In \cref{subsec:model_distillation}, we propose a framework for solving the variational problem \cref{eq:bayes_variational_approximation} and reducing the inferred density $\materialParamsDist$ to the relevant material parameters.\\

\subsection{Hyperelastic constitutive modeling}\label{subsec:hyperelastic_constitutive_modeling}

In the framework of continuum solid mechanics, the first Piola-Kirchhoff stress tensor $\ten{P}$ is derived from a scalar-valued \ac{SEF} $\sca{W}$ as follows
\begin{equation}\label{eq:first_PK_from_SEF_compressible}
    \ten{P} = \frac{\partial \sca{W}(\ten{F}; \materialParams)}{\partial \ten{F}} \quad \text{(compressible case)}.
\end{equation}
Here, $\ten{F} = \operatorname{Grad}\vec{x}$ denotes the deformation gradient and $\vec{x} \elm{3}$ corresponds to the position of a material point in the current configuration. Furthermore, $\materialParams \elm{\numMaterialParams}$ is a vector of material parameters with $\numMaterialParams$ components. In the special case of incompressibility, the constraint $\operatorname{det}\ten{F} = 1$ is enforced via a Lagrange multiplier which can be identified as the hydrostatic pressure $\sca{p}$.  Accordingly, \cref{eq:first_PK_from_SEF_compressible} modifies to
\begin{equation}\label{eq:first_PK_from_SEF_incompressible}
    \ten{P} = \frac{\partial \widebarsca{W}(\ten{F}; \materialParams)}{\partial \ten{F}} - \sca{p} \ten{F}\minusTranspose \quad \text{(incompressible case)}.
\end{equation}
Note that in this special case, $\widebarsca{W}$ denotes only the isochoric part of the \ac{SEF}. The hydrostatic pressure $\sca{p}$ in \cref{eq:first_PK_from_SEF_incompressible} is usually determined from global equilibrium in combination with loading and boundary conditions. The identification of a suitable \ac{SEF} $\widebarsca{W}(\ten{F}; \materialParams)$ in \cref{eq:first_PK_from_SEF_incompressible} from stress-deformation data constitutes a supervised regression problem that requires the formulation of a suitable ansatz for $\widebarsca{W}$. In the context of continuum mechanics, it is of utmost importance that such an ansatz is compatible with fundamental constraints, such as thermodynamic consistency, objectivity, material and Cauchy stress symmetry, non-negativity and polyconvexity. For more details, the reader is referred to standard textbooks, e.g., \cite{holzapfel_nonlinearSolidMechanics_2000}.\\

In the present work, we consider isochoric hyperelastic \acp{SEF} $\widebarsca{W}$ that fulfill the aforementioned constraints by construction and can be written as a linear combination of the model terms $\{ \modelTerm^{(\indModelTerms)} \}_{\indModelTerms=1}^{\numModelTerms}$ as follows
\begin{equation}\label{eq:ansatz_sef_general}
  \widebarsca{W}\bigl( \ten{F}, \{\structuralTensorGeneral\}_{\indStructuralTensors=1}^{\numStructuralTensors}; \materialParams\bigr) =
  \sum_{\indModelTerms=1}^{\numModelTerms} \materialParamLinear^{(\indModelTerms)} \, \modelTerm^{(\indModelTerms)} \bigl(\ten{F}, \{\structuralTensorGeneral\}_{\indStructuralTensors=1}^{\numStructuralTensors}; \materialParamNonlinear^{(\indModelTerms)} \bigr).
\end{equation}
Here, $\materialParamLinear^{(\indModelTerms)}$ and $\materialParamNonlinear^{(\indModelTerms)}$ denote the outer-linear and inner-non-linear material parameters, respectively, composed as $\materialParams = \bigl[\materialParamLinear^{(1)}, \dots, \materialParamLinear^{(\numModelTerms)}, \materialParamNonlinear^{(1)\transpose}, \dots, \materialParamNonlinear^{(\numModelTerms)\transpose}\bigr]\transpose$. In addition, $\{\structuralTensorGeneral\}_{\indStructuralTensors=1}^{\numStructuralTensors}$ denotes a set of structural tensors that describe the class of material symmetry. A reduced representation of the input arguments $\ten{F}$ and $\{\structuralTensorGeneral\}_{\indStructuralTensors=1}^{\numStructuralTensors}$ in \cref{eq:ansatz_sef_general} is obtained with their invariants $I_{\indInvariants}$. The explicit formulations of the constitutive models used will be presented together with the numerical test cases in \cref{sec:results}.

The general form of the model library \cref{eq:ansatz_sef_general} expresses the isochoric hyperelastic \ac{SEF} as a sum over possibly non-linear model terms and includes many classical and modern approaches. For example, in the context of \ac{EUCLID} \cite{flaschel_unsupervisedDiscoveryEUCLID_2021}, constitutive models of type \cref{eq:ansatz_sef_general} were considered, but with linear coefficients only. Even \acp{CANN}, originally introduced in \cite{linka_newCANNs_2023}, can be formulated in the general form given in \cref{eq:ansatz_sef_general}, resulting in model libraries with non-linear parameters, see \cite{urrea_automatedModelDiscovery_2025}.

\subsection{Dataset and notation}\label{subsec:dataset_and_notation}
We consider a dataset 
$\dataSet = 
\bigl\{
    \bigl\{
        \ten{F}^{(\indTests, \indData)}, 
        \{
            \sca{P}^{(\indTests, \indObservedP, \indData)}
        \}_{\indObservedP=1}^{\numObservedPTesti}
    \bigr\}_{\indData=1}^{\numDataTesti} 
\bigr\}_{\indTests=1}^{\numTests}$ 
composed of pairwise stress-deformation measurements from $\numTests$ mechanical tests. In the test $\indTests$, $\numDataTesti$ deformation gradients and stress tensors are measured. Typically, in each mechanical test $\indTests$, only a subset of $\numObservedPTesti \leq 9$ components of the stress tensor $\ten{P}$ are observed. Throughout this paper, deformation related quantities are mainly indexed by a double-index, such as the deformation gradients $\ten{F}^{(\indTests, \indData)}$ in $\dataSet$. Similarly, the stress components are indexed by a triple-index of the form $\sca{P}^{(\indTests, \indObservedP, \indData)}$. Here, $\indTests$, $\indObservedP$, $\indData$ are the indices for the mechanical test, the observed stress component, and the measurement, respectively. For simplicity, the integer value of the index $\indTests$ is sometimes replaced by its respective abbreviation, such as \acs{UT} for a uniaxial tension test. Moreover, if there is no dependence on the measurement point, the respective index $\indData$ will be omitted.

We introduce the observation map $\observationMapGeneral: \mathbb{R}^{3 \times 3} \to \mathbb{R}$ that filters out the stress component $\indObservedP \in \{1, \dots, \numObservedPTesti\}$ observed in the respective mechanical test $\indTests$. For the \ac{UT} test in the Treloar dataset \cite{treloar_vulcanisedRubberData_1944}, e.g., the observation map is defined as $\observationMapSpecific{UT}{1}(\ten{P}) = \sca{P}_{11}$. The specific definitions of the observation maps for all the numerical test cases we consider in this paper are provided in \cref{subsec:appendix_observation_maps}. 

Furthermore, $\setObservedPTesti$ is the set of all observed stress components for the mechanical test $\indTests$ with the cardinality of the set $|\setObservedPTesti| = \numObservedPTesti$. The set of observed stress components is indexed by the observed stress index $\indObservedP$. For example, for the \ac{UT} test in the Treloar dataset, $\setObservedPTestSpecific{UT} = \{\sca{P}_{11}\}$ and $\setObservedPTestSpecific{UT}_{\indObservedP=1} = \sca{P}_{11}$. The set of all observed stress components for the dataset $\dataSet$ is the union of all $\setObservedPTesti$ for all tests $\indTests$, which is $\setObservedPDataset = \setObservedPTestSpecific{1} \cup \cdots \cup \setObservedPTestSpecific{\indTests}$ with $|\setObservedPDataset| = \numObservedP\leq 9$. 
Please note that throughout this paper, we index the sets $\setObservedPTesti$ to refer to specific observed components of the stress tensor.

Finally, similar to the observation map, we use deformation filters $\deformationFilterGeneral$ that filter out the reduced deformation vectors $\reducedFGeneral = \deformationFilterGeneral(\ten{F}) \elm{\numReducedFObservedP}$ from the deformation gradient. The reduced deformation vector $\reducedFGeneral$ only contains the deformation components that are relevant to predict the stress component $\setObservedPTesti_{\indObservedP}$. In the Treloar dataset, e.g., the stress component $\sca{P}_{11}$ depends only on the deformation components $\sca{F}_{11}$ and $\sca{F}_{22}$, such that $\reducedF^{(\mathrm{UT},1)}=\deformationFilterSpecific{\sca{P}_{11}}(\ten{F}) = [\sca{F}_{11}, \sca{F}_{22}]\transpose$. The remaining components of $\ten{F}$ are irrelevant with regard to $\sca{P}_{11}$, taking into account all mechanical tests carried out to collect the dataset. For the specific definitions of the deformation filters $\deformationFilterGeneral$, we refer to \cref{subsec:appendix_deformation_filters}.\\

A central idea of our framework is that we do not consider individual stress-deformation tensor pairs in isolation, but rather consider stress-deformation functions as entities. However, the various steps in our framework require the functions to be discretized. The vector $\func^{(\indTests,\indObservedP)} \elm{\numFuncPointsTesti}$ represents the discretized scalar-valued stress-deformation function for the $\indTests$-th mechanical test, which contains the $\indObservedP$-th stress component at a total of $\numFuncPointsTesti$ points. The corresponding functions for all tests can then be stacked in one vector as follows
\begin{equation}\label{eq:stacked_functions}
\func = \begin{bmatrix}
    \func^{(\indTests=1, \indObservedP=1)} \\
    \scriptstyle\vdots \\
    \func^{(\indTests=1, \indObservedP=\numObservedPTestSpecific{1})} \\
    \vdots \\
    \func^{(\indTests=\numTests, \indObservedP=1)} \\
    \scriptstyle\vdots \\
    \func^{(\indTests=\numTests, \indObservedP=\numObservedPTestSpecific{\numTests})} \\
\end{bmatrix}
\elm{\numFuncPoints} \quad \text{with} \quad \numFuncPoints = \sum_{\indTests=1}^{\numTests} \numObservedPTesti \numFuncPointsTesti,
\quad
\func^{(\indTests,\indObservedP)} \elm{\numFuncPointsTesti},
\end{equation}
where $\numFuncPoints$ is the total number of discretization points. Furthermore, the stresses in $\func^{(\indTests,\indObservedP)}$ are associated with the deformation gradients in $\funcPointsDeformationGradientsTesti = \bigl[\ten{F}^{(\indTests,\indFuncPoints=1)\transpose}, \dots, \allowbreak \ten{F}^{(\indTests,\indFuncPoints=\numFuncPointsTesti)\transpose}\bigr]\transpose \elmm{3 \numFuncPointsTesti}{3}$. If the dataset comprises several mechanical tests, the total number of discretized functions is $\numFuncs = \sum_{\indTests=1}^{\numTests} \numObservedPTesti$. 

The number of discretization points per test $\numFuncPointsTesti$ is a hyperparameter which determines the resolution of the discretized stress-deformation functions $\func^{(\indTests,\indObservedP)}$ and is generally not the same as $\numDataTesti$. Accordingly, the deformation gradients used to discretize the stress-deformation functions are not identical to the deformation gradients in $\dataSet$. Instead, we intentionally introduce a discretization of the stress-deformation functions that is totally independent of the measurements and provides us with the following advantages: First, we can choose the resolution of the function discretization independently of the measurement data, e.g., based on their expected smoothness. Second, we can also add discretization points between measurement points that may be sparse to augment the data and in consequence the resolution of the stress-deformation functions. Throughout this contribution, deformations are evenly distributed between the minimum and maximum deformation values for the respective mechanical test $\indTests$ taken from the dataset. Finally, the index-notation introduced in this section is summarized in \cref{tab:notation}.

\begin{table}[!htbp]
\centering
\caption{Summary of the notation.}
\label{tab:notation}

\begin{tabularx}{\textwidth}{
l
>{\raggedright\arraybackslash}X
}

\toprule

$\indTests$
& index for mechanical test \\
$\indObservedP$
& index for observed stress component \\
$\indData$
& index for measurement \\
$\numTests$
& number of mechanical tests \\
$\numObservedPTesti$
& number of observed stress components in test $\indTests$ ($\numObservedPTesti\leq 9$) \\
$\numObservedP$
& total number of observed stress components ($\numObservedP\leq 9$) \\
$\numDataTesti$
& number of measurements in test $\indTests$ \\

$\ten{F}^{(\indTests, \indData)}$
& deformation gradient for measurement $\indData$ in test $\indTests$ \\
$\sca{P}^{(\indTests, \indObservedP, \indData)}$
& observed stress component $\indObservedP$ for measurement $\indData$ in test $\indTests$ \\

$\observationMapGeneral$
& observation map for observed stress component $\indObservedP$ in test $\indTests$ \\
$\setObservedPTesti$
& set of observed stress components in test $\indTests$ ($|\setObservedPTesti| = \numObservedPTesti$) \\
$\setObservedP$
& union of all sets of observed stress components $\setObservedPTesti$ ($|\setObservedPDataset| = \numObservedP$) \\
$\setObservedPTesti_{\indObservedP}$
& observed stress component $\indObservedP$ from set $\setObservedPTesti$ \\
$\deformationFilterGeneral$
& deformation filter for observed stress component $\indObservedP$ in set $\setObservedPTesti$\\
$\reducedFGeneral$
& reduced deformation vector for stress component $\indObservedP$ in test $\indTests$ \\

$\func^{(\indTests,\indObservedP)}$
& discretized stress-deformation function for observed stress component $\indObservedP$ in test $\indTests$ \\
$\numFuncs$
& number of discretized stress-deformation functions \\
$\func$
& discretized stress-deformation function for all tests and observed  stress components \\
$\indFuncPoints$
& index for discretization point \\
$\numFuncPointsTesti$
& number of discretization points in test $\indTests$ \\
$\numFuncPoints$
& total number of discretization points \\
$\ten{F}^{(\indTests, \indFuncPoints)}$
& deformation gradient for discretization point $\indFuncPoints$ in test $\indTests$ \\

\bottomrule

\end{tabularx}
\end{table}

\subsection{Distilling statistical constitutive models from GP posteriors}\label{subsec:model_distillation}

We propose a framework for \ac{UQ} in model discovery that solves the variational problem in \cref{eq:bayes_variational_approximation} and further reduces the inferred \ac{PDF} $\materialParamsDist$ to the most relevant material parameters in four subsequent steps:

\begin{enumerate}[label=(\roman*)]

    \item For each observed stress component in $\setObservedP$, we train a \ac{GP} on a subset of the dataset $\dataSet$. In the following, we refer to the set of \acp{GP} for the different observed stress components as an independent multi-output \ac{GP}. The inferred \ac{GP} posterior $\funcDistGP$ models the probability distribution over stress-deformation functions conditioned on the data $\dataSet$, i.e., $\funcDistGP \approx p(\func \mid \dataSet)$, which is required for the variational problem formulation in \cref{eq:bayes_variational_approximation}. In subsequent steps, we use the \ac{GP} posterior for data augmentation, drawing on the idea of generative modeling. Note that the \ac{GP} posterior may not be physically consistent. Thus, stress-deformation functions $\funcGP$ sampled from the \ac{GP}, i.e., $\funcGP \sim \funcDistGP$, may violate the aforementioned physical constraints, such as thermodynamic consistency.

    \item We distill a physically consistent and interpretable statistical constitutive model from the \ac{GP} posterior. For this purpose, we match the distribution $\funcDistGP$ on the one hand and the distribution defined by the statistical constitutive model $\funcDistModel$ on the other by solving the variational problem \cref{eq:bayes_variational_approximation}. Note that the distribution $\funcDistModel$ over stress-deformation functions corresponds to the \ac{PDF} of the push-forward measure $\parametersToStateMapPush\materialParamsDistPush$ in \cref{eq:bayes_variational_approximation} and is induced by the distribution of the material parameters $\materialParamsDist$. Throughout this paper, $\funcDistModel$ is determined by $\funcDistModel = \int \delta(\func - \parametersToStateMap(\materialParams)) \, \materialParamsDist \, \mathrm{d}\materialParams$, where $\delta$ is the Dirac delta function. The deterministic map $\parametersToStateMap: \materialParams \mapsto \vec{g} \elm{\numFuncPoints}$ is implicitly defined by \cref{eq:first_PK_from_SEF_incompressible} and the generalized model library for the \ac{SEF} in \cref{eq:ansatz_sef_general}. In this contribution, we approximate the distribution of the material parameters by a \ac{NF} $\materialParamsDistNF$ parameterized in $\nfParams$, i.e., $\materialParamsDist \approx \materialParamsDistNF$. In comparison to standard distributions, such as multivariate Gaussian distributions, using \acp{NF} also enables us to approximate more complex, high-dimensional joint distributions. The distribution of material parameters, in turn, induces a parameterized distribution over stress-deformation functions as follows
    \begin{equation}\label{eq:parameterized_functional_dustribution}
        \funcDistModelNF = \int \delta(\func - \parametersToStateMap(\materialParams)) \, \materialParamsDistNF \, \mathrm{d}\materialParams.
    \end{equation}
    In order to match the distributions, we minimize the Wasserstein-1 distance $\mathcal{W}_{1}(\funcDistGP, \funcDistModelNF)$ between them with respect to the distributional parameters $\nfParams$ according to the variational problem formulation \cref{eq:bayes_variational_approximation}. An advantageous property of the Wasserstein-1 distance is that this metric can be estimated solely from samples of both distributions, which in our case can be easily generated. Moreover, in view of \eqref{eq:parameterized_functional_dustribution} and for $\numMaterialParams$ smaller than $\numFuncPoints$, the density $\funcDistModelNF$ will be a low-dimensional sub-manifold of $\numFuncPoints$. Hence, the measures associated to $\funcDistModelNF$ and $\funcDistGP$ will be mutually singular. The Wasserstein-1 distance is still applicable in this setting, whereas the Kullback-Leibler divergence and other distances are not. Since the deterministic map $\parametersToStateMap$ satisfies all aforementioned constraints by construction, all functions $\funcModel \sim \funcDistModelNF$ are physically consistent. However, after this step, the model does not necessarily have to be sparse and $\materialParamsDistNF$ is defined as a joint distribution of all parameters that were originally included in the model library. The sparsity of $\materialParamsDistNF$ is induced in the next step.

    \item We reduce the joint distribution $\materialParamsDistNF$ to the most relevant material parameters to promote interpretability and generalization. Therefore, we perform a sensitivity analysis with respect to the material parameters $\materialParams \sim \materialParamsDistNF$ and remove all non-relevant parameters from the joint distribution $\materialParamsDistNF$. As a measure of sensitivity, we consider the total-order Sobol' indices of the material parameters. We then remove all material parameters whose total sensitivity indices fall below a predefined threshold.

    \item We perform a recalibration of the constitutive model discovered in steps (i)-(iii) by repeating step (ii) but now using only the relevant model terms and corresponding material parameters. Finally, we obtain the interpretable, physically consistent and sparse statistical constitutive model
    \begin{equation}\label{eq:discovered_reduced_model}
      \widebarsca{W}\bigl( \ten{F}, \{\structuralTensorGeneral\}_{\indStructuralTensors=1}^{\numStructuralTensors}; \reducedMaterialParams\bigr) =
      \sum_{\indModelTerms=1}^{\numReducedModelTerms} \materialParamLinear^{(\indModelTerms)} \, \modelTerm^{(\indModelTerms)} \bigl(\ten{F}, \{\structuralTensorGeneral\}_{\indStructuralTensors=1}^{\numStructuralTensors}; \materialParamNonlinear^{(\indModelTerms)} \bigr),
      \quad \reducedMaterialParams \sim \reducedMaterialParamsDistNF,
    \end{equation}
    where $\numReducedModelTerms \ll \numModelTerms$ denotes the number of relevant model terms $\modelTerm^{(\indModelTerms)}$ and $\reducedMaterialParams \in \mathbb{R}^{\numReducedMaterialParams}$ the sparse vector of material parameters with $\numReducedMaterialParams \ll \numMaterialParams$. Recalibration aims to refine the statistical model and eliminate possible dependencies on terms removed in step (iii). In general, it can be assumed that the accuracy of the approximation $\reducedMaterialParamsDistNF$ increases for smaller $\numReducedMaterialParams$ since with the number of parameters also the complexity of the joint distribution decreases. It is generally possible that even after recalibration some material parameters render non-sensitive and can be removed. In this case, a further recalibration step may be useful.
    
\end{enumerate}

The framework outlined above remains generic and particularly suitable for model discovery. Instead of a fully Bayesian approach, we propose a partially Bayesian two-step inference procedure that does not require prior selection for the material parameters. In the first step, a \ac{GP} posterior is inferred from the available stress-deformation data. In a second step, we then distill a physically consistent and interpretable statistical constitutive model from the \ac{GP} posterior. Furthermore, \acp{NF} are very flexible and enable the approximation of complex high-dimensional distributions. Therefore, we do not need to make strong assumptions about the type of distribution of the material parameters or their correlation. In addition, \acp{NF} can be directly used for both sampling and density estimation
\cite{papamakarios_normalizingFlows_2021}, whereas \ac{MCMC}-based methods provide only samples of material parameters. Finally, the Sobol' sensitivity analysis in step (iii) provides further insights into the model selection process as we show in our numerical tests. It is worth noting that, in general, our framework can also be used for model calibration. In this special case, steps (iii) and (iv) are omitted. The complete four-step approach to \ac{UQ} in model discovery is visualized in \cref{fig:workflow}. Steps (i)-(iii) and the corresponding methods are explained in more detail in the following subsections.\\

\noindent \textbf{Remark:} The two-step inference procedure introduces an additional approximation step, and the associated error must be carefully controlled. In a one-step approach, instead one could consider the model $\funcDistGP = \gp(\parametersToStateMap(\materialParams),\mathbf{K})$ and infer $\materialParams$ as hyperparameters of the \ac{GP}. Although this procedure would be more principled from a statistical point of view, assigning a prior and inferring $\materialParams$ within this framework poses a significant challenge. Using a \ac{NF} to approximate the distribution over the material parameters would add another hierarchy and additional complexity. We therefore prefer to omit formulating a prior and inferring $\materialParams$ as hyperparameters.

\newcommand{\colorWorkflowComponent}{teal!50!gray!20}
\newcommand{\colorWorkflowDistribution}{cyan!50!gray!20}
\newcommand{\colorWorkflowResult}{teal!30!}
\newcommand{\colorWorkflowProcess}{purple!50!gray!20}
\begin{figure}[ht!]
    \centering
    \resizebox{0.92\textwidth}{!}{
    \begin{tikzpicture}

    \tikzstyle{component} = [
        shape=rectangle, 
        minimum width=2.5cm, 
        minimum height=1cm,
        inner xsep = 2mm,
        inner ysep = 2mm,
        align = center,
        font=\scriptsize\linespread{2},
        draw=gray, 
        fill=\colorWorkflowComponent
    ]
    \tikzstyle{functionaldistribution} = [
        shape=rectangle,
        minimum width=2.5cm, 
        minimum height=1cm,
        inner xsep = 2mm,
        inner ysep = 2mm,
        align = center,
        font=\scriptsize\linespread{2},
        draw=gray, 
        fill=\colorWorkflowDistribution
    ]
    \tikzstyle{functionaldistributionscope} = [
        shape=rectangle, 
        rounded corners, 
        minimum width=2.5cm, 
        minimum height=2.9cm,
        inner xsep = 2mm,
        inner ysep = 2mm,
        align = center,
        font=\scriptsize\linespread{2},
        draw=gray, 
        fill=\colorWorkflowDistribution
    ]
    \tikzstyle{result} = [
        shape=rectangle, 
        minimum width=2.5cm, 
        minimum height=1cm,
        inner xsep = 2mm,
        inner ysep = 2mm,
        align = center,
        font=\scriptsize\linespread{2},
        draw=teal, 
        fill=\colorWorkflowResult
    ]
    \tikzstyle{process} = [
        shape=rectangle, 
        rounded corners, 
        minimum width=2.5cm, 
        minimum height=0.5cm,
        inner xsep = 2mm,
        inner ysep = 2mm,
        align = center, 
        font=\scriptsize,
        draw=purple, 
        fill=\colorWorkflowProcess
    ]
    \tikzstyle{processscope} = [
        shape=rectangle, 
        rounded corners, 
        minimum width=2.5cm, 
        minimum height=4.3cm,
        inner xsep = 2mm,
        inner ysep = 2mm,
        align = center, 
        font=\scriptsize,
        draw=purple, 
        fill=\colorWorkflowProcess
    ]
    \tikzstyle{headernode} = [
        minimum width=2.0cm,
        font=\strut\ttfamily,
        align=center,
        text depth=+.3ex, 
        draw=#1,
        fill=white,
    ]
    \tikzstyle{header} = [
        inner ysep=+4mm,
        append after command={
            \pgfextra{\let\TikZlastnode\tikzlastnode}
            node [headernode] (header-\TikZlastnode) at (\TikZlastnode.north) {#1}
            node [span=(\TikZlastnode)(header-\TikZlastnode)] at (fit bounding box) (h-\TikZlastnode) {}
        }
    ]
    \newcommand{\stepNumberShapeSize}{0.8cm}
    \tikzstyle{stepNumber} = [
        shape=circle, 
        minimum width=\stepNumberShapeSize, 
        minimum height=\stepNumberShapeSize,
        inner xsep = 2mm,
        inner ysep = 2mm,
        align = center, 
        font=\small,
        draw=orange, 
        fill=yellow!70!gray!20
    ]
    \tikzstyle{arrow} = [thick,->,>=stealth, rounded corners]

    \node (dataset) [
        component, 
        header = dataset,
        xshift = -0.5\linewidth
    ] {$\dataSet = \bigl\{ \ten{F}^{(\indData)}, \ten{P}^{(\indData)} \bigr\}_{\indData=1}^{\numData}$};
    
    \node (gptraining) [
        process,
        below = 0.3cm of dataset,
        xshift = 0.65cm,
        minimum width = 1.3cm,
    ]{train GP};
    
    \node (stepNumberOne) [
        stepNumber,
        left = 0.0cm of gptraining
    ]{1};

    \node (mechanics) [
        component,
        header = mechanics,
        right = 0.5cm of dataset
    ] {$\ten{P} = \frac{\partial \widebarsca{W}(\ten{F}; \materialParams)}{\partial \ten{F}} - \sca{p} \ten{F}\minusTranspose$};

    \node (modellibrary) [
        component,
        header = model library,
        right = 0.5cm of mechanics
    ] {$\widebarsca{W}(\ten{F};\materialParams) = 
    \sum_{\indModelTerms=1}^{\numModelTerms} 
    \materialParamLinear^{(\indModelTerms)} \, \modelTerm^{(\indModelTerms)} (\ten{F}; \materialParamNonlinear^{(\indModelTerms)})$
    };

    \node (mechanicalmodel) [
        component,
        header = mechanical model,
        below = 1.0cm of modellibrary,
    ] {$\parametersToStateMap(\materialParams) = 
        \begin{bmatrix*}[c]
            \sca{P}^{(1)}(\materialParams)\\
            \vdots\\
            \sca{P}^{(\numFuncPoints)}(\materialParams)\\
        \end{bmatrix*}$};

    \node (normalizingflow) [
        component,
        header = normalizing flow,
        left = -0.5cm of mechanicalmodel,
        xshift = -1.cm
    ] {$\materialParamsDistNF = \nfBaseSamplesDist \biggl| \det \frac{\partial \nfTransform(\nfBaseSamples;\nfParams)}{\partial \nfBaseSamples} \biggr|^{-1}$};

    \draw [arrow] (mechanics) -- +(0,-1cm) -| (h-mechanicalmodel);
    \draw [arrow] (modellibrary) -- (h-mechanicalmodel);

    \node (gpposterior) [
        functionaldistribution,
        header = GP posterior,
        below = 6.2cm of dataset.west,
        anchor = west
    ] {$\funcGP \sim \funcDistGP = \gp(\vec{0}, \gpCovarianceMatrix)$};
    
    \draw [arrow] (dataset) -- +(0,-4.5cm) -| (h-gpposterior);

    \node (statisticalmodel) [
        functionaldistribution,
        header = statistical model,
        below = 6.2cm of modellibrary.east,
        anchor = east
    ] {$\funcModel \sim \funcDistModelNF = \int \delta(\func - \parametersToStateMap(\materialParams)) \materialParamsDistNF \mathrm{d}\materialParams$};

    \scoped[on background layer]
        \node (functionaldistributions) [
            functionaldistributionscope,
            header = distribution of \\ discretized functions,
            fit=(gpposterior)(statisticalmodel)
        ]{};

    \draw [arrow] (normalizingflow) -- +(0,-1.2cm) -| (h-statisticalmodel);
    \draw [arrow] (mechanicalmodel) -- +(0,-1.2cm) -| (h-statisticalmodel);

    \node (wasserstein) [
        process,
        header = Wasserstein-1 minimization,
        below = 5.5cm of gpposterior.west,
        anchor=west,
        xshift = \stepNumberShapeSize
    ]{
        $\{\nfParamsOptimal, \lipschitzParamsOptimal \} = \argmin_{\nfParams} \argmax_{\lipschitzParams} 
        \bigl[
            \expectation{\funcDistGPMeasure}\bigl(\wassersteinLipschitzNetwork(\funcGP; \lipschitzParams)\bigr) - \expectation{\funcDistModelMeasure(\cdot;\nfParams)}\bigl(\wassersteinLipschitzNetwork(\funcModel; \lipschitzParams)\bigr)
        \bigr]$ 
    };
    \node (stepNumberTwo) [
        stepNumber,
        left = 0.0cm of wasserstein,
    ]{2};

    \node (lipschitznetwork) [
        component,
        header = {Lipschitz-1\\ network},
        below = 3.8cm of statisticalmodel.east,
        anchor=east,
        inner ysep = 6mm
    ]{$\wassersteinLipschitzNetwork(\func; \lipschitzParams)$};

    \scoped[on background layer]
        \node (distillation) [
            processscope,
            header = distillation of \\ parameter distribution,
            fit=(stepNumberTwo)(wasserstein)(lipschitznetwork),
        ]{};

    \draw [arrow] (gpposterior) -- +(0,-1cm) -| (h-distillation);
    \draw [arrow] (statisticalmodel) -- +(0,-1cm) -| (h-distillation);
    \draw [arrow] (lipschitznetwork.west)  -| (h-wasserstein);

    \node (sensitivity) [
        process,
        header = sensitivity analysis,
        below = 8.8cm of gpposterior.west,
        anchor=west,
        xshift = \stepNumberShapeSize
    ]{
        $\totalSobolIndex(\materialParamOne) = 1 - \frac{
        \variance{\materialParams_{\sim\indMaterialParam}} \Bigl( 
            \expectation{\materialParamOne} \bigl( 
               \modelOutput(\materialParams) \mid \materialParams_{\sim\indMaterialParam}
            \bigr)
        \Bigr)
        }{\mathbb{V} \bigl(
            \modelOutput(\materialParams)
        \bigr)
        }$ 
    };
    \node (stepNumberThree) [
        stepNumber,
        left = 0.0cm of sensitivity,
    ]{3};

    \draw [arrow] (wasserstein) -- +(0,-1.0cm) -| (h-sensitivity);

    \node (result) [
        result,
        header = {parameter\\ distribution},
        below = 8.8cm of statisticalmodel.east,
        anchor=east,
        inner ysep = 6mm
    ]{
        $p_{\reducedMaterialParams}(\reducedMaterialParams; \nfParams^{*})$ 
    };

    \node (refinement) [
            process,
            below = 0.6cm of distillation,
            xshift = 1.8cm,
            minimum width = 1.3cm,
    ]{refine};
    \node (stepNumberFour) [
        stepNumber,
        left = 0.0cm of refinement
    ]{4};

    \draw [arrow] (sensitivity) -- (result);
    \draw [arrow] (sensitivity.east) -| ([xshift=2.45cm]distillation);
        
    \end{tikzpicture}}
    \caption{Workflow for the quantification of uncertainty in model discovery by distilling interpretable material constitutive models from \ac{GP} posteriors. In above workflow, the color \raisebox{-0.5pt}{\color{\colorWorkflowComponent}\normalsize $\blacksquare$} encodes components, such as the data or the model library, \raisebox{-0.5pt}{\color{\colorWorkflowDistribution}\normalsize $\blacksquare$} the distributions over discretized stress-deformation functions, \raisebox{-0.5pt}{\color{\colorWorkflowProcess}\normalsize $\blacksquare$} processes, such as the Wasserstein-1 minimization or the sensitivity analysis, and \raisebox{-0.5pt}{\color{\colorWorkflowResult}\normalsize $\blacksquare$} the resulting joint distribution over the relevant material parameters. The numbers \num{1} to \num{4} in yellow refer to the four steps of the proposed framework which are (\num{1}) \ac{GP} training (\cref{subsubsec:gps}), (\num{2}) model distillation (\cref{subsubsec:normalizing_flows} and \cref{subsubsec:wasserstein_minimization}), (\num{3}) sensitivity analysis and model sparsification (\cref{subsubsec:sensitivity_analysis}), and (\num{4}) model recalibration.}
    \label{fig:workflow}
\end{figure}

\subsubsection{Gaussian process posterior}\label{subsubsec:gps}

We infer an independent multi-output \ac{GP} posterior from the dataset $\dataSet$ which is used for data augmentation in the subsequent steps. In the following, we consider the components of the stress tensor to be independent of each other and model each component that is observed at least in one mechanical test separately with a single-output \ac{GP}. Furthermore, we use the reduced deformation vectors $\reducedFGeneral = \deformationFilterGeneral(\ten{F}) \elm{\numReducedFObservedP}$, introduced in \cref{subsec:dataset_and_notation}, as input to the \ac{GP} for the stress component $\setObservedPTesti_{\indObservedP}$. This allows us to reduce the number of \ac{GP} hyperparameters. For the specific definitions of the deformation filters $\deformationFilterGeneral$, we refer to \cref{subsec:appendix_deformation_filters}. 

In the following, we explain the inference of the \ac{GP} posterior for a single stress component. However, for the sake of clarity, we omit the explicit dependence on indices $\indTests$ and $\indObservedP$ in the notation such that $\reducedF := \reducedFGeneral$, $\gp := \gpGeneral$, $\numData := \numDataTesti $, $\numReducedF = \numReducedFObservedP$, etc. For simplicity, we take the mean functions of the \acp{GP} to be zero, i.e., $\gpMeanFunction(\reducedF) = \sca{0}$. The \acp{GP} are further specified by a covariance function $\gpCovarianceFunction(\reducedF, \reducedFDifferent; \gpParams)$ with hyperparameters $\gpParams$. We choose scaled squared-exponential covariance functions which are defined as follows
\begin{equation}\label{eq:gp_squared_exponential_covariance_func}
    \gpCovarianceFunction(\reducedF, \reducedFDifferent; \gpParams) = 
    \gpParamsOutputScaleSquared \operatorname{exp}\Biggl(
        -\half \sum_{i=1}^{\numReducedF} \biggl(
            \frac{\reducedFComponent_{i} - \reducedFComponentDifferent_{i}}{\gpParamsLengthScalesComponent_{i}}
        \biggr)^{2}
    \Biggr),
\end{equation}
since sample paths and the mean functions of \acp{GP} with squared exponential kernel functions are smooth \cite{rasmussen_gaussianProcesses_2005}. Here, $\gpParamsLengthScales \elm{\numReducedF}$ comprises the length scales for each input dimension and $\gpParamsOutputScale \in \mathbb{R}$ is the output scale, i.e., $\gpParams = [\gpParamsLengthScales\transpose, \gpParamsOutputScale]\transpose$. Finally, the joint distribution of all stresses corresponding to the $\numData$ deformation states that comprise the random vector $\observedPGeneral = [\sca{P}^{(1)}, \dots, \sca{P}^{(\numData)}]\transpose$ is
\begin{equation}\label{eq:gp_joint_distribution}
    \observedPGeneral \sim 
    \gp\bigl( \vec{0}, \gpCovarianceMatrix(\reducedFAll,\reducedFAll;\gpParams)  \bigr),
\end{equation}
with $\observedPGeneral: \sampleSpaceObservedPGeneral \rightarrow \mathbb{R}^{\numData}$, where $\sampleSpaceObservedPGeneral$ is the sample space of $\observedPGeneral$. Furthermore, $\reducedFAll = [\reducedF^{(1)}, \dots, \reducedF^{(\numData)}] \elmm{\numReducedF}{\numData}$ denotes the matrix of all reduced deformation inputs and $\gpCovarianceMatrix$ is the covariance matrix with $\gpCovarianceMatrixElement = \gpCovarianceFunction(\reducedF^{(i)},\reducedF^{(j)};\gpParams)$.\\

\noindent \textbf{GP prior:} First, we fit the hyperparameters $\{\gpParamsObservedPDataset\}_{\indObservedP=1}^{\numObservedP}$ of the $\numObservedP$ \acp{GP} to the dataset $\dataSet$. 
In order to fit the \acp{GP} to the training data, we select the hyperparameters $\gpParamsObservedPDataset$ by maximizing the marginal logarithmic likelihood \cite{rasmussen_gaussianProcesses_2005}, which is also known as the empirical Bayes method. 

In our statistical framework, we assume that the measured stresses $\observedPGeneral$ are noisy observations of the true but hidden stresses $\observedPGeneralTrue$ with independent additive Gaussian noise $\noiseTerm$. Since the measured stress-deformation curves vary for different samples of the same material, the data model contains an additional term $\samplesVariabilityTerm$ that describes the variability for different samples, i.e., $\observedPGeneral = \observedPGeneralTrue + \noiseTerm + \samplesVariabilityTerm$. The sample variability term is also assumed to be Gaussian distributed. Depending on the tested material and the mechanical test, the uncertainty resulting from $\samplesVariabilityTerm$ may exceed the measurement noise $\noiseTerm$. Both the measurement noise and the sample variability represent inherent randomness in mechanical testing and thus are aleatoric uncertainties. In this contribution, we assume $\samplesVariabilityTerm$ to consist of independent entries that are also independent of $\noiseTerm$ and combine both measurement noise $\noiseTerm$ and sample variability $\samplesVariabilityTerm$ into one Gaussian error term $\errorTerm = \noiseTerm + \samplesVariabilityTerm$. Furthermore, we assume that both the measurement noise and the sample variability are functions of the amount of stress. Thus, for the total error term $\errorTerm$, we use a heteroscedastic Gaussian error model 
\begin{equation}\label{eq:gp_error_model}
    \errorTerm = \mathcal{N}\bigl(
        \vec{0}, 
        \errorCovariance(\minErrorStddev, \relativeErrorStddev, \observedPGeneralTrue) 
    \bigr),
\end{equation}
with zero mean and a positive definite covariance matrix $\errorCovariance$. The covariance matrix is a function of the minimum error standard deviation $\minErrorStddev$, the relative error standard deviation $\relativeErrorStddev$ and the stresses. It is defined as
\begin{equation}\label{eq:gp_error_model_covariance}
    \errorCovariance(\minErrorStddev, \relativeErrorStddev, \observedPGeneralTrue) = 
    \operatorname{diag}\bigl(
        \operatorname{max}\bigl\{\minErrorStddev^{2} \vec{1}, \relativeErrorStddev^{2} \, \observedPGeneralTrueSquared\bigr\}
    \bigr),
\end{equation}
where $\observedPGeneralTrueSquared$ are the element-wise squares of the true stresses, $\vec{1} \elm{\numData}$ is a vector of ones, and $\operatorname{max}$ is the operator which selects the element-wise maximum of the two passed vectors. Thus, the data model simplifies to $\observedPGeneral \approx \observedPGeneralTrue + \errorTerm$. We are aware that our assumptions about the error model might not fully reflect reality, since, in particular, the variability term is generally neither Gaussian nor additive. However, we expect that the datasets we use in our numerical tests contain the averaged stress measurements for several samples. Since the error contribution of each sample is lower on average, we can model the error contributions more roughly, and the effects of our simplified data model are limited. Ultimately, since the numerical tests did not reveal any problems with our assumptions, we consider these assumptions reasonable for the model discovery task presented in this work.

As the \ac{GP} corresponds to a multivariate Gaussian distribution and the assumptions on the noise defined in \cref{eq:gp_error_model} and \cref{eq:gp_error_model_covariance} also lead to a Gaussian likelihood, there exists a closed form for the logarithmic marginal likelihood which yields
\begin{equation}\label{eq:gp_log_marginal_likelihood}
    \log p(\observedPGeneral \mid \reducedFAll; \gpParams) =
    - \half \observedPGeneralTransposed 
    \gpCovarianceMatrixNoise(\reducedFAll, \reducedFAll; \gpParams)^{-1} \observedPGeneral \\
    - \half \log \det \gpCovarianceMatrixNoise(\reducedFAll, \reducedFAll; \gpParams)
    - \frac{\numData}{2} \log 2 \pi,
\end{equation}
with $\gpCovarianceMatrixNoise(\reducedFAll, \reducedFAll; \gpParams) = \gpCovarianceMatrix(\reducedFAll, \reducedFAll; \gpParams) + \errorCovariance(\minErrorStddev, \relativeErrorStddev, \observedPGeneralTrue)$. In order to find an appropriate point estimate for the hyperparameters $\gpParams$, for each \ac{GP}, we define the optimization problem
\begin{equation}\label{eq:gp_optimization_problem}
    \gpParamsOptimal = \argmax_{\gpParams}  \log p(\observedPGeneral \mid \reducedFAll; \gpParams),
\end{equation}
and optimize their hyperparameters using the AdamW gradient-based optimization algorithm \cite{loshchilov_AdamW_2018}. However, we observed that the length scales obtained by solving the optimization problem in \cref{eq:gp_optimization_problem} tend to be too large. Excessively large length scales eventually induce a class of stress-deformation functions that vary only slowly with their input and, in fact, may cause an underestimation of the uncertainty. In general, the uncertainty intervals become narrower as the length scales increase, see \cite{rasmussen_gaussianProcesses_2005}. Therefore, we reduce the optimized length scales by a factor of \num{0.6}. This factor is determined manually, striking a balance between physical consistency and the coverage of the \ac{GP} posterior as a measure of how well it quantifies the uncertainty in the data. For a definition of the coverage, see \cref{sec:appendix_ci_and_estimated_coverage}. For completeness, we would like to point out that in our implementation we normalize the reduced deformation inputs to the range $[0, 1]$. Finally, the \acp{GP} with the selected hyperparameters then represent our prior belief about the stochastic processes behind the observed stress-deformation functions. \\

\noindent \textbf{GP posterior:} Second, we condition the \ac{GP} prior on the corresponding subset of the observed data points in $\dataSet$ and thus infer the \ac{GP} posterior. Through conditioning, the mean and covariance functions change as follows
\begin{equation}\label{eq:gp_posterior_mean_and_covariance}
\begin{aligned}
    \gpMeanVectorPosterior(\reducedFUnseen; \gpParamsOptimal) &= 
    \gpCovarianceMatrix(\reducedFUnseen, \reducedF; \gpParamsOptimal)
    \gpCovarianceMatrixNoise(\reducedF, \reducedF; \gpParamsOptimal)^{-1}
    \observedPGeneral,\\
    \gpCovarianceMatrixPosterior(\reducedFUnseen, \reducedFUnseen; \gpParamsOptimal) &= 
    \gpCovarianceMatrix(\reducedFUnseen, \reducedFUnseen; \gpParamsOptimal)
    - \gpCovarianceMatrix(\reducedFUnseen, \reducedF; \gpParamsOptimal)
    \gpCovarianceMatrixNoise(\reducedF, \reducedF; \gpParamsOptimal)^{-1}
    \gpCovarianceMatrix(\reducedF, \reducedFUnseen; \gpParamsOptimal),
\end{aligned}
\end{equation}
where ${\reducedF, \observedPGeneral}$ denote the training data and ${\reducedFUnseen, \observedPGeneralUnseen}$ the unseen data. From the function space view, the \ac{GP} posterior describes a distribution over functions conditioned on the measured data \cite{rasmussen_gaussianProcesses_2005}. Since we are interested in the posterior over the stress-deformation functions and not in noisy predictions, we consider the \ac{GP} posterior instead of the predictive \ac{GP} posterior; please refer to \cite{rasmussen_gaussianProcesses_2005} for more details. The covariance of the \ac{GP} posterior in \cref{eq:gp_posterior_mean_and_covariance} decreases when more training data is used. From this, we can conclude that the uncertainty of the \ac{GP} posterior is epistemic. 

Finally, from the \ac{GP} posteriors, we can sample discretized stress-deformation functions stacked in one random vector according to \cref{eq:stacked_functions} as follows
\begin{equation}\label{eq:sampled_functions_gp}
\funcGP \sim
    \begin{bmatrix*}[c]
        \gpSpecific{\indTests=1}{1} \Bigl( 
        \gpMeanVectorPosterior(\reducedFAll^{(1, 1)}), \gpCovarianceMatrixPosterior(\reducedFAll^{(1, 1)}, \reducedFAll^{(1, 1)})  
        ; \gpParamsOptimalObservedPSpecific{1}{1}
        \Bigr) \\
        \scriptstyle\vdots \\
        \gpSpecific{\indTests=1}{\numObservedPTestSpecific{1}} \Bigl( \gpMeanVectorPosterior(\reducedFAll^{(1, \numObservedPTestSpecific{1})}), \gpCovarianceMatrixPosterior(\reducedFAll^{(1, \numObservedPTestSpecific{1})}, \reducedFAll^{(1, \numObservedPTestSpecific{1})})
        ; \gpParamsOptimalObservedPSpecific{1}{\numObservedPTestSpecific{1}}
        \Bigr) \\
        \vdots \\
        \gpSpecific{\indTests=\numTests}{1} \Bigl( 
        \gpMeanVectorPosterior(\reducedFAll^{(\numTests, 1)}), \gpCovarianceMatrixPosterior(\reducedFAll^{(\numTests, 1)}, \reducedFAll^{(\numTests, 1)})
        ; \gpParamsOptimalObservedPSpecific{\numTests}{1}
        \Bigr) \\
        \scriptstyle\vdots \\
        \gpSpecific{\indTests=\numTests}{\numObservedPTestSpecific{\numTests}} \Bigl( \gpMeanVectorPosterior(\reducedFAll^{(\numTests, \numObservedPTestSpecific{\numTests})}), \gpCovarianceMatrixPosterior(\reducedFAll^{(\numTests, \numObservedPTestSpecific{\numTests})}, \reducedFAll^{(\numTests, \numObservedPTestSpecific{\numTests})})
        ; \gpParamsOptimalObservedPSpecific{\numTests}{\numObservedPTestSpecific{\numTests}}
        \Bigr)
    \end{bmatrix*},
\end{equation}
with $\funcGP: \sampleSpaceFuncGP \rightarrow \mathbb{R}^{\numFuncPoints}$, where $\sampleSpaceFuncGP$ is the sample space of $\funcGP$. Furthermore, $\reducedFAll^{(\indTests, \indObservedP)} = [\deformationFilterGeneral(\ten{F}^{(\indTests, 1)}), \dots, \deformationFilterGeneral(\ten{F}^{(\indTests, \numFuncPointsTesti)})] \elmm{\numReducedFObservedP}{\numFuncPointsTesti}$ is the matrix comprising all reduced deformation inputs for the $\indObservedP$-th observed stress component in the $\indTests$-th mechanical test. Again, $\numFuncPointsTesti$ indicates the number of points at which the stress-deformation function for the test $\indTests$ is discretized and $\numFuncPoints$ the total number of points for all $\numFuncs$ functions stacked in $\funcGP$. For simplicity, in \cref{eq:sampled_functions_gp}, we use $\gp\bigl( \gpMeanVectorPosterior(\reducedFAll), \gpCovarianceMatrixPosterior(\reducedFAll, \reducedFAll); \gpParams \bigr)$ as short form for $\gp\bigl( \gpMeanVectorPosterior(\reducedFAll; \gpParams), \gpCovarianceMatrixPosterior(\reducedFAll, \reducedFAll; \gpParams) \bigr)$.\\

\noindent \textbf{Remark:} It is worth noting that, for hyperelastic materials, all stress components are derived from the \ac{SEF} and are coupled. However, in the present work, for a given deformation state, the data typically provide measurements only for a small subset of stress components. Therefore, the observations available to identify the correlation between the stress components are very limited. As a consequence, adopting a correlated multi-output \ac{GP} would necessitate additional structural assumptions or strong priors on the correlation, which are typically also very limited in model discovery. For these reasons, we adopt independent multi-output \acp{GP} for the observed stress components as a pragmatic choice for constructing the target distribution used in the distillation step. Importantly, physical coupling is reintroduced at the distillation stage. Developing a surrogate model that incorporates physically informed cross-component correlations remains an important direction for future work.

\subsubsection{Normalizing flows}\label{subsubsec:normalizing_flows}

We approximate the joint distribution $\materialParamsDist$ using a \ac{NF} which enables modeling of complex, high-dimensional and multi-modal distributions and potential dependencies between the material parameters \cite{papamakarios_normalizingFlows_2021}.

The idea of \acp{NF} is to express the random vector $\materialParams$ with values in $\mathbb{R}^{\numMaterialParams}$ via a bijective transport map based on another $\numMaterialParams$-dimensional random vector $\nfBaseSamples$. However, the vector $\nfBaseSamples$ can be sampled from the base distribution which is generally simpler than the one expected for $\materialParams$. A common choice for the base distribution is a standard multivariate normal distribution \cite{papamakarios_normalizingFlows_2021}, which we also choose in our numerical tests. The transformation $\nfTransform$ with distributional parameters $\nfParams$ then induces a distribution over the random vector $\materialParams$ as follows 
\begin{equation}\label{eq:nf_induced_distribution}
    \materialParams = \nfTransform(\nfBaseSamples; \nfParams) \quad \text{with} \quad \nfBaseSamples \sim \mathcal{N}(\vec{0}, \ten{I}),
\end{equation}
where $\vec{0} \elm{\numMaterialParams}$ is a vector of zeros and $\ten{I} \elmm{\numMaterialParams}{\numMaterialParams}$ is the identity matrix, respectively \cite{papamakarios_normalizingFlows_2021}. Provided that the transformation $\nfTransform$ is invertible and both $\nfTransform$ and $\nfTransform^{-1}$ are differentiable, the probability density $\materialParamsDistNF$ can be obtained by a change of variables \cite{rudin_realAndComplexAnalysis_1987,bogachev_measureTheory_2006} as follows
\begin{equation}\label{eq:nf_probability_density}
    \materialParamsDistNF 
    = \nfBaseSamplesDist \biggl| \det \frac{\partial \nfTransform(\nfBaseSamples;\nfParams)}{\partial \nfBaseSamples} \biggr|^{-1}
    = p_{\mathrm{u}}\bigl(\nfTransform^{-1}(\materialParams;\nfParams)\bigr) \biggl| \det \frac{\partial \nfTransform^{-1}(\materialParams; \nfParams)}{\partial \materialParams} \biggr|,
\end{equation}
where $\nfTransform^{-1}(\materialParams;\nfParams)$ is the inverse transformation. In general, the parameters of the base distribution, such as the standard deviation, can also be trainable. However, in theory, the parameters of the base distribution can be absorbed in the transformation and thus be considered to be fixed. 

The transformation $\nfTransform$ is usually a composition of a finite number of sub-transformations with the general form
\begin{equation}\label{eq:nf_transformation_composition}
\begin{aligned}
    \nfTransform: \mathbb{R}^{\numMaterialParams} &\to \mathbb{R}^{\numMaterialParams}, \\
    \nfBaseSamples &\mapsto 
    \nfTransform(\nfBaseSamples;\nfParams) = 
    \bigl( \nfTransform^{(\nfNumTransforms + 1)}(\bullet;\nfParams^{(\nfNumTransforms + 1)}) \circ \dots \circ \nfTransform^{(1)}(\bullet;\nfParams^{(1)}) \bigr)  
    \bigl(\nfBaseSamples \bigr).
\end{aligned}
\end{equation}
Here, $\bullet$ denotes the input of the sub-transformation which corresponds to the output of the next inner sub-transformation.
Each sub-transformation in the composition \cref{eq:nf_transformation_composition} is defined as
\begin{equation}\label{eq:nf_transformations}
\begin{aligned}
    \nfTransform^{(\nfIndTransforms)}: \mathbb{R}^{\numMaterialParams} &\to \mathbb{R}^{\numMaterialParams}, \\
    \nfIntermediateSamples^{(k-1)} &\mapsto 
    \nfTransform^{(\nfIndTransforms)}(\nfIntermediateSamples^{(k-1)};\nfParams^{(\nfIndTransforms)}) = \nfIntermediateSamples^{(k)}
    , \quad \nfIndTransforms = 1, \dots, \nfNumTransforms + 1,
\end{aligned}
\end{equation}
where we assume that $\nfIntermediateSamples^{(0)} = \nfBaseSamples$ and $\nfIntermediateSamples^{(\nfNumTransforms + 1)} = \materialParams$, respectively. Finally, the parameters of all $\nfTransform^{(\nfIndTransforms)}$ can be combined in $\nfParams = \bigl\{\nfParams^{(\nfIndTransforms)} \bigr\}_{1 \leq \nfIndTransforms \leq \nfNumTransforms+1}$. \\

In this paper, we use \acp{IAF} \cite{kingma_inverseAutoregressiveFlow_2016} to estimate the joint probability density $\materialParamsDist$. 
In \acp{IAF}, the transformations $\nfTransform^{(\nfIndTransforms)}$ in \cref{eq:nf_transformations} are based on (inverse) autoregressive transformations defined as 
\begin{equation}\label{eq:nf_iaf_transformation}
    \nfTransform^{(\nfIndTransforms)}(\nfIntermediateSamples^{(k-1)};\nfParams^{(\nfIndTransforms)}) 
    = \nfIntermediateSamples^{(k)}
    = \nfIAFLocation^{(k)}(\nfIntermediateSamples^{(k-1)};\nfParams^{(\nfIndTransforms)}) 
    + \nfIAFScaleConstrained^{(k)}(\nfIntermediateSamples^{(k-1)};\nfParams^{(\nfIndTransforms)})
    \odot \nfIntermediateSamples^{(k-1)},
\end{equation}
for each sub-transformation $\nfIndTransforms = 1, \dots, \nfNumTransforms.$ Here, $\nfIAFLocation^{(k)} \elm{\numMaterialParams}$ and $\nfIAFScaleConstrained^{(k)} \elm{\numMaterialParams}$, short for $\nfIAFLocation^{(k)}(\nfIntermediateSamples^{(k-1)};\nfParams^{(\nfIndTransforms)}), \ \nfIAFScaleConstrained^{(k)}(\nfIntermediateSamples^{(k-1)};\nfParams^{(\nfIndTransforms)})$, are the location and constrained scale vectors of the affine transformation where the latter is obtained from the unconstrained scale vector $\nfIAFScaleUnconstrained$ as $\nfIAFScaleConstrained = \operatorname{sigmoid}(\nfIAFScaleUnconstrained)$. In addition, $\odot$ denotes the elementwise Hadamard product. The $\nfIndTransforms$-th location and unconstrained scale vectors are functions of the output $\nfIntermediateSamples^{(k-1)}$ of the previous sub-transformation, parameterized in $\nfParams^{(\nfIndTransforms)}$ and implemented using \ac{MADE} networks \cite{germain_MADE_2015}. \Ac{MADE} networks with $\nfMADENumLayers - 1$ hidden layers have the general form
\begin{equation}\label{eq:nf_made}
    \nfMADEHiddenState^{(\nfMADEIndLayer)} = \nfMADEActivation \bigl( \nfMADEBiases^{(\nfMADEIndLayer)} + (\nfMADEMask^{(\nfMADEIndLayer)} \odot \nfMADEWeights^{(\nfMADEIndLayer)}) \nfMADEHiddenState^{(\nfMADEIndLayer - 1)} \bigr)
    , \quad \nfMADEIndLayer = 1, \dots, \nfMADENumLayers,
\end{equation}
where we assume that $\nfMADEHiddenState^{(0)} = \nfIntermediateSamples^{(k-1)}$, $\nfMADEHiddenState^{(L)} = [\nfIAFLocation^{(k)\transpose}, \nfIAFScaleUnconstrained^{(k)\transpose}]\transpose$ and $\nfMADEHiddenState^{(l)} \elm{\nfMADEHiddenStateSize}$ for $\nfMADEIndLayer = 1, \dots, \nfMADENumLayers-1$, respectively. Note that instead of two separate \ac{MADE} networks for the location vector and the unconstrained scale vector, we only use one network with double output size. The hyperparameter $\nfMADEHiddenStateSize \in \mathbb{N}$ controls the size of the $\nfMADEIndLayer$-th hidden layer. Furthermore, $\nfMADEActivation$ denotes an elementwise activation function and all weight matrices $\nfMADEWeights^{(\nfMADEIndLayer)}$ and bias vectors $\nfMADEBiases^{(\nfMADEIndLayer)}$ can be summarized in $\nfParams^{(\nfIndTransforms)} = \bigl\{\nfMADEWeights^{(\nfMADEIndLayer)}, \nfMADEBiases^{(\nfMADEIndLayer)} \bigl\}_{1 \leq \nfMADEIndLayer \leq \nfMADENumLayers}$. The binary matrices $\nfMADEMask^{(\nfMADEIndLayer)}$ with indices $u$ and $v$ are assembled as 
\begin{equation}\label{eq:nf_made_mask}
    \sca{M}_{u,v}^{(\nfMADEIndLayer)} = 
    \begin{cases}
        1, & \text{if } \nfMADEMaskFunc^{(\nfMADEIndLayer)}(u) \geq \nfMADEMaskFunc^{(\nfMADEIndLayer-1)}(v) \\
        0, & \text{otherwise}
    \end{cases},
\end{equation}
and ensure the autoregressive property of the \ac{IAF}. Here, the scalar-valued bijective functions $\nfMADEMaskFunc^{(\nfMADEIndLayer)}$ are defined as $\nfMADEMaskFunc^{(0)}(v)=v$ with $v \in \{1, \dots, \numMaterialParams\}$ and $\nfMADEMaskFunc^{(\nfMADENumLayers)}(u)=u-1$ with $u \in \{1, \dots, \numMaterialParams\}$ or assign a pre-set or random integer from the range $[1, \dots, \numMaterialParams-1]$ for all $\nfMADEMaskFunc^{(\nfMADEIndLayer)}$ with $\nfMADEIndLayer = 1, \dots, \nfMADENumLayers-1$. Since we use one \ac{MADE} network with double output size, the binary mask of the output layer is composed of two separate sub-masks, one for the location vector and one for the unconstrained scale vector.

The autoregressive property enables modeling the dependencies between the material parameters while estimating the joint density $\materialParamsDist$. Furthermore, under mild assumptions, it can be shown that autoregressive flows are universal approximators \cite{bogachev_triangularTransformationsOfMeasures_2005}. We choose \acp{IAF} in particular because they are fast to evaluate and scale well to high-dimensional distributions \cite{kingma_inverseAutoregressiveFlow_2016}. However, note that there are other flow-based models which include both additional autoregressive models, such as \acp{MAF} \cite{papamakarios_maskedAutoregressiveFlow_2017} or \ac{realNVP} flows \cite{dinh_realNVP_2017}, as well as non-autoregressive models, like Residual Flows \cite{chen_residualFLows_2019}. For more details on \acp{NF}, the reader is referred to \cite{papamakarios_normalizingFlows_2021}.

In order to ensure that the \ac{SEF} \cref{eq:discovered_reduced_model} parameterized by $\materialParams \sim \materialParamsDistNF$ remains physically admissible, we assume the parameters to be non-negative. To enforce the non-negativity constraint numerically, we use an exponential function as the last sub-transformation, which is defined as
\begin{equation}\label{eq:nf_nonnegativity_constraint}
    \nfTransform^{(\nfNumTransforms+1)}(\nfIntermediateSamples^{(\nfNumTransforms)};\nfParams^{(\nfNumTransforms+1)}) 
    = \operatorname{exp}(\nfIntermediateSamples^{(\nfNumTransforms)})
    = \materialParams,
\end{equation}
where $\operatorname{exp}$ is applied element-wise. Note that the exponential function is invertible and both the exponential and its inverse function are differentiable, so that the overall transformation $\nfTransform$ is still a bijective transport map.

To fit the flow-based model to the target distribution, which we refer to as distillation of the distribution over the material parameters, the \ac{NF} parameters $\nfParams$ must be optimized. For this purpose, a divergence or distance between the target distribution and the distribution estimated by the \ac{NF} is generally minimized. In our case, the \ac{NF} parameters $\nfParams$ are optimized so that they minimize the Wasserstein-1 distance between the \ac{GP} posterior and the distribution of stress-deformation functions induced by $\materialParamsDistNF$, as we will elaborate in the following subsection.

\subsubsection{Wasserstein-1 distance minimization}\label{subsubsec:wasserstein_minimization}

We distill the parameter distribution $\materialParamsDistNF$ by minimizing the Wasserstein-1 distance between the target distribution $\funcDistGP$ defined in \cref{eq:sampled_functions_gp} and our statistical model $\funcDistModelNF$. As can be seen from \cref{eq:parameterized_functional_dustribution}, the distribution $\funcDistModelNF$ is induced by the parameter distribution $\materialParamsDistNF$. Furthermore, the mapping $\parametersToStateMap: \materialParams \mapsto \vec{g} \elm{\numFuncPoints}$ in \cref{eq:parameterized_functional_dustribution} is deterministically defined by the stress-deformation relation \cref{eq:first_PK_from_SEF_incompressible} and the generalized ansatz for the \ac{SEF} in \cref{eq:ansatz_sef_general}. 

Given a realization of the material parameters $\materialParamsRealization$ from their distribution $\materialParamsDistNF$, we can calculate the discretized stress-deformation function  $\func$ induced by $\materialParamsRealization$ as follows
\begin{equation}\label{eq:sampled_functions_model}
\parametersToStateMap(\materialParamsRealization) =
    \begin{bmatrix*}[c]
        \sca{P}^{(1,1,1)}(\materialParamsRealization)\\
        \scriptstyle\vdots\\  
        \sca{P}^{(1,\numObservedPTestSpecific{1},\numFuncPointsTestSpecific{1})}(\materialParamsRealization)\\
        \vdots\\
        \sca{P}^{(\numTests,1,1)}(\materialParamsRealization)\\
        \scriptstyle\vdots\\  
        \sca{P}^{(\numTests,\numObservedPTestSpecific{\numTests},\numFuncPointsTestSpecific{\numTests})}(\materialParamsRealization)\\
    \end{bmatrix*}
    \elm{\numFuncPoints},
\end{equation}
where $\modelOutputGeneral(\materialParamsRealization)$ is, according to \cref{eq:first_PK_from_SEF_incompressible}, calculated as
\begin{equation}\label{eq:model_output}
    \modelOutputGeneral(\materialParamsRealization) = 
    \observationMapSpecific{\indTests}{\indObservedP} \Biggl(
        \frac{\partial \widebarsca{W}(\ten{F}^{(\indTests, \indFuncPoints)}; \materialParamsRealization)}{\partial \ten{F}} - \sca{p} \ten{F}^{(\indTests, \indFuncPoints)\minusTranspose}
    \Biggr).
\end{equation}
Here, $\observationMapGeneral$ denotes the observation map as defined in \cref{subsec:dataset_and_notation} and $\ten{F}^{(\indTests, \indFuncPoints)}$ are the $s$-th deformation gradients for the $t$-th test which are identical to those at which $\funcGP$ is sampled. Please note that the statistical model $\funcDistModelNF$ will never generate function samples that violate physics as long as the model library \cref{eq:ansatz_sef_general} is compatible with the principles of continuum mechanics, see \cref{subsec:hyperelastic_constitutive_modeling}. 

Our ultimate goal is to find the distribution over the material parameters $\materialParams \sim \materialParamsDistNF$ that induces a distribution of discretized functions $\funcDistModelNF$ whose Wasserstein-1 distance to $\funcDistGP$ is minimal. In the following, we frame the problem of matching the two probability distributions as an optimization problem in terms of the parameters $\nfParams$. \\

Here, we employ the dual form of the Wasserstein-1 distance according to the Kantorovich-Rubinstein theorem, which reads
\begin{equation}\label{eq:w1_distance_dualform}
    \mathcal{W}_{1}(\funcDistGPMeasure, \funcDistModelMeasure)
    = \supop{\norm{\phi}_{L} \leq 1}
    \expectation{\funcDistGPMeasure}\bigl( \phi(\funcGP) \bigr)
    - \expectation{\funcDistModelMeasure}\bigl( \phi(\funcModel) \bigr),
\end{equation}
where for a density $p$ on $\mathbb{R}^{\numFuncPoints}$, the expectation operator is defined as
\begin{equation}
    \expectation{p}\bigl( \phi(\func) \bigr) = \int_{\mathbb{R}^{\numFuncPoints}}
    \phi(\func)
    p(\func)
    \mathrm{d}\func.
\end{equation}
The supremum in \cref{eq:w1_distance_dualform} is taken over all Lipschitz-1 functions $\phi$. 
In fact, the dual form leads to a functional maximization over $\phi$ on the difference of two expectations of $\phi$ with respect to $\funcDistGPMeasure$ and $ \funcDistModelMeasure$. For a derivation of the above dual form of the Wasserstein-1 distance, the reader is referred to \cite{villani_optimalTransport_2009}.

In order to optimize the parameters $\nfParams$, we consider the optimization problem 
\begin{equation}\label{eq:w1_distance_optimization}
    \{\nfParamsOptimal, \lipschitzParamsOptimal \} = \argmin_{\nfParams} \argmax_{\lipschitzParams}
        \bigl[
            \expectation{\funcDistGPMeasure}\bigl(\wassersteinLipschitzNetwork(\funcGP; \lipschitzParams)\bigr) - \expectation{\funcDistModelMeasure(\cdot;\nfParams)}\bigl(\wassersteinLipschitzNetwork(\funcModel; \lipschitzParams)\bigr)
        \bigr],
\end{equation}
where $\wassersteinLipschitzNetwork(\func; \lipschitzParams): \mathbb{R}^{\numFuncPoints} \to \mathbb{R}$ is a Lipschitz-1 continuous function implemented as a feedforward \ac{NN} that is parameterized in $\lipschitzParams$ \cite{goodfellow_GANs_2014,arjovsky_wassersteinGANs_2017}. The Lipschitz-1 continuity constraint is enforced by a gradient penalty as proposed by \cite{gulrajani_improvedWassersteinGANs_2017}.

The optimization problem defined in \cref{eq:w1_distance_optimization} yields a minimax problem which is solved alternately for the optimal parameters $\nfParamsOptimal$ and $\lipschitzParamsOptimal$. While one of the two parameter sets is optimized, the other set is kept constant and denoted as $\nfParamsConstant$ and $\lipschitzParamsConstant$, respectively. First, we start with maximizing the loss $\lossLipschitz(\nfParamsConstant,\lipschitzParams)$ with respect to $\lipschitzParams$ for $\numItersLipschitz$ iterations, where $\lossLipschitz(\nfParamsConstant,\lipschitzParams)$ is defined as
\begin{equation}\label{eq:w1_distance_optimization_lipschitz}
    \lossLipschitz(\nfParamsConstant,\lipschitzParams) = 
    \lossWasserstein(\nfParamsConstant,\lipschitzParams)
    + \lipschitzPenaltyCoefficient \expectation{\lipschitzCombinedFuncDist}  \biggl[ 
        \Bigl( 
            \norm{\nabla_{\lipschitzCombinedFunc} \wassersteinLipschitzNetwork(\lipschitzCombinedFunc;\lipschitzParams)} - 1
        \Bigr)^{2}
    \biggr].
\end{equation}
Here, $\lossWasserstein(\nfParamsConstant,\lipschitzParams)$ is the loss for the Wasserstein-1 distance defined below and the second part is the Lipschitz-1 gradient penalty. The hyperparameter $\lipschitzPenaltyCoefficient$ controls the weight of the penalty term. The function $\lipschitzCombinedFunc \sim \lipschitzCombinedFuncDist(\lipschitzCombinedFunc)$ is defined as $\lipschitzCombinedFunc = \alpha \funcModel + (1 - \alpha)\funcGP$ where $\alpha$ is uniformly distributed in the interval $[0,1]$. In addition, $\nabla_{\lipschitzCombinedFunc}\wassersteinLipschitzNetwork(\lipschitzCombinedFunc;\lipschitzParams)$ is the gradient of the \ac{NN} with respect to its input $\lipschitzCombinedFunc$.
Second, we minimize the loss $\lossWasserstein(\nfParams,\lipschitzParamsConstant)$ with respect to the parameters $\nfParams$ in one iteration. The loss $\lossWasserstein(\nfParams,\lipschitzParamsConstant)$ is defined as follows
\begin{equation}\label{eq:w1_distance_optimization_wasserstein}
    \lossWasserstein(\nfParams,\lipschitzParamsConstant) = \expectation{\funcDistGPMeasure}\bigl(\wassersteinLipschitzNetwork(\funcGP; \lipschitzParamsConstant)\bigr) - \expectation{\funcDistModelMeasure(\cdot;\nfParams)}\bigl(\wassersteinLipschitzNetwork(\funcModel; \lipschitzParamsConstant)\bigr).
\end{equation}
We repeat this optimization procedure for a total of $\numItersWasserstein$ iterations. In both \cref{eq:w1_distance_optimization_lipschitz} and \cref{eq:w1_distance_optimization_wasserstein}, we follow \cite{tran_functionalPriorForBNN_2022} and estimate the expectation operators by Monte Carlo sampling. In order to sample from $\funcDistModelNF$, we first sample a random material parameter from $\materialParamsDistNF$ and then calculate the stress-deformation function induced by these material parameters according to \cref{eq:sampled_functions_model}. Furthermore, we use the spectral norm \cite{miyato_spectralNorm_2018} for the hidden layers of the Lipschitz-1 \ac{NN}, as we observed that it accelerates convergence and stabilizes the training dynamics. For more details on the Wasserstein-1 distance optimization, see \cite{tran_functionalPriorForBNN_2022, gulrajani_improvedWassersteinGANs_2017}. \\

\noindent \textbf{Remark:} Finally, we would like to make three important remarks: First, since the statistical model is constrained to physically admissible functions by construction but the  \ac{GP} posterior is not, these two distributions cannot be matched exactly. This means that the Wasserstein-1 distance between the two distributions generally converges toward a positive value greater than zero during the optimization process described above. However, this mismatch is intended as we aim to distill a distribution of physically admissible stress-deformation functions from a distribution over functions that may not be physically admissible. Second, we need to distinguish between the physical consistency of the distilled model and the validity of the quantified uncertainty. In principle, the discovered model is physically consistent by construction, but there is no guarantee that the quantified uncertainty is valid. Therefore, we propose to assess the validity of the quantified uncertainty based on coverage tests, as detailed in \cref{sec:appendix_ci_and_estimated_coverage}. Third, since the \ac{GP} posterior models epistemic uncertainty and represents the target distribution, the uncertainties in the material parameters are also epistemic.

\subsubsection{Sobol' sensitivity analysis and model refinement}\label{subsubsec:sensitivity_analysis}

We analyze the sensitivity of the statistical model \cref{eq:discovered_reduced_model} with respect to the material parameters $\materialParams \sim \materialParamsDistNF$ and remove all non-sensitive parameters from the joint distribution $\materialParamsDistNF$. To this end, we carry out a global, variance-based sensitivity analysis and consider the total-order Sobol' index \cite{sobol_globalSensitivityIndices_2001}. This sensitivity index measures the total effect of the parameters $\materialParamOne$ with $\indMaterialParam \in [1, \dots\allowbreak, \numMaterialParams]$ on the variance of the statistical model output and also takes the interactions with the other parameters into account. The total-order Sobol' index for the $\indFuncPoints$-th deformation gradient and the $\indObservedP$-th observed stress component in the $\indTests$-th mechanical test with respect to $\materialParamOne$ is defined as
\begin{equation}\label{eq:total_sobol_index}
    \totalSobolIndexGeneral(\materialParamOne) = 1 - \frac{
        \variance{\materialParams_{\sim\indMaterialParam}} \Bigl( 
            \expectation{\materialParamOne} \bigl( 
               \modelOutputGeneral(\materialParams) \mid \materialParams_{\sim\indMaterialParam}
            \bigr)
        \Bigr)
    }{\mathbb{V} \bigl(
        \modelOutputGeneral(\materialParams)
    \bigr)
    },
\end{equation}
with the stress component $\modelOutputGeneral(\materialParams)$ calculated according to \cref{eq:model_output}. In addition, $\mathbb{V}$ is the variance operator and $\expectation{\materialParamOne} (\modelOutputGeneral(\materialParams) \mid \materialParams_{\sim\indMaterialParam})$ is the conditional expectation. Here, $\materialParams_{\sim\indMaterialParam}$ includes all material parameters except $\materialParamOne$. In this contribution, we calculate the total-order Sobol' indices \cref{eq:total_sobol_index} using the Saltelli sampling method \cite{saltelli_sensitivityIndices_2002,saltelli_sensitivityAnalysis_2010}. As sampling bounds, we use the minimum and maximum values of each parameter which we estimate from the distribution $\materialParamsDistNF$.

In order to obtain a global sensitivity measure for the material parameters in all tests $\indTests = \{1, \dots, \numTests\}$, all $\numObservedPTesti$ observed stress components, and $\numFuncPointsTesti$ deformation gradients, we propose averaging the total-order Sobol' index as follows
\begin{equation}\label{eq:total_sobol_index_avergared}
    \totalSobolIndexAveraged(\materialParamOne) = 
    \frac{1}{\numTests} \sum_{\indTests=1}^{\numTests}
    \frac{1}{\numObservedPTesti\numFuncPointsTesti} \sum_{\indObservedP=1}^{\numObservedPTesti}
    \sum_{\indFuncPoints=1}^{\numFuncPointsTesti}
    \totalSobolIndexGeneral(\materialParamOne).
\end{equation}
In this paper, we assign the same weight to each test and each stress component. In principle, it would also be possible to give certain tests a greater weight.

After the averaged total-order Sobol' sensitivities have been calculated for all material parameters $\materialParams$, we remove those parameters from the joint distribution that fall below a specified threshold, and are therefore considered non-sensitive. Finally, we again optimize the Wasserstein-1 distance with the remaining model terms $\{\modelTerm^{(\indModelTerms)} \}_{\indModelTerms=1}^{\numReducedModelTerms}$ and parameters $\reducedMaterialParams \elm{\numReducedMaterialParams}$ to eliminate potential dependencies of the selected parameters on the removed ones. By removing the non-sensitive parameters, we induce sparsity and favor interpretability.

We admit that one weakness of Sobol' sensitivity analysis is that it does not provide a clear distinction between relevant and non-relevant parameters. Instead, a threshold for the total-order Sobol' index must be selected depending on the use case. Parameters whose total-order Sobol' index falls below that threshold are considered non-relevant. However, a strong reason for the Sobol' sensitivity analysis is that it provides further information on the model selection process as we show in our numerical tests. The threshold for the total-order Sobol' index is set to \num{1e-4} which leads to a reasonable trade-off between the sparsity of the discovered statistical model and its ability to quantify uncertainty for all numerical tests. For more details on the threshold selection, the reader is referred to \cref{sec:appendix_hyperparameters}.


\section{Results}\label{sec:results}

In this section, we demonstrate the proposed framework for experimental datasets with incompressible hyperelastic materials. We start with an isotropic rubber material in \cref{subsec:isotropic_data} and then consider an anisotropic human cardiac tissue in \cref{subsec:anisotropic_data}. In each test case, we study the sensitivity of the individual model terms. Sensitivities are used to select the relevant terms from the model library while promoting sparsity.
Throughout this paper, we quantify the uncertainty in the stress-deformation functions using centered \SI{95}{\percent}-intervals and validate the uncertainty based on the coverage of the synthetic data and the estimated coverage of the experimental data. For further details on these criteria, see \cref{sec:appendix_ci_and_estimated_coverage}. The hyperparameters of the \ac{GP} prior, \ac{NF}, Wasserstein-1 minimization and the Sobol' sensitivity analysis are specified in \cref{sec:appendix_hyperparameters} for all numerical test cases. Therein, we also present the considerations on which we base the hyperparameter selection and qualitatively discuss  the influence of the hyperparameters on runtime. In order to take into account the effect of the random initialization of the \acp{NN} used in the \ac{NF} and as Lipschitz-1 function, we tested five random initializations for each numerical test case. The results of this sensitivity analysis are summarized in \cref{sec:appendix_random_initialization}. In the following, we present representative results for each numerical test case.

\subsection{Isotropic experimental data}\label{subsec:isotropic_data}

First, we consider the Treloar dataset \cite{treloar_vulcanisedRubberData_1944}. This dataset comprises one \ac{UT}, \ac{EBT} and \ac{PS} test of isotropic incompressible vulcanized rubber material with $\numData^{(\mathrm{UT})} = \num{25}$, $\numData^{(\mathrm{EBT})} =\num{14}$ and $\numData^{(\mathrm{PS})} = \num{14}$ data points. In all $\numTests = \num{3}$ mechanical tests, only the first principal stress component $\sca{P}_{11}$ is measured, such that the number of measured scalar-valued stress-deformation functions is $\numFuncs = 3$. Given the controlled stretches for the specific tests, i.e., $\sca{\lambda}^{\mathrm{UT}}_{1}=\sca{\lambda}^{\mathrm{EBT}}_{1}=\sca{\lambda}^{\mathrm{EBT}}_{2}=\sca{\lambda}^{\mathrm{PS}}_{1}=\sca{\lambda}$, the remaining stretch components are obtained under consideration of the geometry, isotropy and incompressibility conditions: $\sca{\lambda}^{\mathrm{UT}}_{2} = \sca{\lambda}^{\mathrm{UT}}_{3}= 1 / \sqrt{\sca{\lambda}}$, 
$\sca{\lambda}^{\mathrm{EBT}}_{3} = 1 /\sca{\lambda^2}$, $\sca{\lambda}^{\mathrm{PS}}_{2}= 1, \sca{\lambda}^{\mathrm{PS}}_{3}= 1 /\sca{\lambda}$. In the following, our aim is to estimate a joint distribution over the material parameters that correspond to a physically admissible and interpretable statistical constitutive model. \\

We start with the formulation of a general model library for the isochoric \ac{SEF} of isotropic hyperelastic materials. Our model library combines both generalized \ac{MR} \cite{rivlin_MooneyRivlinModel_1947} and Ogden \cite{ogden_OgdenModel_1972} features and yields
\begin{equation}\label{eq:ansatz_sef_isotropic}
    \widebarsca{W}\bigl(\ten{F}; \materialParams\bigr) =
    \sum_{\substack{
        \indModelTermsMROne, \indModelTermsMRTwo \geq 0 \\
        1 \leq \indModelTermsMROne + \indModelTermsMRTwo \leq \numModelTermsMR
    }}
    \materialParamLinear^{(\indModelTermsMROne, \indModelTermsMRTwo)} 
    \modelTermMR^{(\indModelTermsMROne, \indModelTermsMRTwo)} \bigl(\ten{F}\bigr)
    + \sum_{\indModelTermsOgden = 1}^{\numModelTermsOgden}
    \materialParamLinear^{(\indModelTermsOgden)} 
    \modelTermOgden^{(\indModelTermsOgden)} \bigl(\ten{F}\bigr).
\end{equation}
Here, $\numModelTermsMR$ is the degree of the generalized \ac{MR} model, i.e., the maximum degree of the polynomial, and $\numModelTermsOgden$ denotes the number of Ogden terms. Note that under the assumption of isotropy, the general model library \cref{eq:ansatz_sef_general} is simplified since we can neglect the structural tensors $\{\structuralTensorGeneral\}_{\indStructuralTensors=1}^{\numStructuralTensors}$. Furthermore, the specific library \cref{eq:ansatz_sef_isotropic} does not contain any inner-non-linear parameters $\materialParamNonlinear^{(\indModelTerms)}$.
The terms of the generalized \ac{MR} model are defined by the first and second invariants of the right Cauchy-Green tensor $\cauchyTensor = \ten{F}\transpose \ten{F}$ which are calculated as
\begin{equation}\label{eq:invariants_isotropic}
    \begin{aligned}
        \invariantOne &= \tr{\cauchyTensor}, \\
        \invariantTwo &= \half \Bigl( \bigl( \tr{\cauchyTensor} \bigr)^2 - \tr{\cauchyTensor^2}  \Bigr).
    \end{aligned}
\end{equation}
The terms of the generalized Ogden model are defined through the principal stretches $\principalStretch{1}$, $\principalStretch{2}$ and $\principalStretch{3}$. Ultimately, the explicit form of the generalized \ac{MR} and Ogden terms is
\begin{equation}\label{eq:model_terms_isotropic}
    \begin{aligned}
        \modelTermMR^{(\indModelTermsMROne, \indModelTermsMRTwo)}(\invariantOne, \invariantTwo) 
        &= (\invariantOne - 3)^{\indModelTermsMROne} (\invariantTwo - 3)^{\indModelTermsMRTwo}, \\
        \modelTermOgden^{(\indModelTermsOgden)}(\principalStretch{1}, \principalStretch{2}, \principalStretch{3}) &= \principalStretch{1}^{\exponentOgdenGeneral} 
        + \principalStretch{2}^{\exponentOgdenGeneral} 
        + \principalStretch{3}^{\exponentOgdenGeneral} 
        - 3,
    \end{aligned}
\end{equation}
where $\exponentOgdenGeneral$ is the exponent of the Ogden term $\modelTermOgden^{(\indModelTermsOgden)}$.

For our numerical test, we set $\numModelTermsMR = \num{3}$ and $\numModelTermsOgden=8$ with fixed Ogden exponents $\exponentOgdenGeneral \in \{-5, -4, -3, -1, 1, 3, 4, 5\}$. We omit the Ogden terms with exponents $\exponentOgdenGeneral = -2$ and $\exponentOgdenGeneral = 2$ since they correspond to the \ac{MR} terms $\modelTermMR^{(0, 1)}$ and $\modelTermMR^{(1, 0)}$, respectively \cite{holzapfel_nonlinearSolidMechanics_2000}. The vector of material parameters is 
\begin{equation}
    \materialParams = [\materialParamLinear^{(0,1)}, \materialParamLinear^{(1,0)}, \dots, \materialParamLinear^{(\numModelTermsMR, 0)}, \materialParamLinear^{(1)}, \dots, \materialParamLinear^{(\numModelTermsOgden)}]\transpose \quad \in\mathbb{R}^{\numMaterialParams},
\end{equation} 
with $\numMaterialParams=17$. When optimizing the \ac{GP} hyperparameters and inferring the \ac{GP} posterior, we assume that the minimum and relative error standard deviation are $\minErrorStddev = \SI{0.01}{\kilo\pascal}$ and $\relativeErrorStddev = \SI{5}{\percent}$, respectively. Ultimately, the runtime for distilling the distribution of material parameters on a NVIDIA  \ac{GPU} A100 is approximately $\num{3.5}$ hours. In the remainder of this subsection, we report and discuss the results of this numerical test.\\ 

\begin{figure}[htb]
    \centering
    \includegraphics[width=\linewidth]{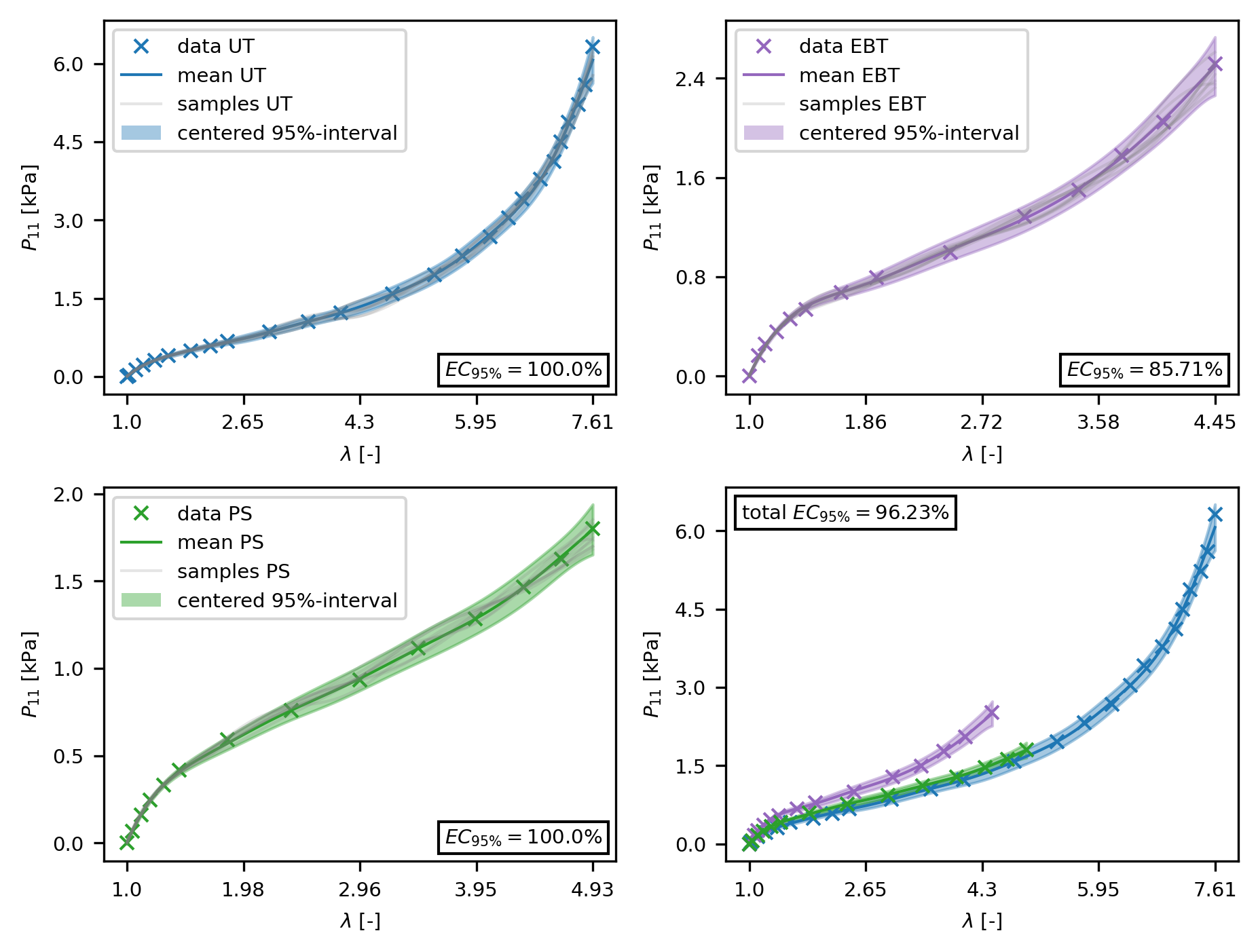}       
    \caption{\textbf{Treloar test case}: \Ac{GP} posterior. The illustrations show the \ac{GP} posterior mean, the centered \SI{95}{\percent}-intervals, some random stress-deformation function samples and the estimated coverages for the \ac{UT}, \ac{EBT} and \ac{PS} test as well as the total estimated coverage. The \ac{GP} posterior is used for data augmentation in the subsequent steps of the proposed framework.}
    \label{fig:gp_isotropic}
\end{figure}

\noindent \textbf{GP posterior:} The \ac{GP} posterior for all mechanical tests included in the Treloar dataset is illustrated in \cref{fig:gp_isotropic}. The total estimated coverage of the centered \SI{95}{\percent}-interval of the \ac{GP} posterior reported in \cref{fig:gp_isotropic} is $\estimatedCoverage=\SI{96.23}{\percent}$, which is very close to $\SI{95}{\percent}$, implying that the uncertainty in the \ac{GP} posterior is well estimated.\\

\begin{figure}[htb]
    \centering
    \includegraphics[width=\linewidth]{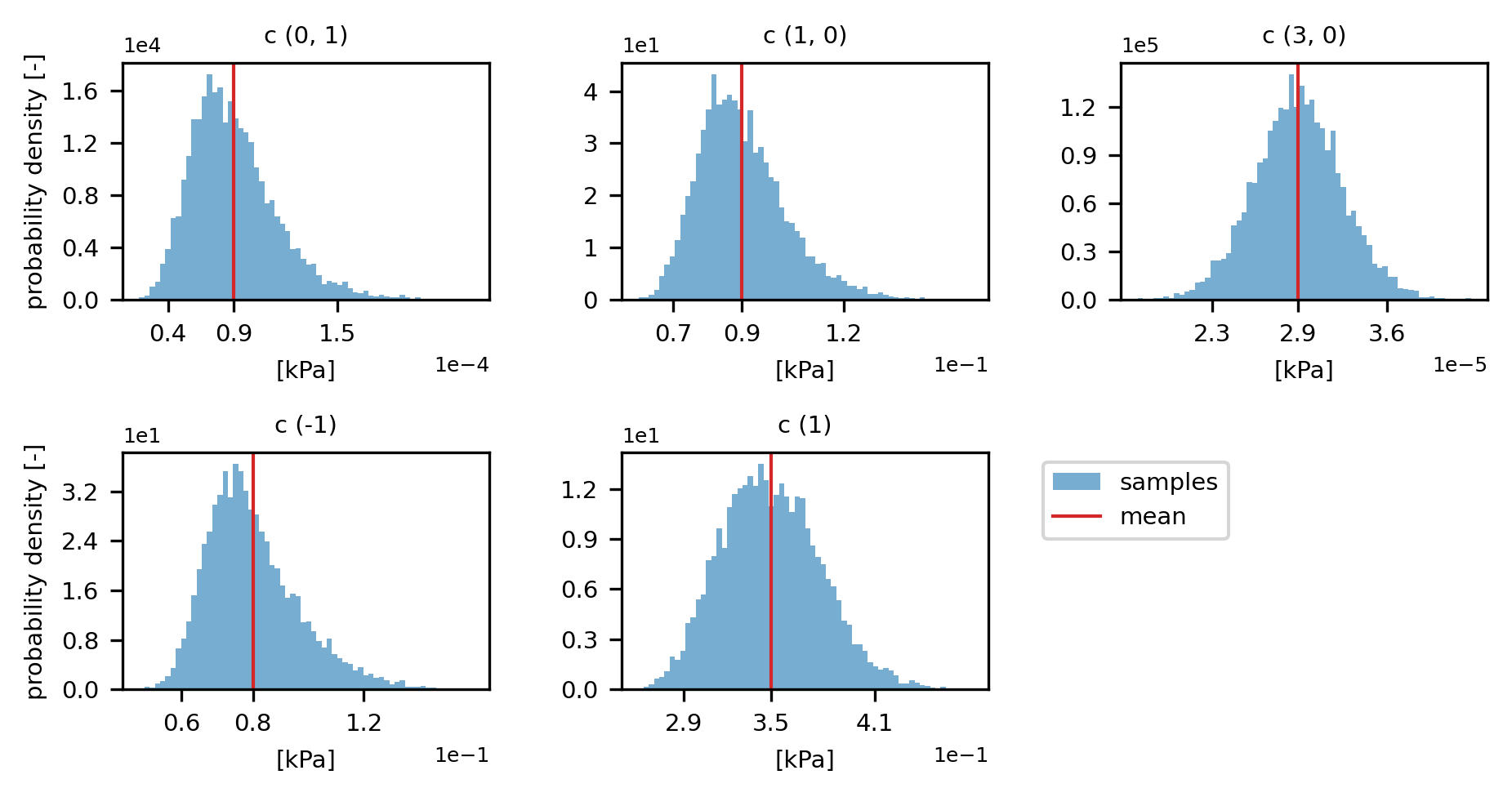}       
    \caption{\textbf{Treloar test case}: Distilled distribution of the material parameters after the sensitivity analysis and model refinement. The distribution over stress-deformation functions induced by this parameter distribution is shown in \cref{fig:model_isotropic}.}
    \label{fig:parameters_isotropic}
\end{figure}

\noindent \textbf{Discovered model:} As a result of our statistical model discovery framework, we ultimately obtain the distribution over the material parameters $\reducedMaterialParams$ shown in \cref{fig:parameters_isotropic}. The associated, discovered \ac{SEF} has the following form
\begin{equation}\label{eq:result_model_isotropic}
    \begin{aligned}
        \widebarsca{W}\bigl(\ten{F}; \reducedMaterialParams\bigr) = \;
        &\materialParamLinear^{(0, 1)} (\invariantTwo - 3)
        + \materialParamLinear^{(1, 0)} (\invariantOne - 3)
        + \materialParamLinear^{(3, 0)} (\invariantOne - 3)^{3} \\
        &+ \materialParamLinear^{(-1)} (\principalStretch{1}^{-1} + \principalStretch{2}^{-1} + \principalStretch{3}^{-1} - 3)
        + \materialParamLinear^{(1)} (\principalStretch{1}^{1} + \principalStretch{2}^{1} + \principalStretch{3}^{1} - 3),
    \end{aligned}
\end{equation}
where the distribution of the reduced material parameters is approximated by the \ac{NF} $\reducedMaterialParamsDistNFOptimized$, i.e., $\reducedMaterialParams \sim \reducedMaterialParamsDistNFOptimized$. For better assignment of the parameters to the terms, we do not consecutively number the Ogden parameters, but we name each parameter $\materialParamLinear^{(\indModelTermsOgden)}$ in the form $\mathrm{\materialParamLinear}^{ (\exponentOgdenGeneral)}$. The sensitivity analysis with respect to the initialization of the involved \acp{NN} summarized in \cref{tab:appendix_initialization_isotropic} underscores that the model discovery converges robustly to a very similar model structure. For three initializations, the model reported in \cref{eq:result_model_isotropic} was discovered and for the other two initializations a four-term model without the term $\materialParamLinear^{(0, 1)} (\invariantTwo - 3)$ was distilled.

In the literature, the Treloar dataset is a frequently used benchmark test for incompressible hyperelastic constitutive models. Our discovered model \cref{eq:result_model_isotropic} is compatible with the literature and matches the structure of well established hyperelastic models, which are primarily constructed of first invariant-based polynomials and a few Ogden terms, see, e.g., \cite{holzapfel_nonlinearSolidMechanics_2000,steinmann_hyperelasticModelsTreloar_2012}. Compared to most of the constitutive models available in the literature, such as the models reported in \cite{steinmann_hyperelasticModelsTreloar_2012,ricker_systematicFittingHyperelasticity_2023,linka_newCANNs_2023,fuhg_extremeSparsificationModelDiscovery_2024}, our model is less complex, remains linear in the material parameters, and is therefore easier to interpret.

\begin{figure}[h!]
    \centering
    \includegraphics[width=\linewidth]{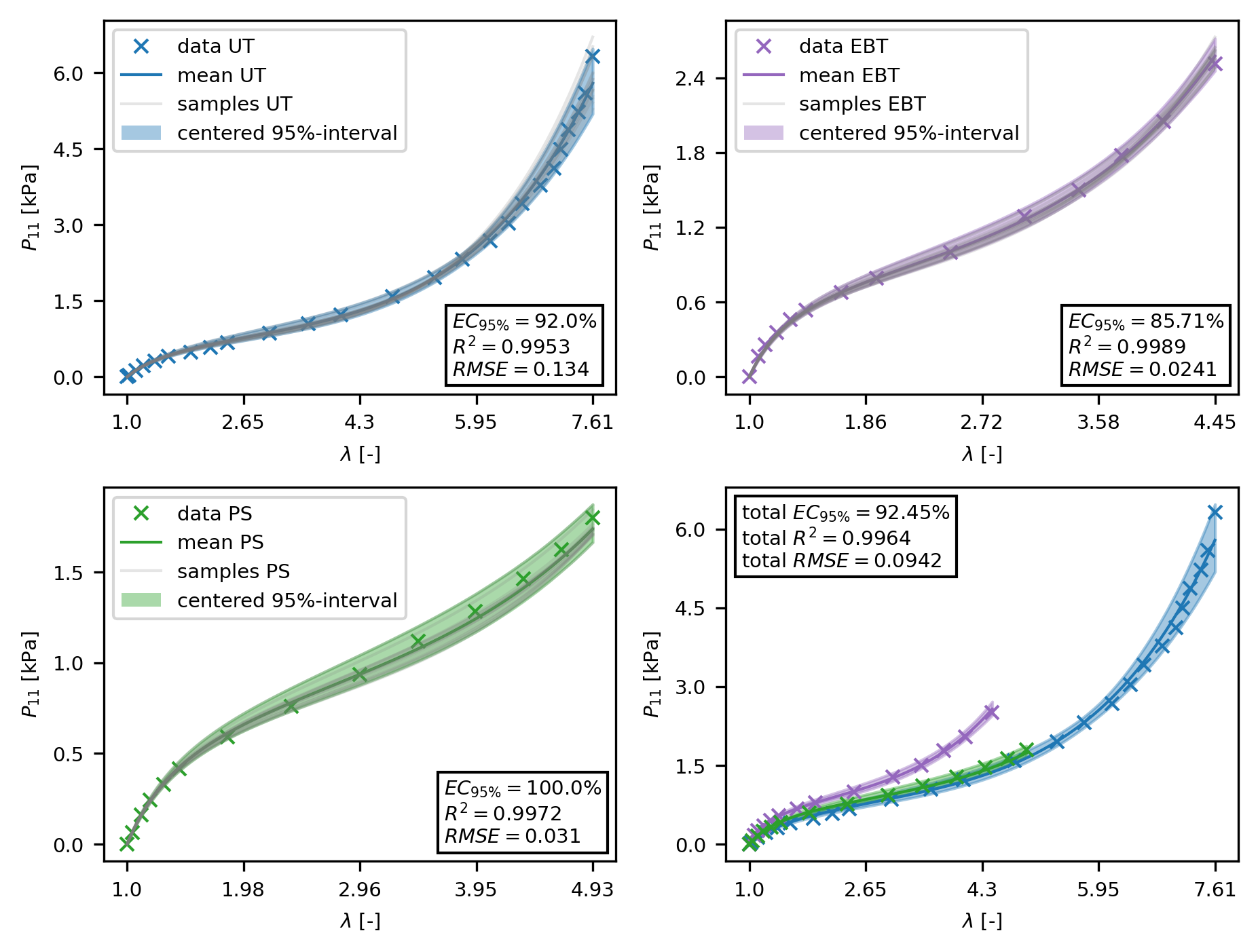}       
    \caption{\textbf{Treloar test case}: Distilled interpretable statistical model. The illustrations show the mean stress-deformation functions, the centered \SI{95}{\percent}-intervals, some random stress-deformation function samples as well as the individual and total validation metrics. The \ac{RMSE} and the \ac{R2} refer to the mean stress-deformation functions, respectively, and show a good fit. Additionally, the results of the coverage estimation prove that the uncertainty is well estimated.}
    \label{fig:model_isotropic}
\end{figure}

The distribution of stress-deformation functions is illustrated in \cref{fig:model_isotropic}. The high \ac{R2} of $\rSquare = \num{0.9964}$ and the low \ac{RMSE} of $\rmse = \num{0.0942}$ indicate a very good fit of the mean function. In addition, from the estimated coverage $\estimatedCoverage = \SI{92.45}{\percent}$, we can conclude that the uncertainty in the distribution of stress-deformation functions and thus the distribution of material parameters is well estimated. The estimated coverage and validation metrics \ac{R2} and \ac{RMSE} also show similarly good values for the other four initializations of the \acp{NN} that were tested. Only for one initialization, the results are slightly worse, but with $\estimatedCoverage = \SI{86.79}{\percent}$, $\rSquare = \num{0.9964}$ and $\rmse = \num{0.0947}$ still in an acceptable range. Nevertheless, especially the distributions for the \ac{UT} and \ac{EBT} tests exhibit some counterintuitive effects, see \cref{fig:model_isotropic}. In the range of medium deformation, the \SI{95}{\percent}-intervals first become larger and then narrow again, although we assume a heteroscedastic error model. However, the described effect also occurs, e.g., in \cite{wollner_BayesianFrameworkUncertaintyEstimationCalibration_2025} for a similar setup. Since the effect does not occur exclusively in our results, we suspect that it is not an artifact of our approach, but rather that it is an attribute of the model library or the discovered model. \\

\begin{figure}[htb]
    \centering
    \includegraphics[width=\linewidth]{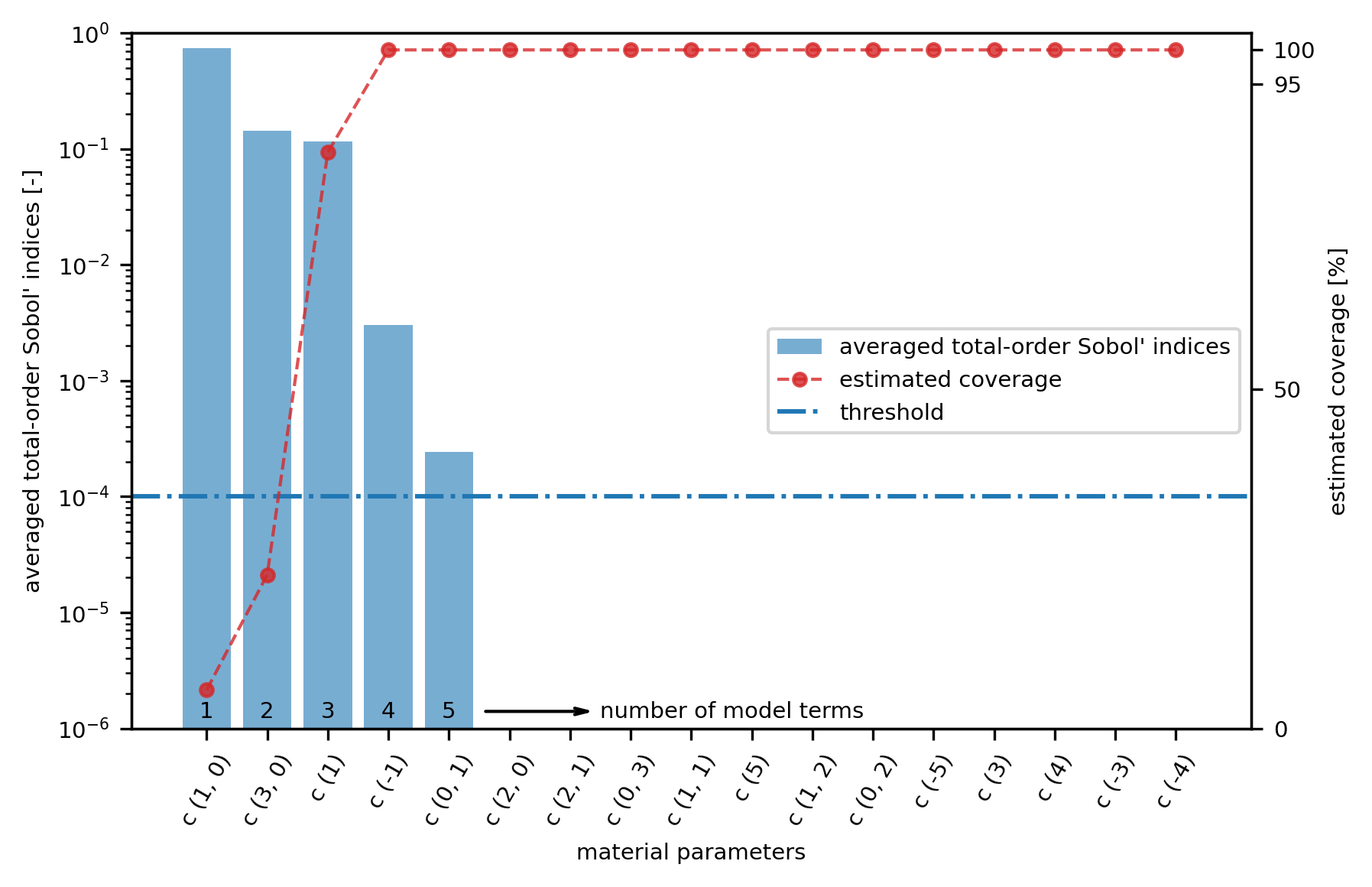}       
    \caption{\textbf{Treloar test case:} Illustration of the model selection process based on the sensitivity analysis. The bars show the total-order Sobol' indices for each material parameter averaged over all mechanical tests, observed stress components and discretization points in their descending order, calculated  according to \cref{eq:total_sobol_index_avergared}. The blue dashed-dotted line indicates the threshold of the total-order Sobol' index below which the material parameters are considered non-relevant. The numbers at the bottom of the bars indicate the total number of model terms when the model term associated with the corresponding material parameter is added to the discovered model. In addition, the red dotted line illustrates how the empirical coverage develops with increasing model size. Note that the estimated coverage may differ slightly after the refinement of the discovered model in the fourth step of the framework.}
    \label{fig:sensitivities_isotropic}
\end{figure}

\noindent \textbf{Sensitivity-based model selection:} For more information on the model selection process, we refer the reader to \cref{fig:sensitivities_isotropic}. This figure illustrates the averaged total-order Sobol' indices of all material parameters in the model library in their descending order and visualizes how coverage develops as the model size increases.

\begin{figure}[h!]
    \centering
    \includegraphics[width=\linewidth]{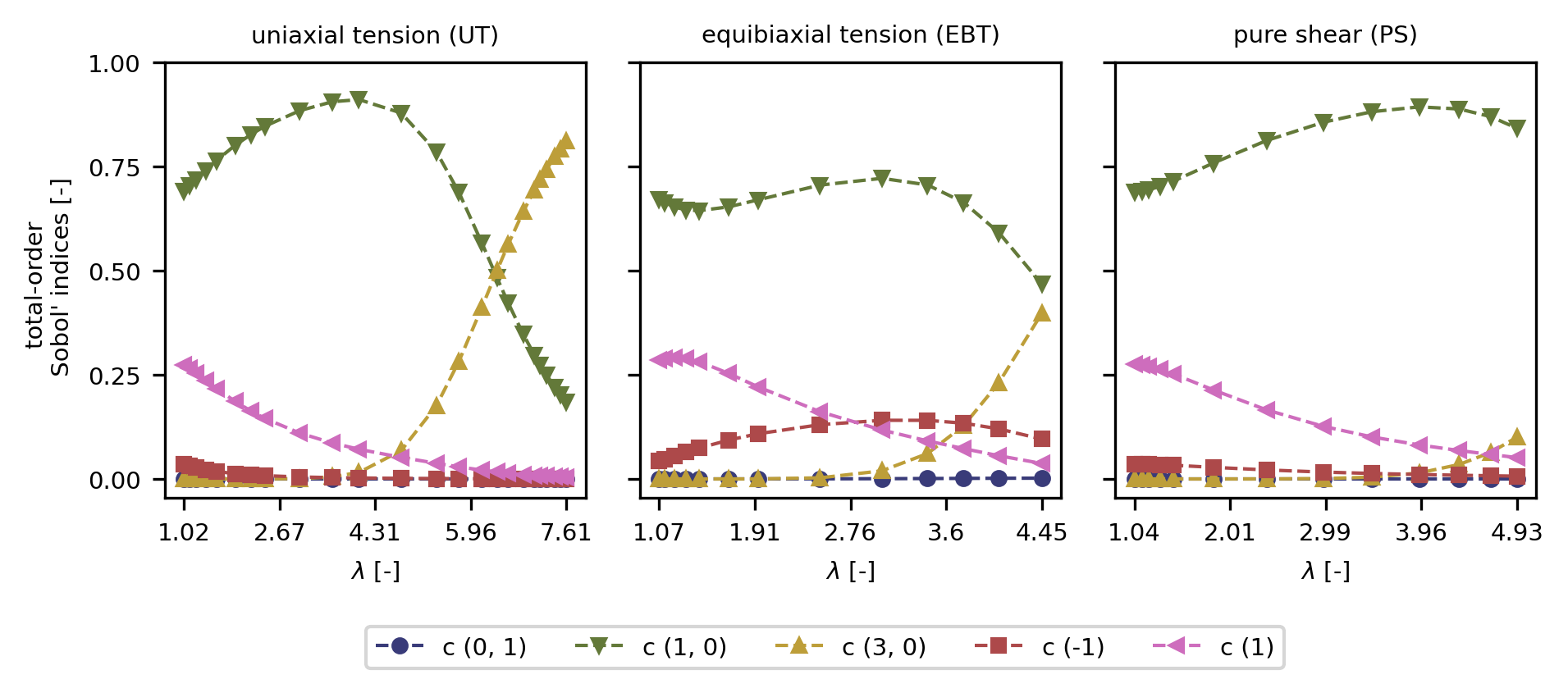}       
    \caption{\textbf{Treloar test case}: Development of total-order Sobol' indices over the course of the mechanical tests. The results show that the stress component $\sca{P}_{11}$ predicted by the statistical model is most sensitive to the terms linearly parameterized in $\materialParamLinear^{(1,0)}$ (Neo-Hookean term) and $\materialParamLinear^{(3,0)}$. However, the effect of the terms differs for the different deformation modes and usually changes with increasing deformation.}
    \label{fig:sensitivities_path_isotropic}
\end{figure}

\cref{fig:sensitivities_path_isotropic} illustrates the development of the total-order Sobol' indices as a measure of the sensitivity against the principal stretches. To the best of the authors' knowledge, deformation-dependent Sobol' indices analysis has not been previously used to analyze the individual contributions of the model terms to the overall stress response. Interestingly, the results of the analysis in \cref{fig:sensitivities_path_isotropic} reveal that each term of the model contributes differently depending on the type of mechanical test and the level of deformation. 
This could explain the diversity of the models proposed or discovered for the Treloar dataset in the past, see, e.g., \cite{steinmann_hyperelasticModelsTreloar_2012,ricker_systematicFittingHyperelasticity_2023,abdolazizi_CKANs_2025,urrea_automatedModelDiscovery_2025}. The results also give a hint of why some classical hyperelastic models fail to capture Treloar's \ac{UT}, \ac{EBT}, and \ac{PS} data with a single parameter set, see, e.g., \cite{steinmann_hyperelasticModelsTreloar_2012}.

The figure shows that the stress component $\sca{P}_{11}$ predicted by the discovered model is most sensitive with respect to the Neo-Hookean term $\materialParamLinear^{(1, 0)} (\invariantOne - 3)$. However, in the \ac{UT} and \ac{EBT} tests, the sensitivity of the Neo-Hookean term decreases significantly for larger deformations. This is consistent with the classical interpretation of rubber elasticity as an entropic response of Gaussian polymer chains. It explains why simple constitutive models, such as the Neo-Hookean model, are able to capture the initial slope in \ac{UT}, \ac{EBT}, and \ac{PS} tests \cite{treloar_physicsRubberElasticity_2005}. With increasing deformation, however, the Sobol' indices reveal a mode-dependent activation of higher-order terms: in \ac{UT} and \ac{EBT}, the \ac{MR} term $\materialParamLinear^{(3, 0)} (\invariantOne - 3)^{3}$ contributions become more dominant. This could reflect the transition to chain alignment and the finite extensibility effects that produce strain stiffening and curvature in the stress–deformation response \cite{arruda_constitutiveModelRubber_1993,tosaka_vulcanizedNaturalRubber_2010}. The deformation-dependent Sobol' indices provide a quantitative interpretation of these observations. They clearly indicate which constitutive contributions dominate the stress response at a given deformation level and deformation mode. It becomes visually apparent why calibration over limited deformation ranges or deformation modes promotes the inclusion or exclusion of specific model terms or invariants.

\subsection{Anisotropic data}\label{subsec:anisotropic_data}

In the following numerical tests, we consider data from anisotropic human cardiac tissue \cite{sommer_humanVentricularMyocardium_2015}. Compared to isotropic materials, the discovery of anisotropic constitutive models is substantially more complex because the mechanical responses of anisotropic materials are direction-dependent. For the underlying dataset, the three mutually orthogonal preferred directions are assumed to be $\normalDirectionVectorFiber$, $\normalDirectionVectorSheet$ and $\normalDirectionVectorNormal$ which are defined in the reference configuration according to \cite{martonova_modelDiscoveryCardiacTissue_2025} and correspond to the local fiber (f), sheet (s), and normal (n) directions, respectively. The directions are finally encoded in the structural tensors
\begin{equation}\label{eq:structural_tensors}
    \structuralTensorFiber = \normalDirectionVectorFiber \otimes \normalDirectionVectorFiber, \quad
    \structuralTensorSheet = \normalDirectionVectorSheet \otimes \normalDirectionVectorSheet, \quad
    \structuralTensorNormal = \normalDirectionVectorNormal \otimes \normalDirectionVectorNormal,
\end{equation}
where $\otimes$ is the tensor product. Considering the structural tensors defined in \cref{eq:structural_tensors} as input to the \ac{SEF}, the class of material symmetry is taken into account.

The dataset comprises the measured deformation and stress data from a total of six \ac{SS} and five \ac{BT} tests. In the six \ac{SS} tests, the respective shear strains increase from $\sca{\gamma}_{\mathrm{min}}=\num{0.0}$ to $\sca{\gamma}_{\mathrm{max}}=\num{0.5}$ and the principal stretches remain equal to $\num{1.0}$. Depending on the direction of the shear deformation, the associated shear stress component of the Cauchy stress tensor was measured. For example, in the test in which shear strain $\sca{\gamma}_{\mathrm{fs}}$ is controlled, Cauchy shear stress $\sca{\sigma}_{\mathrm{sf}}$ is measured.
In the five \ac{BT} tests, the relative stretch $\sca{\lambda}$ is increased from $\sca{\lambda}_{\mathrm{min}}=1.0$ to $\sca{\lambda}_{\mathrm{max}}=1.1$. The absolute stretches in the fiber and the normal direction are controlled by the ratio $\sca{\lambda}_{\mathrm{f}}^{*} \! : \! \sca{\lambda}_{\mathrm{n}}^{*}$ of the parameters $\sca{\lambda}_{\mathrm{f}}^{*}$ and $\sca{\lambda}_{\mathrm{n}}^{*}$ and are calculated as follows
\begin{equation}\label{eq:fiber_and_normal_stretch_ratios_anisotropic}
    \sca{\lambda}_{\mathrm{f}} = 1 + \sca{\lambda}_{\mathrm{f}}^{*} (\sca{\lambda} - 1), \quad \sca{\lambda}_{\mathrm{n}} = 1 + \sca{\lambda}_{\mathrm{n}}^{*} (\sca{\lambda} - 1). 
\end{equation}
For $\sca{\lambda}_{\mathrm{f}}^{*} \! : \! \sca{\lambda}_{\mathrm{n}}^{*}$, the ratios $1\!:\!1$, $1\!:\!0.75$, $0.75\!:\!1$, $1\!:\!0.5$ and $\!0.5:\!1$ are considered. In each test, the principal Cauchy stresses $\sca{\sigma}_{\mathrm{ff}}$ and $\sca{\sigma}_{\mathrm{nn}}$ were measured.
In total, the dataset thus includes $\numTests = \num{11}$ mechanical tests, but $\numFuncs = \num{16}$ measured scalar-valued stress-deformation functions as defined in \cref{subsec:dataset_and_notation}. For each of these stress-deformation functions, $\numData = \num{11}$ data points were measured.

The human cardiac tissue under consideration is assumed to be perfectly incompressible. Under this assumption, the Cauchy stress tensor is derived from the scalar-valued isochoric \ac{SEF} $\widebarsca{W}$ as follows
\begin{equation}\label{eq:cauchy_stress_from_SEF_incompressible}
    \ten{\sigma} 
    = \sca{J}^{-1} \ten{P} \ten{F}\transpose 
    = \frac{
        \partial \widebarsca{W}(
            \ten{F},
            \{\structuralTensorGeneral\}_{\indStructuralTensors=1}^{\numStructuralTensors}; \materialParams
        )
    }{
        \partial \ten{F}
    } \ten{F}\transpose - \sca{p} \ten{I}.
\end{equation}
Here, $\ten{I}$ denotes the identity matrix. The incompressibility constraint $\sca{J} = \operatorname{det}\ten{F} = 1$ is again enforced by a Lagrange multiplier, see \cref{eq:first_PK_from_SEF_incompressible}.

For our numerical tests, we adopt a modified version of the model library from \cite{martonova_modelDiscoveryCardiacTissue_2025}. In addition to the isotropic invariants defined in \cref{eq:invariants_isotropic}, this library also contains terms which are functions of the anisotropic fourth or eighth invariants formed by combining the Cauchy-Green tensor $\cauchyTensor$ with the structural tensors defined in \cref{eq:structural_tensors}. The stretch-related fourth invariants are
\begin{equation}\label{eq:invariants_anisotropic_fourth}
    \begin{aligned}
        \invariantFourF(\cauchyTensor, \structuralTensorFiber) &= \normalDirectionVectorFiber \cdot \cauchyTensor \normalDirectionVectorFiber, \\
        \invariantFourS(\cauchyTensor, \structuralTensorSheet) &= \normalDirectionVectorSheet \cdot \cauchyTensor \normalDirectionVectorSheet, \\
        \invariantFourN(\cauchyTensor, \structuralTensorNormal) &= \normalDirectionVectorNormal \cdot \cauchyTensor \normalDirectionVectorNormal.
    \end{aligned}
\end{equation}
The eighth invariants considering the coupling between directions are defined as
\begin{equation}\label{eq:invariants_anisotropic_eight}
    \begin{aligned}
        \invariantEightFS(\cauchyTensor, \structuralTensorFiber, \structuralTensorSheet) 
        &= \normalDirectionVectorFiber \cdot \cauchyTensor \normalDirectionVectorSheet, \\
        \invariantEightFN(\cauchyTensor, \structuralTensorFiber, \structuralTensorNormal) 
        &= \normalDirectionVectorFiber \cdot \cauchyTensor \normalDirectionVectorNormal, \\
        \invariantEightSN(\cauchyTensor, \structuralTensorSheet, \structuralTensorNormal) 
        &= \normalDirectionVectorSheet \cdot \cauchyTensor \normalDirectionVectorNormal.
    \end{aligned}
\end{equation}
Based on the invariants in \cref{eq:invariants_isotropic,eq:invariants_anisotropic_fourth,eq:invariants_anisotropic_eight}, the model library for the isochoric \ac{SEF} yields
\begin{equation}\label{eq:ansatz_sef_anisotropic}
    \begin{aligned}
        \widebarsca{W}\bigl( \ten{F}, \{\structuralTensorGeneral\}_{\indStructuralTensors=1}^{\numStructuralTensors}; \materialParams\bigr) &=
        \materialParamLinear^{(2,1)} \bigr[ \invariantOne - 3 \bigl]
        + \materialParamLinear^{(2,2)} \Bigl( 
            \operatorname{exp}\bigl( 
                \materialParamNonlinearSingle^{(1,2)} \bigl[ \invariantOne - 3\bigr]
            \bigr) - 1
        \Bigr) \\
        &+ \materialParamLinear^{(2,3)} \bigr[ \invariantOne - 3 \bigl]^{2}
        + \materialParamLinear^{(2,4)} \Bigl( 
            \operatorname{exp}\bigl( 
                \materialParamNonlinearSingle^{(1,4)} \bigr[ \invariantOne - 3 \bigl]^{2}
            \bigr) - 1
        \Bigr) \\
        &+ \materialParamLinear^{(2,5)} \bigr[ \invariantTwo - 3 \bigl]
        + \materialParamLinear^{(2,6)} \Bigl( 
            \operatorname{exp}\bigl( 
                \materialParamNonlinearSingle^{(1,6)} \bigl[ \invariantTwo - 3\bigr]
            \bigr) - 1
        \Bigr) \\
        &+ \materialParamLinear^{(2,7)} \bigr[ \invariantTwo - 3 \bigl]^{2}
        + \materialParamLinear^{(2,8)} \Bigl( 
            \operatorname{exp}\bigl( 
                \materialParamNonlinearSingle^{(1,8)} \bigr[ \invariantTwo - 3 \bigl]^{2}
            \bigr) - 1
        \Bigr) \\
        &+ \materialParamLinear^{(2,11)} \bigr[ \invariantFourFBar - 1 \bigl]^{2}
        + \materialParamLinear^{(2,12)} \Bigl( 
            \operatorname{exp}\bigl( 
            \materialParamNonlinearSingle^{(1,12)} \bigl[ \invariantFourFBar - 1\bigr]^{2}
            \bigr) - 1
        \Bigr) \\
        &+ \materialParamLinear^{(2,15)} \bigr[ \invariantFourSBar - 1 \bigl]^{2}
        + \materialParamLinear^{(2,16)} \Bigl( 
            \operatorname{exp}\bigl( 
            \materialParamNonlinearSingle^{(1,16)} \bigl[ \invariantFourSBar - 1\bigr]^{2}
            \bigr) - 1
        \Bigr) \\
        &+ \materialParamLinear^{(2,19)} \bigr[ \invariantFourNBar - 1 \bigl]^{2}
        + \materialParamLinear^{(2,20)} \Bigl( 
            \operatorname{exp}\bigl( 
            \materialParamNonlinearSingle^{(1,20)} \bigl[ \invariantFourNBar - 1\bigr]^{2}
            \bigr) - 1
        \Bigr) \\
        &+ \materialParamLinear^{(2,23)} \bigr[ \invariantEightFS \bigl]^{2}
        + \materialParamLinear^{(2,24)} \Bigl( 
            \operatorname{exp}\bigl( 
            \materialParamNonlinearSingle^{(1,24)} \bigl[ \invariantEightFS \bigr]^{2}
            \bigr) - 1
        \Bigr) \\
        &+ \materialParamLinear^{(2,27)} \bigr[ \invariantEightFN \bigl]^{2}
        + \materialParamLinear^{(2,28)} \Bigl( 
            \operatorname{exp}\bigl( 
            \materialParamNonlinearSingle^{(1,28)} \bigl[ \invariantEightFN \bigr]^{2}
            \bigr) - 1
        \Bigr) \\
        &+ \materialParamLinear^{(2,31)} \bigr[ \invariantEightSN \bigl]^{2}
        + \materialParamLinear^{(2,32)} \Bigl( 
            \operatorname{exp}\bigl( 
            \materialParamNonlinearSingle^{(1,32)} \bigl[ \invariantEightSN \bigr]^{2}
            \bigr) - 1
        \Bigr).
    \end{aligned}
\end{equation}
Here, $\invariantFourFBar = \operatorname{max}\{\invariantFourF, 1\}$, $\invariantFourSBar = \operatorname{max}\{\invariantFourS, 1\}$, and $\invariantFourNBar = \operatorname{max}\{\invariantFourN, 1\}$ such that the terms based on the fourth invariants are activated only for tensile stretches. In the literature, the \ac{SEF} in \cref{eq:ansatz_sef_anisotropic} is known as \ac{CANN} \cite{linka_newCANNs_2023}. Compared to the original \ac{SEF} proposed in \cite{martonova_modelDiscoveryCardiacTissue_2025}, in \cref{eq:ansatz_sef_anisotropic}, we remove the terms based on the corrected fourth invariants $\bigl[ \invariantFourFBar - 1\bigr]$, $\bigl[ \invariantFourSBar - 1\bigr]$ and $\bigl[ \invariantFourNBar - 1\bigr]$ and the terms based on the eighth invariants $\invariantEightFS$, $\invariantEightFN$ and $\invariantEightSN$, since these terms may induce stresses in deformation-free states. For a better comparison with the results in \cite{martonova_modelDiscoveryCardiacTissue_2025}, we stick to the numbering of material parameters from \cite{martonova_modelDiscoveryCardiacTissue_2025}. However, when naming the parameters, we distinguish between linear parameters $\materialParamLinear$ and non-linear parameters $\materialParamNonlinearSingle$. Ultimately, the model library in \cref{eq:ansatz_sef_anisotropic} has a total of $\numMaterialParams = \num{30}$ material parameters.

When optimizing the \ac{GP} hyperparameters and inferring the \ac{GP} posterior, we assume that the minimum and relative error standard deviation are $\minErrorStddev = \SI{0.01}{\kilo\pascal}$ and $\relativeErrorStddev = \SI{5}{\percent}$, respectively. The runtime for distilling the distribution of material parameters on a NVIDIA \ac{GPU} A100 is approximately $\num{6}$ hours for both synthetic (\cref{subsubsec:anisotropic_synthetic_data}) and experimental data (\cref{subsubsec:anisotropic_experimental_data}). \\

\subsubsection{Synthetic data}\label{subsubsec:anisotropic_synthetic_data}
In order to ensure controlled conditions, we start validating our framework using a synthetic dataset. The structure of the synthetic dataset is identical to the experimental one, which is defined at the beginning of this section. For data generation, we use the deterministic four-term model discovered from the experimental dataset in  \cite{martonova_modelDiscoveryCardiacTissue_2025}, which is reported as
\begin{equation}\label{eq:ansatz_sef_anisotropic_4term}
    \begin{aligned}
        \widebarsca{W}_{\mathrm{4-term}}\bigl( \ten{F}, \{\structuralTensorGeneral\}_{\indStructuralTensors=1}^{\numStructuralTensors}; \materialParams\bigr) &=
        5.162 \bigr[ \invariantTwo - 3 \bigl]^{2} \\
        &+ 0.081 \Bigl( 
            \operatorname{exp}\bigl( 
            21.151 \bigl[ \invariantFourFBar - 1\bigr]^{2}
            \bigr) - 1
        \Bigr) \\
        &+ 0.315 \Bigl( 
            \operatorname{exp}\bigl( 
            4.371 \bigl[ \invariantFourNBar - 1\bigr]^{2}
            \bigr) - 1
        \Bigr) \\
        &+ 0.486 \Bigl( 
            \operatorname{exp}\bigl( 
            0.508 \bigl[ \invariantEightFS \bigr]^{2}
            \bigr) - 1
        \Bigr).
    \end{aligned}
\end{equation}
After generating the data, we added heteroscedastic Gaussian noise to the data according to the data model defined in \cref{subsubsec:gps}. Here, we also assume a minimum and a relative error standard deviation of $\minErrorStddev = \SI{0.01}{\kilo\pascal}$ and $\relativeErrorStddev = \SI{5}{\percent}$, respectively. \\

\begin{figure}[h!]
    \centering
    \includegraphics[width=\linewidth]{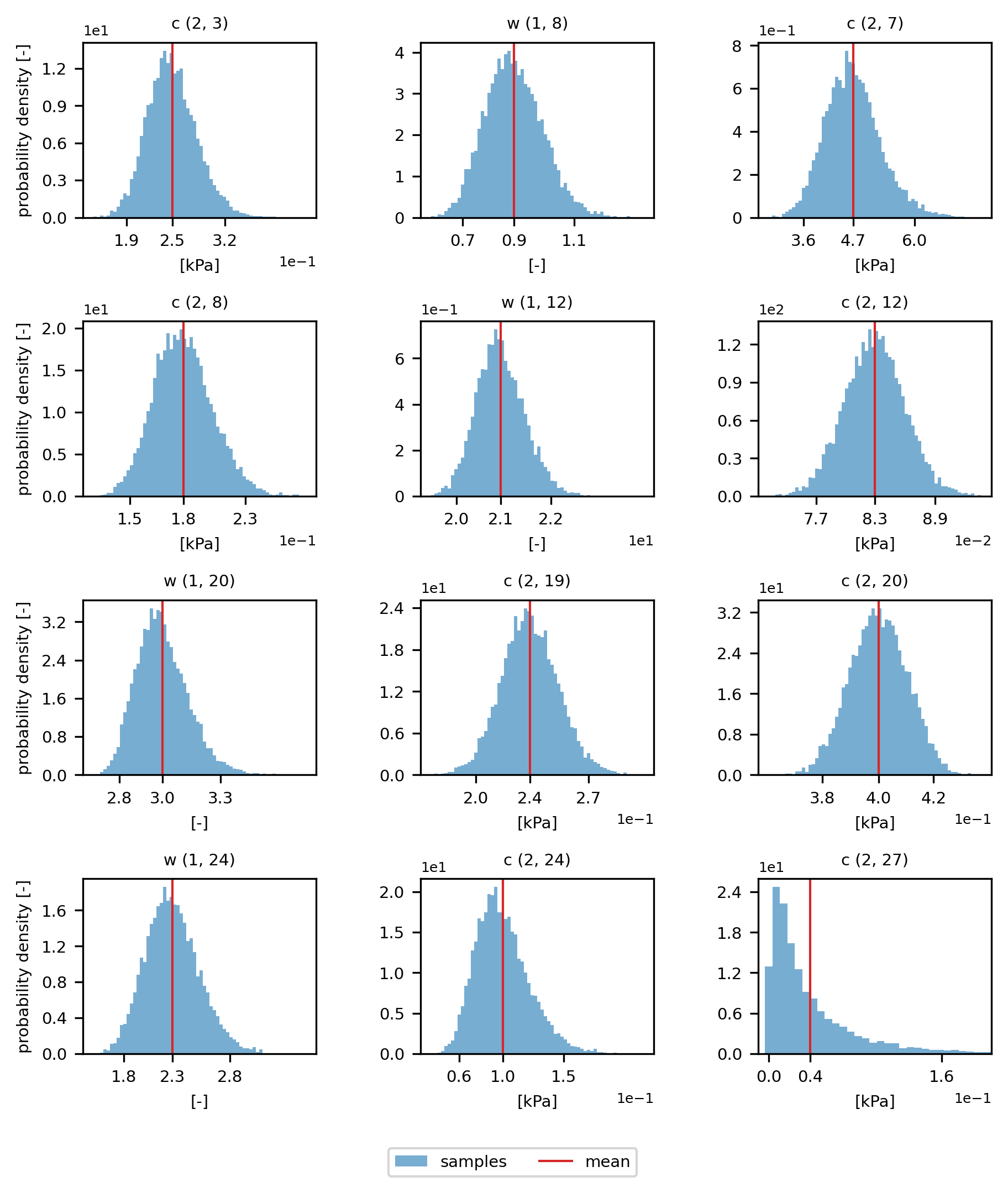}       
    \caption{\textbf{Synthetic anisotropic test case}: Distilled distribution of the material parameters after the sensitivity analysis and model refinement. The distribution over stress-deformation functions induced by this parameter distribution is shown in \cref{fig:model_anisotropic_synthetic}.}
    \label{fig:parameters_anisotropic_synthetic}
\end{figure}

\noindent \textbf{Discovered model:}
We discover an isochoric \ac{SEF} of the form
\begin{equation}\label{eq:result_model_anisotropic_synthetic}
    \begin{aligned}
        \widebarsca{W}\bigl( \ten{F}, \{\structuralTensorGeneral\}_{\indStructuralTensors=1}^{\numStructuralTensors}; \reducedMaterialParams\bigr) &=
        \materialParamLinear^{(2,3)} \bigr[ \invariantOne - 3 \bigl]^{2} \\
        &+ \materialParamLinear^{(2,7)} \bigr[ \invariantTwo - 3 \bigl]^{2}
        + \materialParamLinear^{(2,8)} \Bigl( 
            \operatorname{exp}\bigl( 
                \materialParamNonlinearSingle^{(1,8)} \bigr[ \invariantTwo - 3 \bigl]^{2}
            \bigr) - 1
        \Bigr) \\
        &+ \materialParamLinear^{(2,12)} \Bigl( 
            \operatorname{exp}\bigl( 
            \materialParamNonlinearSingle^{(1,12)} \bigl[ \invariantFourFBar - 1\bigr]^{2}
            \bigr) - 1
        \Bigr) \\
        &+ \materialParamLinear^{(2,19)} \bigr[ \invariantFourNBar - 1 \bigl]^{2}
        + \materialParamLinear^{(2,20)} \Bigl( 
            \operatorname{exp}\bigl( 
            \materialParamNonlinearSingle^{(1,20)} \bigl[ \invariantFourNBar - 1\bigr]^{2}
            \bigr) - 1
        \Bigr) \\
        &+ \materialParamLinear^{(2,24)} \Bigl( 
            \operatorname{exp}\bigl( 
            \materialParamNonlinearSingle^{(1,24)} \bigl[ \invariantEightFS \bigr]^{2}
            \bigr) - 1
        \Bigr) \\
        &+ \materialParamLinear^{(2,27)} \bigr[ \invariantEightFN \bigl]^{2}.
    \end{aligned}
\end{equation}
The distribution of the reduced material parameters $\reducedMaterialParams$ is approximated by the \ac{NF} $\reducedMaterialParamsDistNFOptimized$ and is shown in \cref{fig:parameters_anisotropic_synthetic}. Compared to the four-term model used for data generation, the discovered model contains four additional terms, but only two of them are based on invariants which were not considered for data generation, namely the first invariant $\invariantOne$ and the eighth invariant $\invariantEightFN$. The results of the sensitivity analysis with respect to the initialization of the involved \acp{NN} are reported in \cref{tab:appendix_initialization_anisotropic_synthetic}. For all five initializations, isochoric \acp{SEF} similar to \cref{eq:result_model_anisotropic_synthetic} were discovered: the five model structures discovered have a total of between eight and ten terms, with seven terms in common. All five models contain the four terms used for data generation.

The results show that for none of the five initializations of the \acp{NN}, the constitutive model used for data generation was discovered and that all five discovered models have more than four terms. One key reason for this is that we do not aim to fit the model to the specific stress-deformation function defined in \cref{eq:ansatz_sef_anisotropic_4term}. Instead, we aim to match a parameterized statistical model to the distribution defined by the \ac{GP} posterior. By some probability, this distribution may yield some stress-deformation functions with deviating characteristics that can only be described by slightly different constitutive models. Even in a deterministic setting, noise can make the problem ill-posed or at least complicate the identifiability, see, e.g., \cite{flaschel_brainEUCLID_2023}.
Therefore, we assume that additional terms are needed to achieve the required flexibility of the statistical model. Another possible reason is the collinearity between the different terms, which may also complicate model discovery \cite{urrea_automatedModelDiscovery_2025,mcculloch_LPRegularizationModelDiscovery_2024} and cause identifiability problems \cite{hartmann_identifiabilityMaterialParameters_2018}. Eventually, collinearity prevents the discovery of a unique model.\\

\begin{figure}[p!]
    \centering
    \includegraphics[width=\linewidth]{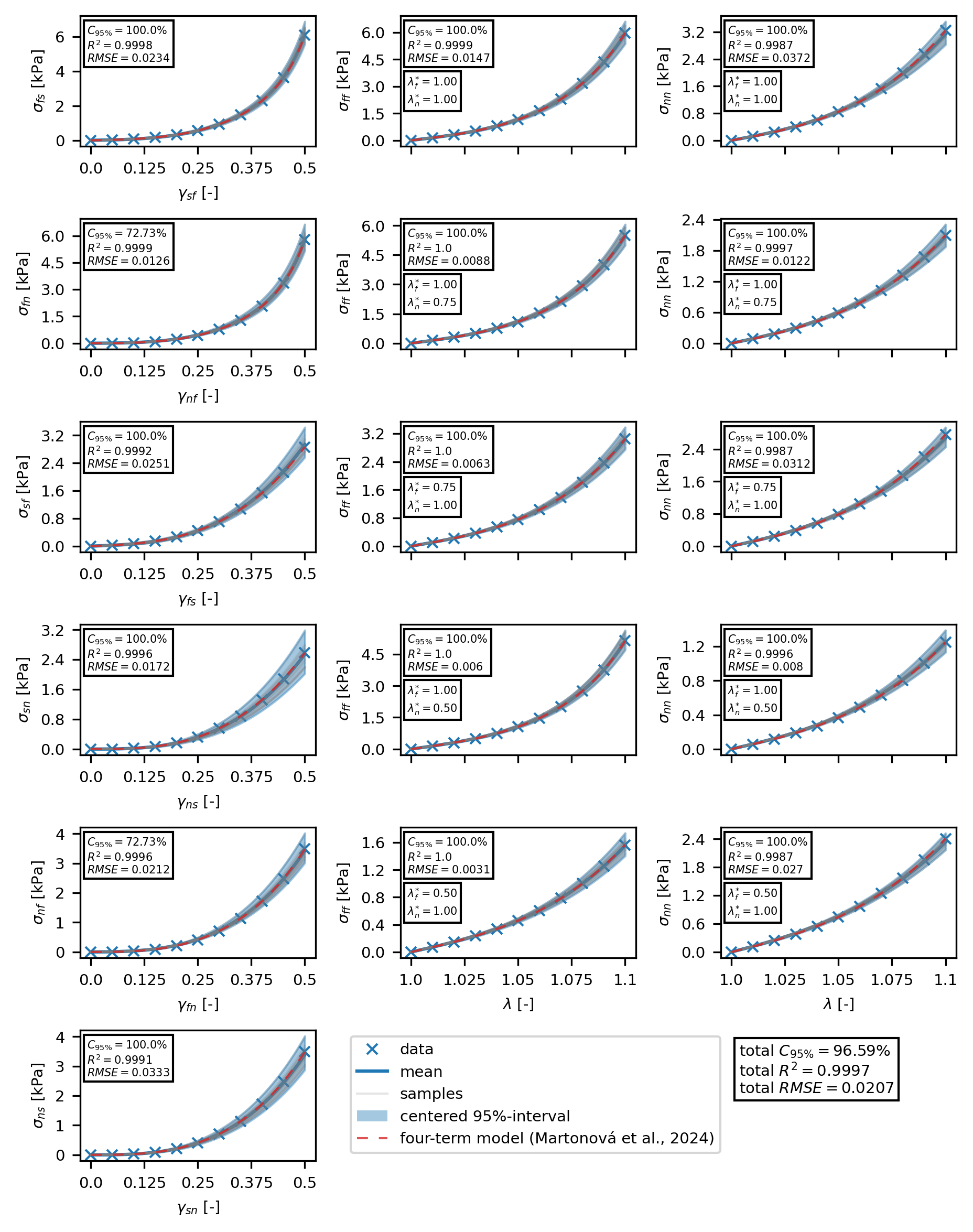}       
    \caption{\textbf{Synthetic anisotropic test case}: Distilled interpretable statistical model. The illustrations show the mean stress-deformation functions, the centered \SI{95}{\percent}-intervals, some random stress-deformation function samples as well as the individual and total validation metrics fo the six \ac{SS} and five \ac{BT} tests. The \ac{RMSE} and the \ac{R2} refer to the mean stress-deformation functions, respectively, and show a good fit. In addition, the total estimated coverage is close to $\SI{95}{\percent}$ and thus indicates a good estimation of the uncertainty.}
    \label{fig:model_anisotropic_synthetic}
\end{figure}

The distribution of stress-deformation functions induced by the distribution over material parameters is shown in \cref{fig:model_anisotropic_synthetic}. In contrast to the experimental datasets, the true stress values are known for the synthetic test case. The validation metrics \ac{R2}, \ac{RMSE}, and the coverage are therefore calculated using the true stress values. Please note, however, that the reported coverages are still only estimates, as they are calculated based on relatively few data points, see \cref{sec:appendix_ci_and_estimated_coverage}. Both the high $\rSquare = \num{0.9997}$ and the low $\rmse = \num{0.0207}$ values prove a very good fit of the mean stress-deformation function to the true data. The coverage of the true data $\coverage = \SI{96.59}{\percent}$ is close to the optimal value of $\SI{95}{\percent}$. With the four other \ac{NN} initializations, for which similar constitutive models were distilled, comparable validation metrics are achieved, see \cref{tab:appendix_initialization_anisotropic_synthetic}. The \textbf{\ac{GP} posterior} representing the target distribution has a coverage of $\coverage = \SI{89.77}{\percent}$, as shown in \cref{fig:gp_anisotropic_synthetic}. The coverages of the five discovered models reported in \cref{tab:appendix_initialization_anisotropic_synthetic} are slightly higher. However, the overestimation of the uncertainty measured in the coverage is for all five initializations only between \SI{5.68}{\percent} and \SI{9.09}{\percent}. We suspect that the slight overestimation is not only affected by the initialization, but also by the choice of model library and the hyperparameters listed in \cref{sec:appendix_hyperparameters}.\\

\begin{figure}[htb]
    \centering
    \includegraphics[width=\linewidth]{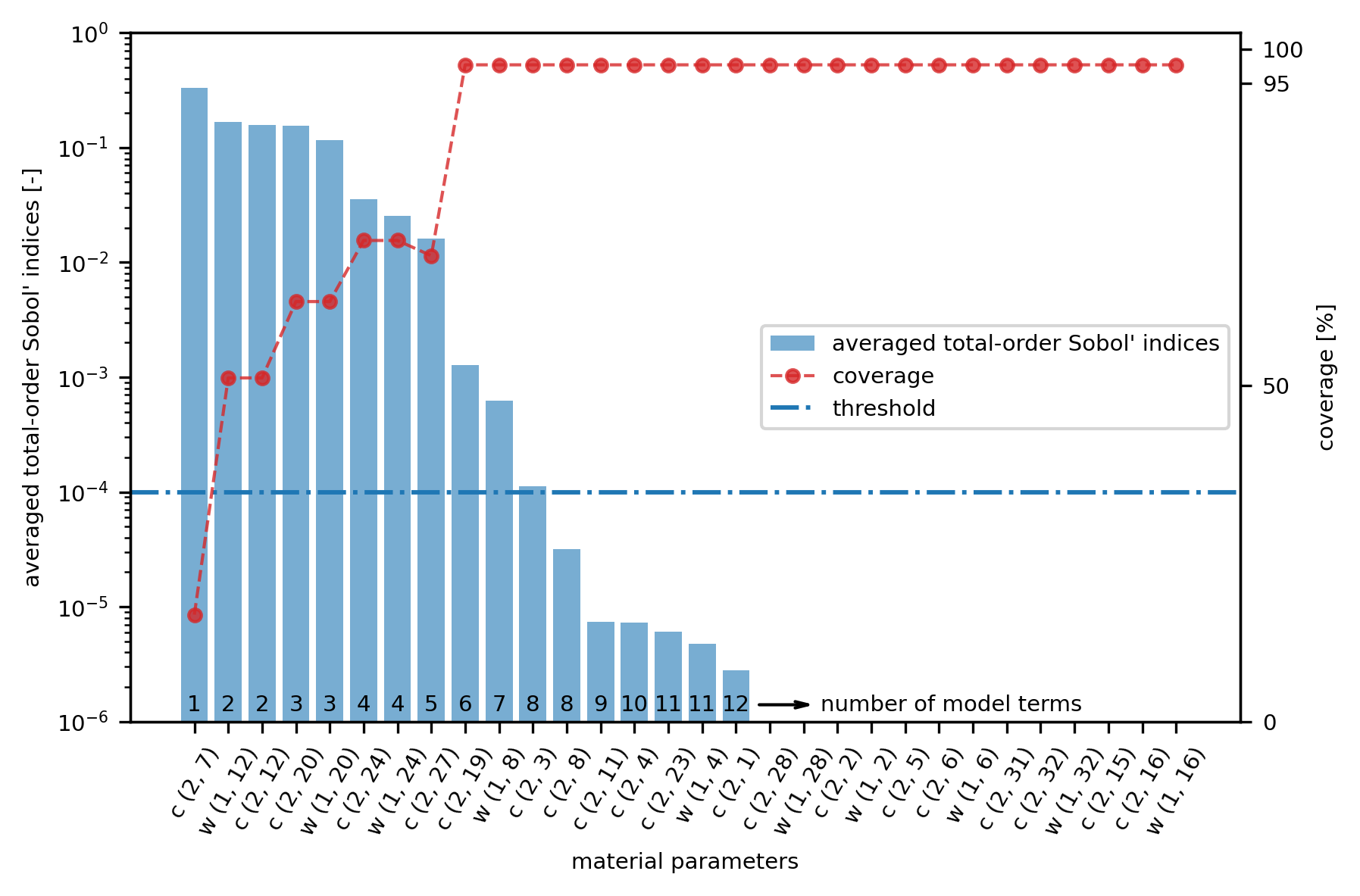}       
    \caption{\textbf{Synthetic anisotropic test case:} Illustration of the model selection process based on the sensitivity analysis. The bars show the total-order Sobol' indices for each material parameter averaged over all mechanical tests, observed stress components and discretization points in their descending order, calculated  according to \cref{eq:total_sobol_index_avergared}. The blue dashed-dotted line indicates the threshold of the total-order Sobol' index below which the material parameters are considered non-relevant. The numbers at the bottom of the bars indicate the total number of model terms when the model term associated with the corresponding material parameter is added to the discovered model. In addition, the red dotted line illustrates how the coverage develops with increasing model size. Note that the estimated coverage may differ slightly after the refinement of the discovered model in the fourth step of the framework.}
    \label{fig:sensitivities_anisotropic_synthetic}
\end{figure}

\begin{figure}[p!]
    \centering
    \includegraphics[width=\linewidth]{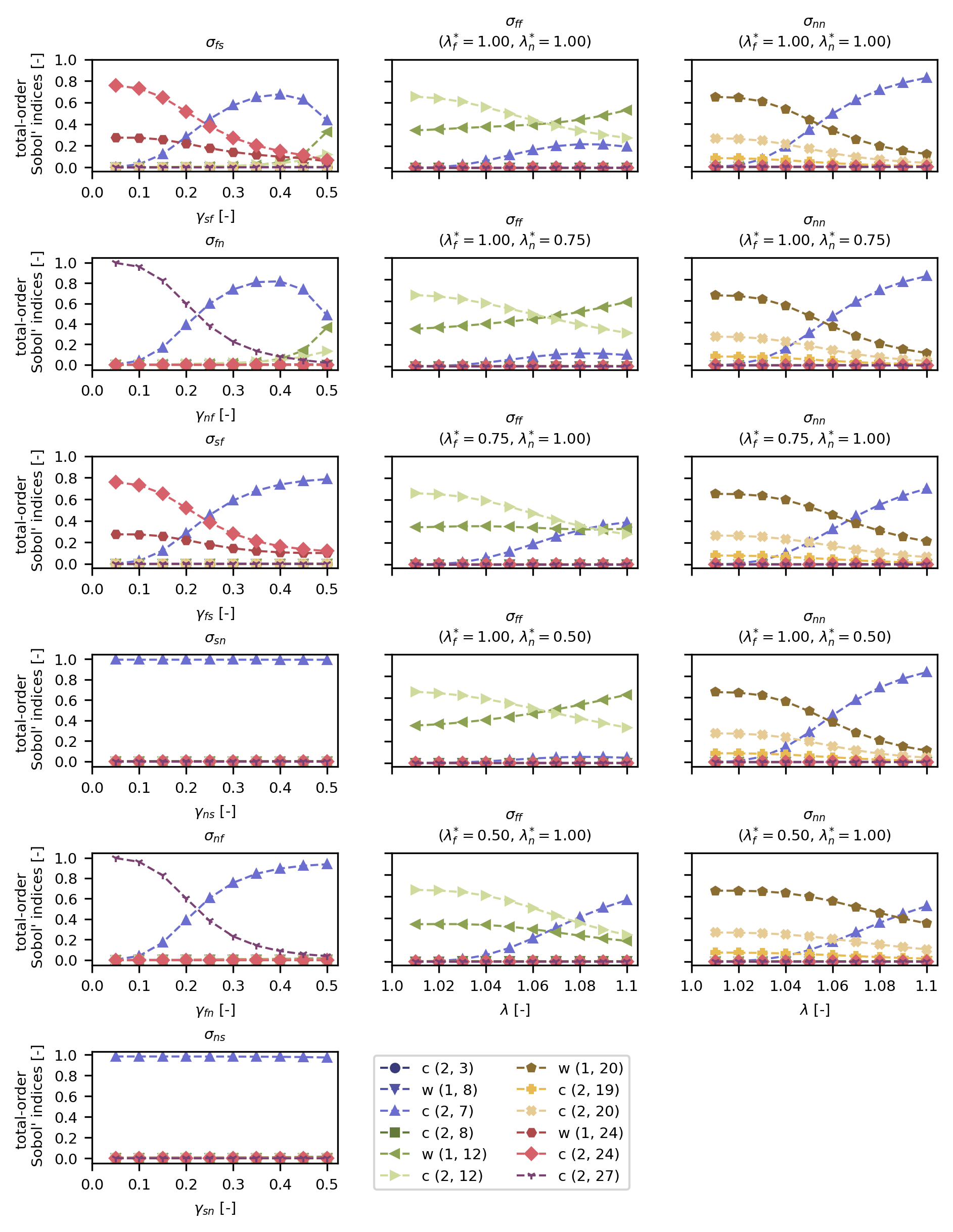}       
    \caption{\textbf{Synthetic anisotropic test case}: Development of total-order Sobol' indices over the course of the mechanical tests. From the results we can make the following observations: \textbf{(i)} Different mechanical tests activate different subsets of \ac{SEF} terms, i.e., terms that are negligible in one deformation mode can dominate in another, and \textbf{(ii)} term importance varies with deformation level. Overall, the deformation-dependent Sobol' analysis clarifies how anisotropic mechanisms emerge under different loading paths and illustrates that the discovered model structure is shaped by both deformation mode and deformation range.
    }    
    \label{fig:sensitivities_path_anisotropic_synthetic}
\end{figure}

\noindent \textbf{Sensitivity-based model selection:} For an illustration of the model selection process, see \cref{fig:sensitivities_anisotropic_synthetic}. In \cref{fig:sensitivities_path_anisotropic_synthetic}, we illustrate the development of the total-order Sobol' indices for increasing deformation for all 11 tests and associated measured stress components. From the results, we can observe that the stresses in the various mechanical tests are sensitive to different terms in the \ac{SEF}. This is expected due to the direction-dependent properties of the anisotropic material. In the \ac{BT} tests, the stress component $\sigma_{\mathrm{ff}}$ in the fiber-direction is dominated by the term based on the invariant $\invariantFourF$ and the component $\sigma_{\mathrm{nn}}$ in the normal direction is more sensitive to the term based on the invariant $\invariantFourN$ and the isotropic invariant $\invariantTwo$. The shear stresses in the fiber-shear plane $\sigma_{\mathrm{fs}}$ and $\sigma_{\mathrm{sf}}$ and fiber-normal plane $\sigma_{\mathrm{fn}}$ and $\sigma_{\mathrm{nf}}$ clearly show sensitivity to the respective eighth invariants $\invariantEightFS$ and $\invariantEightFN$ for small to medium deformation ranges. As deformations increase, the influence of terms based on the eighth invariants decreases, and in particular the terms based on $\invariantTwo$ and $\invariantFourFBar$ gain importance. In contrast, the sensitivity of the shear stresses in the shear-normal plane $\sigma_{\mathrm{sn}}$ and $\sigma_{\mathrm{ns}}$ is clearly dominated by the isotropic second invariant $\invariantTwo$. Similar to the isotropic test case, the sensitivities are generally not constant and may vary with increasing deformation.

\subsubsection{Experimental data}\label{subsubsec:anisotropic_experimental_data}

Finally, we apply the framework presented in \cref{sec:methodology} to the experimental dataset collected in mechanical tests with anisotropic human cardiac tissue. The structure of the dataset is described at the beginning of the section and is identical to that of the synthetic dataset that we considered before in \cref{subsubsec:anisotropic_synthetic_data}.\\

\noindent \textbf{GP posterior:} 
The \ac{GP} posterior for all mechanical tests is shown in the appendix, see \cref{fig:gp_anisotropic_experimental}. The coverage estimation results in a total estimated coverage of the centered \SI{95}{\percent}-interval of $\estimatedCoverage=\SI{93.75}{\percent}$ and thus indicates that the \ac{GP} posterior correctly reflects the uncertainty in the measured data. Note that the \ac{GP} posterior is not physically consistent in a few cases, e.g., in the \ac{BT} test for $\sca{\lambda}_{\mathrm{f}}^{*} = 1.0$ and $\sca{\lambda}_{\mathrm{n}}^{*} = 0.5$. A non-physical \ac{GP} posterior may lead to a mismatch between the distributions, i.e., a Wasserstein-1 distance greater than zero, but the statistical model is ultimately physically consistent by construction of the model library for the \ac{SEF}, see \cref{sec:methodology}.\\

\begin{figure}[h!]
    \centering
    \includegraphics[width=\linewidth]{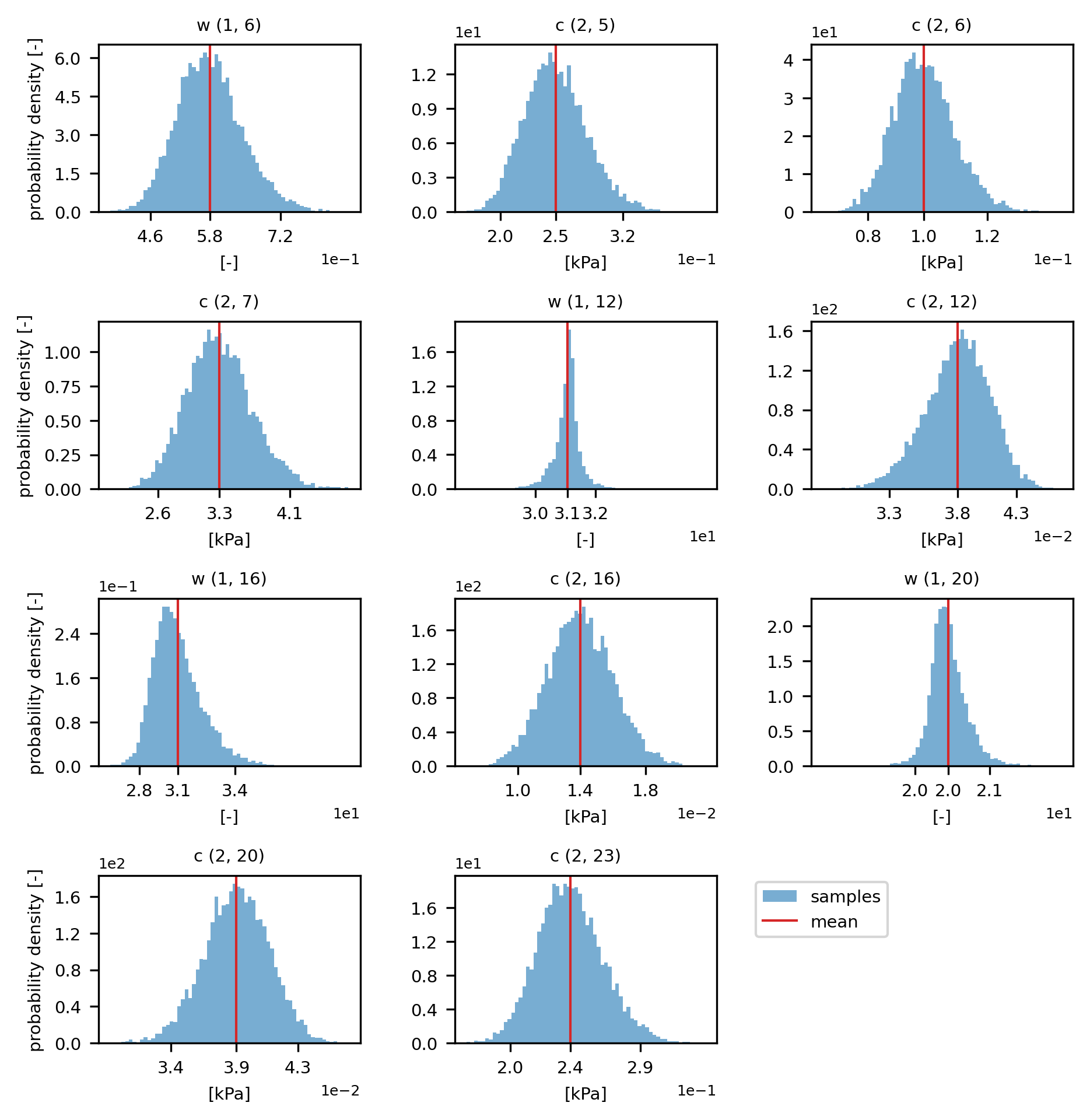}       
    \caption{\textbf{Experimental anisotropic test case}: Distilled distribution of the material parameters after the sensitivity analysis and model refinement. The distribution over stress-deformation functions induced by this parameter distribution is shown in \cref{fig:model_anisotropic_experimental}.}
    \label{fig:parameters_anisotropic_experimental}
\end{figure}

\noindent \textbf{Discovered model:} The minimization of the Wasserstein-1 distance between the \ac{GP} posterior and the statistical model yields the following explicit form of the isochoric \ac{SEF}
\begin{equation}\label{eq:result_model_anisotropic_experimental}
    \begin{aligned}
        \widebarsca{W}\bigl( \ten{F}, \{\structuralTensorGeneral\}_{\indStructuralTensors=1}^{\numStructuralTensors}; \reducedMaterialParams\bigr) &=
        \materialParamLinear^{(2,5)} \bigr[ \invariantTwo - 3 \bigl]
        + \materialParamLinear^{(2,6)} \Bigl( 
            \operatorname{exp}\bigl( 
                \materialParamNonlinearSingle^{(1,6)} \bigl[ \invariantTwo - 3\bigr]
            \bigr) - 1
        \Bigr) \\
        &+ \materialParamLinear^{(2,7)} \bigr[ \invariantTwo - 3 \bigl]^{2} \\
        &+ \materialParamLinear^{(2,12)} \Bigl( 
            \operatorname{exp}\bigl( 
            \materialParamNonlinearSingle^{(1,12)} \bigl[ \invariantFourFBar - 1\bigr]^{2}
            \bigr) - 1
        \Bigr) \\
        &+ \materialParamLinear^{(2,16)} \Bigl( 
            \operatorname{exp}\bigl( 
            \materialParamNonlinearSingle^{(1,16)} \bigl[ \invariantFourSBar - 1\bigr]^{2}
            \bigr) - 1
        \Bigr) \\
        &+ \materialParamLinear^{(2,20)} \Bigl( 
            \operatorname{exp}\bigl( 
            \materialParamNonlinearSingle^{(1,20)} \bigl[ \invariantFourNBar - 1\bigr]^{2}
            \bigr) - 1
        \Bigr) \\
        &+ \materialParamLinear^{(2,23)} \bigr[ \invariantEightFS \bigl]^{2},
    \end{aligned}
\end{equation}
where the distribution of the reduced material parameters is approximated by the \ac{NF} $\reducedMaterialParamsDistNFOptimized$ and is shown in \cref{fig:parameters_anisotropic_experimental}. The sensitivity analysis with respect to the initialization of the \acp{NN} demonstrates the robustness of the proposed framework for the experimental anisotropic dataset. For all five initializations of the \acp{NN} involved, the same model structure defined in \cref{eq:result_model_anisotropic_experimental} was consistently discovered, see \cref{tab:appendix_initialization_anisotropic_experimental}.

\begin{figure}[p!]
    \centering
    \includegraphics[width=\linewidth]{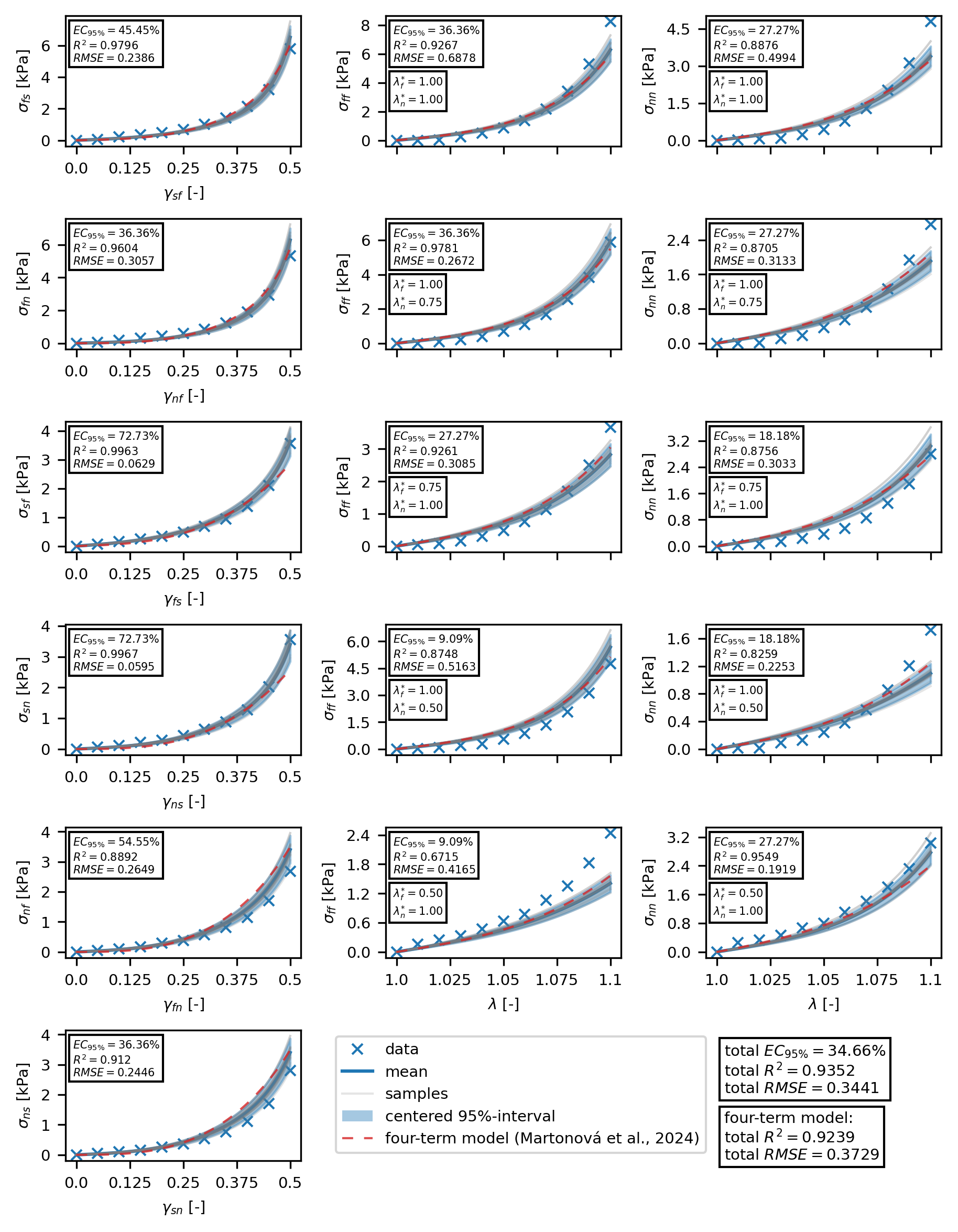}       
    \caption{\textbf{Experimental anisotropic test case}: Distilled interpretable statistical model. The illustrations show the mean stress-deformation functions, the centered \SI{95}{\percent}-intervals, some random stress-deformation function samples as well as the individual and total validation metrics for the six \ac{SS} and five \ac{BT} tests. The \ac{RMSE} and the \ac{R2} refer to the mean stress-deformation functions, respectively. For comparison, we also show the four-term model from \cref{eq:ansatz_sef_anisotropic_4term} that was discovered for the experimental dataset in \cite{martonova_modelDiscoveryCardiacTissue_2025}.}
    \label{fig:model_anisotropic_experimental}
\end{figure}

The distribution of the stress-deformation functions induced by the parameter distribution is illustrated in \cref{fig:model_anisotropic_experimental}. The quality of the fit is with $\rSquare = \num{0.9352}$ and $\rmse = \num{0.3441}$ comparable to the one observed with the four-term model \cref{eq:ansatz_sef_anisotropic_4term} previously discovered in \cite{martonova_modelDiscoveryCardiacTissue_2025} ($\rSquare = \num{0.9239}, \rmse = \num{0.3729}$). However, the estimated coverage of the centered \SI{95}{\percent}-interval is only $\estimatedCoverage=\SI{34.66}{\percent}$ and thus significantly lower than the target value of $\SI{95}{\percent}$. Almost identical values for the validation metrics are obtained for the remaining four initializations. 
\\

Recall that for the synthetic anisotropic dataset with artificial noise, we achieved both a good mean fit and good coverage, as reported in \cref{subsubsec:anisotropic_synthetic_data}. We have thus demonstrated that our framework is generally capable of reliably discovering statistical anisotropic constitutive models. However, in the synthetic test case, we generated the dataset using terms from the same model library that was subsequently used for model discovery. 

In contrast, for the experimental dataset, the same model library may lack flexibility to some extent. A lack of flexibility is strongly related to possibly incorrect modeling assumptions or missing physics. An example of an incorrect assumption would be that the actual orientation of the tissue fibers does not exactly match the orientation assumed in the model library. In addition, the tested cardiac tissue may exhibit viscous effects that are not taken into account in the model library. If the assumptions made when formulating the model library do not accurately reflect reality, then the library's flexibility is insufficient, and the incorrect assumptions will likely induce a model-reality mismatch. For a discussion on the modeling assumptions and missing physics in the constitutive model formulation, we refer to \cite{vaverka_modificationModelMyocardium_2025,brown_cardiacMechanicsModeling_2025}. Other possible reasons for the slight deterioration in mean fit and low estimated coverage compared to the synthetic dataset could be measurement artifacts.

Furthermore, we point out that a reasonable mean fit is not a sufficient condition for accurate coverage. The requirements for the flexibility of the model library are significantly higher for uncertainty quantification than for achieving a good mean fit. For uncertainty quantification, the model library must be flexible enough to model all physically admissible stress-deformation functions from the \ac{GP} posterior, not just the one function that represents a good mean fit.\\

\begin{figure}[htb]
    \centering
    \includegraphics[width=\linewidth]{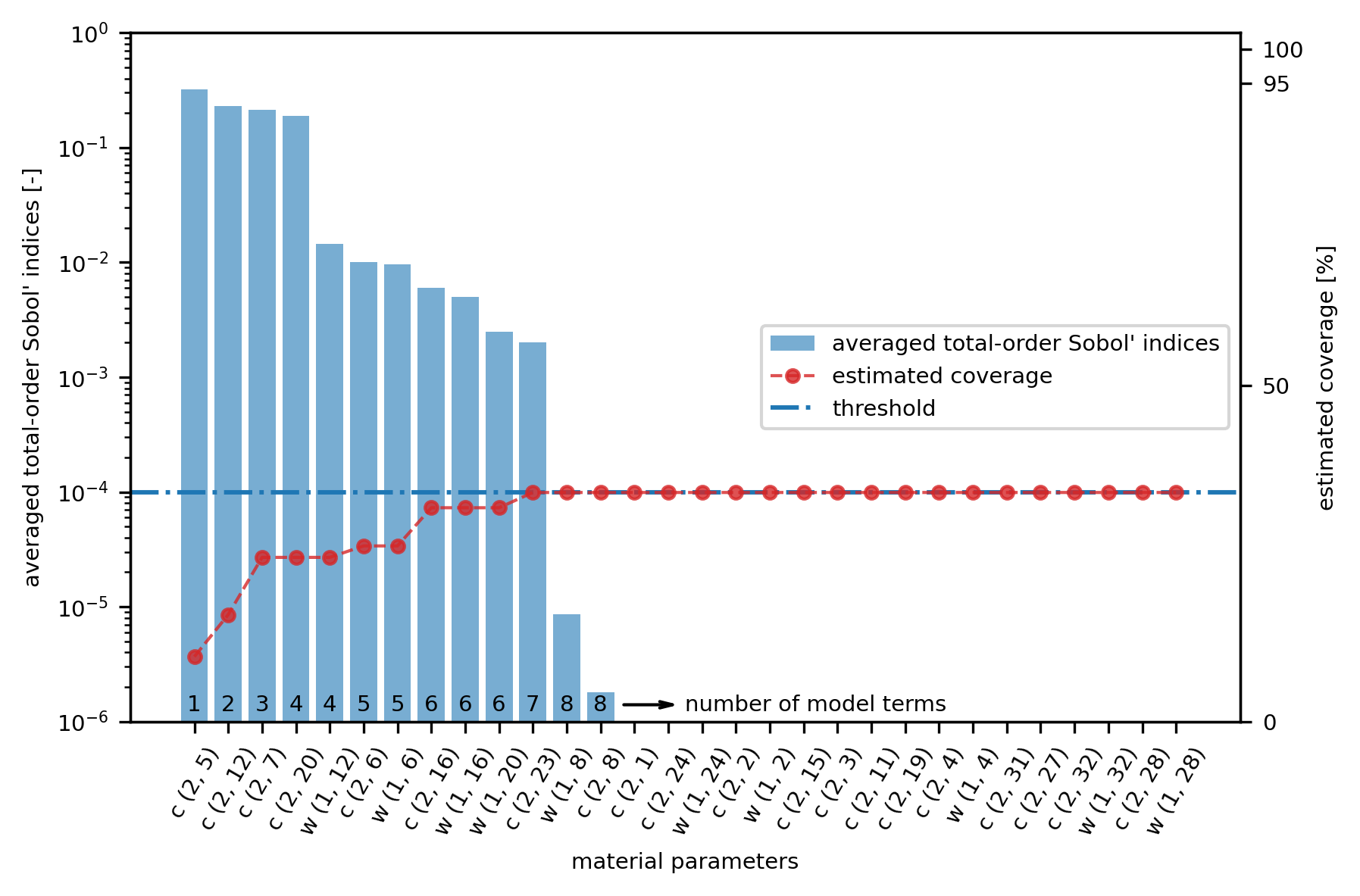}
    \caption{\textbf{Experimental anisotropic test case:} Illustration of the model selection process based on the sensitivity analysis. The bars show the total-order Sobol' indices for each material parameter averaged over all mechanical tests, observed stress components and discretization points in their descending order, calculated  according to \cref{eq:total_sobol_index_avergared}. The blue dashed-dotted line indicates the threshold of the total-order Sobol' index below which the material parameters are considered non-relevant. The numbers at the bottom of the bars indicate the total number of model terms when the model term associated with the corresponding material parameter is added to the discovered model. In addition, the red dotted line illustrates how the empirical coverage develops with increasing model size. Note that the estimated coverage may differ slightly after the refinement of the discovered model in the fourth step of the framework.}
    \label{fig:sensitivities_anisotropic_experimental}
\end{figure}

\begin{figure}[p!]
    \centering
    \includegraphics[width=\linewidth]{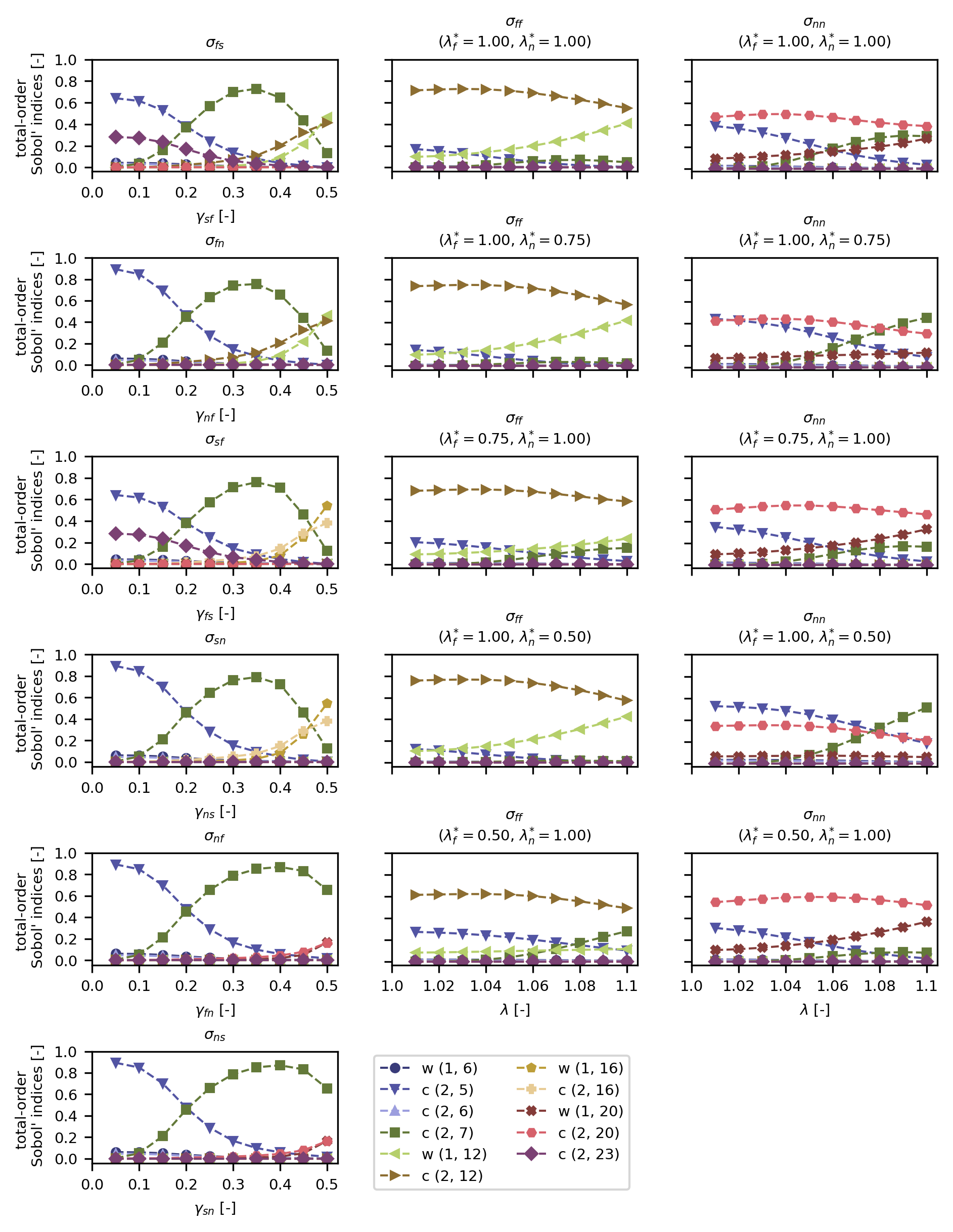}       
    \caption{\textbf{Experimental anisotropic test case}: Development of total-order Sobol' indices over the course of the mechanical tests. From the results we can make the following observations: \textbf{(i)} Different mechanical tests activate different subsets of \ac{SEF} terms, i.e., terms that are negligible in one deformation mode can dominate in another, and \textbf{(ii)} term importance varies with deformation level. For the anisotropic test case, this deformation-dependent sensitivity of the model terms provides a data-driven interpretation of which isotropic and direction-specific mechanisms are actually supported by the dataset.}    
    \label{fig:sensitivities_path_anisotropic_experimental}
\end{figure}

\noindent \textbf{Sensitivity-based model selection:} For information on the model selection process, the reader is referred to \cref{fig:sensitivities_anisotropic_experimental}. In \cref{fig:sensitivities_path_anisotropic_experimental}, we illustrate the development of the total-order Sobol' indices with increasing deformation for all \num{11} mechanical tests and associated stress components. The results show that the relevance of individual terms in the \ac{SEF} is strongly dependent on the deformation mode, loading direction, and deformation level, reflecting the anisotropic behavior of the cardiac tissue. Across the \ac{BT} tests, the Sobol’ indices exhibit a hierarchy of dominant mechanisms. The normal stress $\sigma_{\mathrm{ff}}$ in the fiber-direction is primarily governed by the fiber reinforcement term $\invariantFourFBar$, whereas the transverse normal stress $\sigma_{\mathrm{nn}}$ is primarily governed by the corresponding transverse reinforcement term $\invariantFourNBar$. This behavior is consistent with the experimentally observed preferential stiffness along the myocardial fiber-direction. In addition, there is evidence that, under loading transverse to the fibers, the load is shared by secondary structural directions (sheet and sheet‑normal) that increasingly contribute at higher stretches. This motivates the use of separate direction‑specific reinforcement terms in constitutive models \cite{holzapfel_constitutiveModellingMyocardium_2009,laita_myocardiumModelingMultimodalDeformations_2025}. In contrast to this anisotropic dominance, both stress components show an almost identical deformation-dependent isotropic pattern with respect to $\invariantTwo$. In the small-deformation regime the linear contribution is most influential, whereas at larger stretches the non-linear contribution of $\invariantTwo$ becomes dominant, reflecting an amplification of higher-order isotropic non-linearities with increasing deformation. That is, the Sobol' indices analysis suggests a progressive activation of non-linear stress contributions.

For all shear components, the response at small shear strains is governed by the linear isotropic contribution of $\invariantTwo$, while at intermediate shear strains its quadratic form becomes most influential. Toward the largest shear deformations, the contribution of $\invariantTwo$ decreases again. This coincides with an increase in the sensitivity of model terms based on anisotropic invariants, so that anisotropic mechanisms become non-negligible precisely in the high-shear regime. The observed relevance of direction-specific reinforcement terms is standard for myocardium and other fibrous soft tissues to represent interactions between preferred directions under shear \cite{holzapfel_constitutiveModellingMyocardium_2009,laita_myocardiumModelingMultimodalDeformations_2025}. Our analysis suggests that the coupling term $\invariantEightFS$ enters only as a secondary contribution, rather than acting as the primary driver of the shear response in these experiments. This is consistent with reported myocardial shear measurements \cite{dokos_shearPropertiesMyocardium_2002,owen_modellingCardiovascularSystem_2018}.
In this context, the deformation-dependent Sobol’ analysis provides an interpretation of how different structural mechanisms encoded in the invariant formulation are emphasized under different loading paths and highlights that the specific form of the discovered constitutive model is not only influenced by the deformation modes but by the deformation range explored in the experimental setup.

\section{Conclusion and outlook}\label{sec:conclusion}

In this contribution, we proposed a versatile and general statistical framework for uncertainty quantification in supervised model discovery. The key idea behind this framework is to distill the information and uncertainty encoded in a Gaussian process posterior that augments the available noisy stress-deformation data into a material constitutive model. Sparsity of the discovered model is promoted by a Sobol' sensitivity analysis. As a result, we ultimately obtain interpretable and sparse material constitutive models and a joint probability distribution of their material parameters, which can be used for uncertainty quantification. 

The proposed framework is only partially Bayesian and does not require the selection of a prior for the material parameters. Note that the absence of prior knowledge is the original motivation for model discovery. By using normalizing flows for density estimation, the framework is able to represent complex and high-dimensional joint probability distributions of the sought material parameters.\\

We demonstrated the capability of our framework for several test cases, including experimental datasets with isotropic and anisotropic hyperelastic materials. The results for the experimental isotropic and synthetic anisotropic data clearly show that the mean stress-deformation functions of the discovered statistical models are close to the data with only minor deviations. The coverage estimation also indicated a well-calibrated estimate of the uncertainties. For the experimental anisotropic dataset, the quality of the mean fit is comparable to the reference solution reported in \cite{martonova_modelDiscoveryCardiacTissue_2025}. However, the mean fit has a lower accuracy compared to our synthetic test case and uncertainties are underestimated. Possible reasons for the observed accuracy and coverage deterioration could be measurement artifacts and incorrect modeling assumptions, e.g., with regard to fiber orientation. In particular, we suspect incorrect modeling assumptions and missing physics, such as viscous effects, to cause a lack of flexibility of the model library. In order to better capture the uncertainties in the experimental anisotropic test case, model libraries with more flexibility are required that take into account any missing physics and also fiber dispersion, as proposed, e.g., in \cite{martonova_discoveringDispersion_2025}. For a discussion on the modeling assumptions and missing physics in the constitutive model formulation, the reader is referred to \cite{vaverka_modificationModelMyocardium_2025,brown_cardiacMechanicsModeling_2025}.

In addition, our sensitivity analyses show that the contributions of the selected model terms vary for different mechanical tests but also for different levels of deformation in the same test. We therefore believe that Sobol' indices are a promising technique to get further insights into the model selection process and to support informed decisions in model discovery. Moreover, sensitivity analysis also allows us to optimize the design of the model discovery problem, including the experimental setup and the formulation of the model library.\\

We believe that our framework is a promising approach to uncertainty quantification in model discovery with potential for further developments and applications beyond continuum solid mechanics. Unlike the model library, the Gaussian process posterior is currently not yet constrained to physically admissible stress-deformation functions but is only conditioned on data. Therefore, future work is required to add physical constraints to the Gaussian process posterior which may improve the framework's performance. The monotonicity constraint of the stress-deformation function, e.g., could be fulfilled by using monotonic Gaussian process flows \cite{ustyuzhaninov_monotonicGPFlows_2020}. One possible approach to correlate the stress components with each other is to model the strain energy density function by a correlated multi-output \ac{GP}, as proposed, e.g., by \cite{aggarwal_strainEnergyDensityAsGP_2023}. The stresses could then be obtained by deriving the correlated multi-output \ac{GP} with respect to the kinematic quantities. However, this approach is not straightforward and requires further investigations. 

Furthermore, future research should focus on ensuring that the Gaussian process posterior correctly estimates the uncertainty in the data, since the Gaussian process ultimately represents the target distribution. We believe that it could be promising to infer all hyperparameters of the \ac{GP} posterior, including those of the error model, in a full Bayesian setting. The kernel length scales would then no longer need to be adopted, and the error parameters would no longer be assumed to be known. However, the full Bayesian approach requires the selection of priors for both the error model parameters and the kernel hyperparameters, which is not straightforward. There is also no guarantee that the full Bayesian approach results in well-calibrated uncertainty of the \ac{GP} posterior. 

We also aim to improve the error model for the measurement data. We currently assume that the contribution of sample variability to the overall error is independently distributed, which generally does not fully reflect reality. One option for modeling the variability term in the error model more accurately is to use non-stationary Gaussian processes, as done, e.g., in \cite{narouie_statFEMPolynomialChaos_2025}.\\

We expect that our framework can generally be applied to any constitutive model library formulated as a series expansion with linear and non-linear parameters, c.f. \cref{eq:discovered_reduced_model}. In addition, the application to neural network-based constitutive models that rely on function composition is expected to be possible. However, in the future, it still needs to be investigated whether normalizing flows can also approximate the distribution over material parameters for model libraries and neural networks with a large number of parameters. Finally, we intend to extend the current framework to inelastic material behavior and the unsupervised setting.


\section*{CRediT authorship contribution statement}

\textbf{David Anton:} Conceptualization, Methodology, Software, Validation, Investigation, Data Curation, Writing - Original Draft, Writing - Review \& Editing, Visualization \textbf{Henning Wessels:} Conceptualization, Resources, Writing - Original Draft, Writing - Review \& Editing, Supervision, Funding acquisition \textbf{Ulrich Römer:} Methodology, Writing - Original Draft, Writing - Review \& Editing, Supervision \textbf{Alexander Henkes:} Methodology, Writing - Review \& Editing, Supervision \textbf{Jorge-Humberto Urrea-Quintero:} Conceptualization, Methodology, Investigation, Writing - Original Draft, Writing - Review \& Editing, Supervision

\section*{Declaration of competing interest}
The authors declare that they have no known competing financial interests or personal relationships that could have appeared to influence the work reported in this paper.

\section*{Acknowledgments}
David Anton and Henning Wessels acknowledge support in the project DFG 501798687: \textit{"Monitoring data driven life cycle management with AR based on adaptive, AI-supported corrosion prediction for reinforced concrete structures under combined impacts"} which is a subproject of SPP 2388: \textit{"Hundred plus - Extending the Lifetime of Complex Engineering Structures through Intelligent Digitalization"} funded by the DFG. Alexander Henkes acknowledges support by an ETH Zurich Postdoctoral Fellowship.
The authors also acknowledge the valuable comments of the anonymous reviewers, which led to substantial improvements of the manuscript.

\section*{Data availability}
Our research code is available on GitHub and Zenodo \cite{anton_codeUQInModelDiscovery_2025}.



\appendix

\section{Details on variational perspective}
\label{sec:details_variational}

Instead of targeting $p(\materialParams \mid \dataSet)$ directly, we now detail the two-step approach with a \ac{GP} data surrogate. We assume that there is an unknown, true, stress-deformation function $f$ such that the data is generated as 
\begin{equation}
    \sca{P}^{(i)} = f(\ten{F}^{(i)}) + \epsilon^{(i)}, \quad i=1,\ldots,\numData,
\end{equation}
and $\dataSet = \{ (\ten{F}^{(i)},\sca{P}^{(i)}) \}_{i=1}^{\numData}$. Moreover $\func \elm{\numFuncPoints}$ represents $f$ evaluated at a different grid of $\numFuncPoints$ deformation tensors. The library induces a model of $\func$, given by $\parametersToStateMap: \materialParams \mapsto \parametersToStateMap(\materialParams) \elm{\numFuncPoints}$. Our only aim here is to motivate the variational approach and to establish connections and therefore, we introduce some strong simplifying assumptions. We model $f$, resp. $\func$ as a \ac{GP} and assume that the model library is expressive enough so that for a specific $\materialParams^*$, $\func = \parametersToStateMap(\materialParams^*)$ holds for all possible GP realizations. We also assume that $\parametersToStateMap$ is bijective. 

We proceed by factoring the joint probability distribution $p(\materialParams, \dataSet, \func)$ as
\begin{subequations}
\begin{align}
    p(\materialParams, \dataSet, \func) 
    &=p(\materialParams, \func \mid \dataSet) \, p(\dataSet)
    \label{eq:bayes_joint_pdf_with_functions_a} \\
    &= p(\materialParams \mid \func, \dataSet) \, p(\func, \dataSet) 
    \label{eq:bayes_joint_pdf_with_functions_b} \\
    &= p(\materialParams \mid \func) \, p(\func \mid \dataSet) \, p(\dataSet),
    \label{eq:bayes_joint_pdf_with_functions_c}
\end{align}
\end{subequations}
where we additionally assume conditional independence of $\materialParams$ and $\dataSet$, given $\func$, for simplicity. 
From equating \cref{eq:bayes_joint_pdf_with_functions_a} and \cref{eq:bayes_joint_pdf_with_functions_b}, we obtain the joint posterior
\begin{subequations}
\begin{align}
    p(\materialParams, \func \mid \dataSet) 
    &= \frac{p(\materialParams \mid \func, \dataSet) \, p(\func, \dataSet)}{p(\dataSet)}
    \label{eq:bayes_joint_posterior_with_functions_a} \\
    &= p(\materialParams \mid \func) \, p(\func \mid \dataSet),
    \label{eq:bayes_joint_posterior_with_functions_b}
\end{align}
\end{subequations}
where \cref{eq:bayes_joint_posterior_with_functions_b} is obtained using \cref{eq:bayes_joint_pdf_with_functions_c} and $p(\func \mid \dataSet)$ is the \ac{GP} posterior distribution. 

For the following argument, as a distance in the variational approach \eqref{eq:bayes_variational_approximation}, we consider the Kullback-Leibler divergence $D = \text{KL}$. For a bijective and sufficiently smooth $\parametersToStateMap$, the density $\parametersToStateMapPush\materialParamsDistPush$ can be characterized as 
\begin{equation}
    \parametersToStateMapPush\materialParamsDistPush = p_{\materialParams}\bigl(\parametersToStateMap^{-1}(\func)\bigr) |\det \mathbf{J}_{\parametersToStateMap^{-1}}(\func)|.
\end{equation}
Then, the KL-divergence can be recast as
\begin{equation}\label{eq:appendix_variational_perspective_KL_divergence}
    \text{KL}
        \bigl(
            \parametersToStateMapPush\materialParamsDistPush, p(\func \mid \dataSet)
        \bigr) = \int_{\mathcal{K}} \materialParamsDist \log \frac{\materialParamsDist}{p(\func = \parametersToStateMap(\materialParams) \mid \dataSet) \cdot |\det \mathbf{J}_{\parametersToStateMap}(\materialParams)|} \mathrm{d}\materialParams,
\end{equation}
after pulling the integral back to the parameter domain $\mathcal{K}$. Hence, the minimizer is obtained as 
\begin{equation}
    \materialParamsDistOptimal = p(\func = \parametersToStateMap(\materialParams) \mid \dataSet) \cdot |\det \mathbf{J}_{\parametersToStateMap}(\materialParams)|.
\end{equation}
At the same time, by the change of variables formula
\begin{equation}
    p(\materialParams \mid \dataSet) = p(\func = \parametersToStateMap(\materialParams) \mid \dataSet) |\det \mathbf{J}_{\parametersToStateMap}(\materialParams)|
\end{equation}
and hence $p(\materialParams \mid \dataSet) = \materialParamsDistOptimal$. Note that in the general case, the push-forward density can easily concentrate on a lower dimensional set and the Kullback-Leibler divergence in \cref{eq:appendix_variational_perspective_KL_divergence} becomes infinite. Since we apply \cref{eq:appendix_variational_perspective_KL_divergence} in a much more general and realistic setting than assumed in the above argument, we use the Wasserstein-1 distance throughout this paper instead.


\section{Observation maps and deformation filters definitions}\label{sec:appendix_observation_map_deformation_filter}

This appendix provides the definitions of the observation maps and deformation filters for all numerical test cases. For an introduction to our notation, the reader is referred to \cref{subsec:dataset_and_notation}.

\subsection{Observation maps definitions}\label{subsec:appendix_observation_maps}

The observation map $\observationMapGeneral: \mathbb{R}^{3 \times 3} \to \mathbb{R}$ is used to filter out the stress components $\indObservedP \in \{1, \dots, \numObservedPTesti\}$ observed in the respective mechanical test $\indTests \in \{1, \dots, \numTests\}$. In \cref{subsubsec:appendix_observation_maps_treloar} and \cref{subsubsec:appendix_observation_maps_anisotropic}, we define the observation maps that we use for the Treloar and the anisotropic datasets, respectively.

\subsubsection{Treloar dataset}\label{subsubsec:appendix_observation_maps_treloar}

The Treloar dataset comprises one \acf{UT}, \acf{EBT} and \acf{PS} test. For a clear assignment of the tests, the integer values of the index for the mechanical test $\indTests$ are replaced by their respective abbreviations such that $\indTests \in \{\mathrm{UT}, \mathrm{EBT}, \mathrm{PS}\}$. In all three mechanical tests, only the principal Piola-Kirchhoff stress component $\sca{P}_{11}$ is observed, i.e., $\numObservedPTestSpecific{UT}=\numObservedPTestSpecific{EBT}=\numObservedPTestSpecific{PS}=1$. The observation maps are thus defined as
\begin{equation}\label{eq:appendix_observation_operators_treloar}
    \begin{aligned}
    \observationMapSpecific{UT}{1}(\ten{P}) &= \sca{P}_{11}, \\
    \observationMapSpecific{EBT}{1}(\ten{P}) &= \sca{P}_{11}, \\
    \observationMapSpecific{PS}{1}(\ten{P}) &= \sca{P}_{11}. \\
    \end{aligned}
\end{equation}

\subsubsection{Anisotropic dataset}\label{subsubsec:appendix_observation_maps_anisotropic}

The anisotropic dataset we consider in this paper includes a total of six \acf{SS} and five \acf{BT} tests. According to \cite{sommer_humanVentricularMyocardium_2015}, the three mutually orthogonal directions of the anisotropic human cardiac tissue are the fiber ($\mathrm{f}$), shear ($\mathrm{s}$) and normal ($\mathrm{n}$) directions. In each \ac{SS} test, the corresponding shear stress component of the Cauchy stress tensor is observed, and $\numObservedPTestSpecific{SS}=1$. We again assign the corresponding abbreviations to the integer values of the mechanical test index $\indTests$. The observation maps for the shear tests are then defined as follows
\begin{equation}\label{eq:appendix_observation_operators_anisotropic_ps}
    \begin{aligned}
    \observationMapSpecific{SS_{sf}}{1}(\ten{\sigma}) &= \sca{\sigma}_{\mathrm{fs}}, & \observationMapSpecific{SS_{fs}}{1}(\ten{\sigma}) &= \sca{\sigma}_{\mathrm{sf}}, \\
    \observationMapSpecific{SS_{nf}}{1}(\ten{\sigma}) &= \sca{\sigma}_{\mathrm{fn}}, & \observationMapSpecific{SS_{fn}}{1}(\ten{\sigma}) &= \sca{\sigma}_{\mathrm{nf}}, \\
    \observationMapSpecific{SS_{ns}}{1}(\ten{\sigma}) &= \sca{\sigma}_{\mathrm{sn}}, & \observationMapSpecific{SS_{sn}}{1}(\ten{\sigma}) &= \sca{\sigma}_{\mathrm{ns}}.
    \end{aligned}
\end{equation}
In the \ac{BT} tests, the principal stresses in the fiber and in the normal direction are measured, so that $\numObservedPTestSpecific{BT}=2$. The observation maps for all five \ac{BT} tests are identical and yield
\begin{equation}\label{eq:appendix_observation_operators_anisotropic_bt}
    \begin{aligned}
        \observationMapSpecific{BT}{1}(\ten{\sigma}) &= \sca{\sigma}_{\mathrm{ff}}, &
        \observationMapSpecific{BT}{2}(\ten{\sigma}) &= \sca{\sigma}_{\mathrm{nn}}.
    \end{aligned}
\end{equation}

\subsection{Deformation filters definitions}\label{subsec:appendix_deformation_filters}

In order to reduce the number of \ac{GP} inputs and thus hyperparameters, we introduce the reduced deformation vector $\reducedFGeneral = \deformationFilterGeneral(\ten{F}) \elm{\numReducedFObservedP}$. The reduced deformation vector $\reducedFGeneral$ contains only the deformation gradient components that are relevant to predict the stress component $\setObservedPTesti_{\indObservedP}$ for the components $\indObservedP \in \{1, \dots, \numObservedPTesti\}$ observed in the respective mechanical test $\indTests \in \{1, \dots, \numTests\}$. In \cref{subsubsec:appendix_deformation_filters_treloar} and \cref{subsubsec:appendix_deformation_filters_anisotropic}, we define the deformation filters $\deformationFilterGeneral$ used for the Treloar and the anisotropic datasets, respectively.

\subsubsection{Treloar dataset}\label{subsubsec:appendix_deformation_filters_treloar}

In the case of the Treloar dataset, in all three mechanical tests, only the stress component $\sca{P}_{11}$ is measured. Accordingly, only one deformation filter is required, which is defined as follows
\begin{equation}\label{eq:appendix_deformation_filters_treloar}
    \deformationFilterSpecific{\sca{P}_{11}}(\ten{F}) = [\sca{F}_{11}, \sca{F}_{22}]\transpose.
\end{equation}
Note that for incompressible materials, the deformation gradient component $\sca{F}_{33}$ can be uniquely derived from the other two and is therefore not independent.

\subsubsection{Anisotropic dataset}\label{subsubsec:appendix_deformation_filters_anisotropic}

The anisotropic dataset includes six \acf{SS} and five \acf{BT} tests and we need eight deformation filters which are defined as follows
\begin{equation}\label{eq:appendix_deformation_filters_anisotropic}
    \begin{aligned}
        \deformationFilterSpecific{\sca{\sigma}_{\mathrm{sf}}}(\ten{F}) &= \sca{F}_{\mathrm{fs}}, & 
        \deformationFilterSpecific{\sca{\sigma}_{\mathrm{fs}}}(\ten{F}) &= \sca{F}_{\mathrm{sf}}, \\
        \deformationFilterSpecific{\sca{\sigma}_{\mathrm{nf}}}(\ten{F}) &= \sca{F}_{\mathrm{fn}}, & 
        \deformationFilterSpecific{\sca{\sigma}_{\mathrm{fn}}}(\ten{F}) &= \sca{F}_{\mathrm{nf}}, \\
        \deformationFilterSpecific{\sca{\sigma}_{\mathrm{ns}}}(\ten{F}) &= \sca{F}_{\mathrm{sn}}, & 
        \deformationFilterSpecific{\sca{\sigma}_{\mathrm{sn}}}(\ten{F}) &= \sca{F}_{\mathrm{ns}}, \\
        \deformationFilterSpecific{\sca{\sigma}_{\mathrm{ff}}}(\ten{F}) &= [\sca{F}_{\mathrm{ff}}, \sca{F}_{\mathrm{nn}}]\transpose, & 
        \deformationFilterSpecific{\sca{\sigma}_{\mathrm{nn}}}(\ten{F}) &= [\sca{F}_{\mathrm{ff}}, \sca{F}_{\mathrm{nn}}]\transpose.
    \end{aligned}
\end{equation}
Here, the indices refer to the mutually orthogonal directions of the anisotropic human cardiac tissue which are the fiber ($\mathrm{f}$), shear ($\mathrm{s}$) and normal ($\mathrm{n}$) directions.


\section{Centered 95\%-intervals and estimated coverage}\label{sec:appendix_ci_and_estimated_coverage}

In this paper, the uncertainty in the discretized stress-deformation functions $\func^{(\indTests,\indObservedP)}$ is quantified through point-wise \SI{95}{\percent}-intervals $\uqIntervalGeneral = [ \uqIntervalLowGeneral, \uqIntervalUpGeneral]$. These intervals $\uqIntervalGeneral$ are centered around the mean and contain \SI{95}{\percent} of the probability mass, which is formally defined as $P(\funcValueGeneral \in \uqIntervalGeneral) = 0.95$. Here, $\funcValueGeneral$ is the component $\indFuncPoints$ of the vector $\func^{(\indTests,\indObservedP)} \elm{\numFuncPointsTesti}$.

For the \ac{GP} posterior, the intervals $\uqIntervalGeneral$ are derived based on the posterior distribution and correspond to credible intervals \cite{rasmussen_gaussianProcesses_2005, gelman_bayesianDataAnalysis_2013}. For the statistical model, we determine the intervals $\uqIntervalGeneral$ from a finite set of random samples drawn from the statistical model, since the discretized stress-deformation functions are not necessarily normally distributed. In this case, the lower and upper bounds $\uqIntervalLowGeneral$ and $\uqIntervalUpGeneral$ of the interval are set to the values for which $\SI{2.5}{\percent}$ of the samples lie below or above them, respectively.

As a measure of the validity of the quantified uncertainty in the discretized stress-deformation function $\func^{(\indTests,\indObservedP)}$, we estimate the coverage in our numerical tests based on the measurement data $\dataSet$ as follows
\begin{equation}\label{eq:appendix_estimated_coverage}
    \estimatedCoverage^{(\indTests,\indObservedP)} =
    \frac{1}{100}
    \frac{1}{\numDataTesti}
    \sum_{\indData=1}^{\numDataTesti}
    1_{\uqInterval}(\funcValueGeneralData)
\end{equation}
where $1_{\uqInterval}$ is an indicator function defined as 
\begin{equation}\label{eq:appendix_coverage_indicator_function}
    1_{\uqInterval}(\funcValueGeneralData) =
    \begin{cases}
        1, & \text{if } \funcValueGeneralData \in \uqInterval \\
        0, & \text{otherwise}
    \end{cases}.
\end{equation}
Accordingly, the total estimated coverage for all tests is calculated as follows
\begin{equation}\label{eq:appendix_total_estimated_coverage}
    \estimatedCoverage = 
    \frac{1}{100}
    \frac{1}{\numTests} \sum_{\indTests=1}^{\numTests}
    \frac{1}{\numObservedPTesti\numDataTesti} \sum_{\indObservedP=1}^{\numObservedPTesti}
    \sum_{\indData=1}^{\numDataTesti}
    1_{\uqInterval}(\funcValueGeneralData).
\end{equation}
\noindent \textbf{Remark:} The term coverage is well-defined in frequentist statistics and is calculated using the true stress value. From a frequentist point of view, the uncertainty is correctly quantified if for $\numDataTesti \to \infty$ the measures $\estimatedCoverage^{(\indTests,\indObservedP)}$ and $\estimatedCoverage$ converge to $\SI{95}{\percent}$. However, since the true stress values are generally unknown for experimental data, we use the measured stress values from the dataset for validation. Consequently, the measures $\estimatedCoverage^{(\indTests,\indObservedP)}$ and $\estimatedCoverage$ are only estimates for the true coverage. In contrast, in the synthetic numerical test, we use the known true stress values to determine the coverage, which are also calculated according to \cref{eq:appendix_estimated_coverage,eq:appendix_coverage_indicator_function,eq:appendix_total_estimated_coverage} but referred to as $\coverage^{(\indTests,\indObservedP)}$ and $\coverage$, respectively. Note that even in the synthetic test case, the coverage is only an estimate, since the amount of available data is generally limited in our numerical test cases.


\section{Hyperparameters and scalability}\label{sec:appendix_hyperparameters}

In \cref{tab:appendix_hyperparameters}, we summarize the hyperparameters of the \ac{GP} prior, \ac{NF}, Wasserstein-1 minimization and the Sobol' sensitivity analysis involved in the proposed framework for all test cases considered in \cref{sec:results}. The selection of the hyperparameters is based on the following considerations:

\begin{itemize}
    \item \textbf{\ac{GP} prior:} We choose a scaled squared-exponential kernel because the sample paths and mean functions of \acp{GP} with squared-exponential kernels are smooth \cite{rasmussen_gaussianProcesses_2005}. The results show that the AdamW optimizer with the specified learning rate and number of iterations leads to fast, yet stable, convergence in all numerical tests. As explained in \cref{subsubsec:gps}, the specified scaling factor for the kernel length scales leads to a reasonable trade-off between physical consistency and our prior knowledge about the variety and uncertainty of the stress-deformation functions.
    
    \item \textbf{\ac{NF}:} The standard multivariate normal distribution is a common choice for the base distribution of \acp{NF} \cite{papamakarios_normalizingFlows_2021}, which we also adopt. We use \acp{IAF} since this autoregressive flow is fast to evaluate and scales well to high-dimensional distributions \cite{kingma_inverseAutoregressiveFlow_2016}, see \cref{subsubsec:normalizing_flows}. The \ac{NF} architecture which includes the number of sub-transformations and the width and depth of the \ac{MADE} networks was selected based on preliminary tests. Since the width of the \ac{MADE} network depends on the number of material parameters $\numMaterialParams$, the \ac{NF} architecture, and thus its capacity, automatically adapts to the size of the model library. Preliminary tests also showed that larger \ac{NF} architectures with more trainable parameters lead to an increase in training runtime but do not result in a significant improvement in the predictive capabilities of the statistical model.

    \item \textbf{Wasserstein-1 minimization:} The number of function samples for the estimation of the expectation operators in \cref{eq:w1_distance_optimization_lipschitz} and \cref{eq:w1_distance_optimization_wasserstein}, the number of discretization points $\numFuncPointsTesti$ per stress-deformation function, and the number of iterations $\numItersWasserstein$ were selected systematically by conducting preliminary tests. Further increasing the value of these hyperparameters did not lead to a significant improvement in accuracy. For the remaining hyperparameters, we started with the values used in \cite{wang_normalizingFlowAdaptiveSurrogate_2022,gulrajani_improvedWassersteinGANs_2017} and adjusted these values to our use case based on the results of preliminary tests. The selected optimizers, learning rates, decay rates and number of iterations $\numItersLipschitz$ balance the convergence speed and stability of the iterative optimization procedure. For the Lipschitz-1 penalty coefficient, we found that $\lipschitzPenaltyCoefficient = \num{10}$, which is also used as the default value in \cite{gulrajani_improvedWassersteinGANs_2017}, leads to robust and good results.

    \item \textbf{Sobol' sensitivity analysis:} We use the total-order Sobol' index since it measures the total effect of the corresponding parameter on the variance of the model output and also takes the interactions with the other parameters into account. The total-order Sobol' indices are calculated using the Saltelli sampling method \cite{saltelli_sensitivityIndices_2002,saltelli_sensitivityAnalysis_2010} which is a fairly common strategy. The number of samples used to determine the total-order Sobol' indices was chosen to be so large that no significant variation in the indices was observed when the number was increased further. We estimate the minimum and maximum sample bounds from the distilled distribution $\materialParamsDistNF$ to further automate the sensitivity analysis. Furthermore, the threshold for the total-order Sobol' index below which the parameters are considered non-relevant is set to $\num{1e-4}$. As explained in \cref{subsubsec:sensitivity_analysis}, there is no universally suitable threshold. Instead, the threshold must be selected depending on the use case. In \cref{fig:sensitivities_isotropic}, \cref{fig:sensitivities_anisotropic_synthetic} and \cref{fig:sensitivities_anisotropic_experimental}, the model selection process is illustrated for the isotropic, the synthetic anisotropic and the experimental anisotropic test case, respectively. These plots show that for all three test cases, the selected threshold for the total-order Sobol' index achieves a reasonable trade-off between the sparsity of the discovered model and its ability to quantify the uncertainty.
\end{itemize}
We do not claim to have found the optimal hyperparameters. However, we achieve good results with the selected hyperparameters in all three numerical test cases. In addition, minimization of the Wasserstein-1 distance is robust with the selected hyperparameters, as we show in \cref{sec:appendix_random_initialization} for five different random initializations of the involved \acp{NN} for each numerical test case. Furthermore, we note that the research code is freely available in \cite{anton_codeUQInModelDiscovery_2025} and therefore can be used by the reader for further analysis with different hyperparameters, stress-deformation data, or model libraries.\\


Finally, we discuss the scalability of the framework presented in this paper in qualitative terms and identify the hyperparameters in \cref{tab:appendix_hyperparameters} that are expected to have the most significant impact on computational costs. The computational costs are mainly influenced by the design of experiments in which the stress-deformation data are collected, the resolution of the discretization, as well as the size of the model library. The larger the number of mechanical tests $\numTests$ and discretization points $\numFuncPointsTesti$ are, the larger the size $\numFuncPoints$ of the vector with all discretized stress-deformation functions $\func$ defined in \cref{eq:stacked_functions} is. As the size of $\func$ increases, the size of $\funcGP$ in \cref{eq:sampled_functions_gp} and $\parametersToStateMap$ in \cref{eq:sampled_functions_model} as well as the number of parameters of the \ac{NN} used as Lipschitz-1 function $\wassersteinLipschitzNetwork$ in \cref{eq:w1_distance_optimization_lipschitz} and \cref{eq:w1_distance_optimization_wasserstein} increase. The computational cost for Wasserstein-1 minimization also grows with the number of samples used to estimate the expectation operators in \cref{eq:w1_distance_optimization_lipschitz} and \cref{eq:w1_distance_optimization_wasserstein}. Furthermore, there is a direct correlation between the size of the model library, on the one hand, and memory consumption and runtime, on the other. The more material parameters $\materialParams$ the model library contains, the larger the dimensionality of their joint probability distribution, which is approximated by the \ac{NF}. As the complexity of the joint probability distribution of the material parameters increases, the number of trainable parameters of the \ac{NF} generally must also be increased. In addition, a growing model library also increases the memory consumption and runtime of the sensitivity analysis. However, we found that the Wasserstein-1 minimization is the most computationally intensive step in the proposed framework compared to the inference of the \ac{GP} posterior and the Sobol' sensitivity analysis.

\begin{table}[!htbp]
\centering
\caption{Selected hyperparameters of the \ac{GP} prior, \ac{NF}, Wasserstein-1 minimization and the Sobol' sensitivity analysis involved in the proposed framework for each test case considered in \cref{sec:results}.}
\label{tab:appendix_hyperparameters}

{\small
\begin{tabularx}{\textwidth}{
l
>{\centering\arraybackslash}X
>{\centering\arraybackslash}X
>{\centering\arraybackslash}X}
\toprule
\multirow{3}{*}{hyperparameter} & \multicolumn{3}{c}{test case} \\
                                \cmidrule{2-4}
                                & \multicolumn{1}{c}{\begin{tabular}{@{}c@{}} Treloar\\ (isotropic)\end{tabular}}   & \multicolumn{1}{c}{\begin{tabular}{@{}c@{}} synthetic\\anisotropic\end{tabular}} & \multicolumn{1}{c}{\begin{tabular}{@{}c@{}} experimental\\anisotropic\end{tabular}} \\

\midrule
\textbf{\acf{GP} prior} & & &\\
\midrule
kernel & \multicolumn{3}{c}{scaled squared-exponential \cite{rasmussen_gaussianProcesses_2005}}\\
\midrule
optimizer for $\gpParams$ & \multicolumn{3}{c}{\begin{tabular}{@{}c@{}} AdamW \cite{loshchilov_AdamW_2018}\\ learning rate: \num{0.2} \end{tabular}}\\
\midrule
iterations & \multicolumn{3}{c}{\num{10000}}\\
\midrule
scaling factor for length scales & \multicolumn{3}{c}{\num{0.6}}\\
\midrule

\textbf{\acf{NF}} & & &\\
\midrule
base distribution & \multicolumn{3}{c}{standard multivariate normal}\\
\midrule
method & \multicolumn{3}{c}{\acf{IAF} \cite{kingma_inverseAutoregressiveFlow_2016}}\\
\midrule
sub-transformations & \multicolumn{3}{c}{\num{16}}\\
\midrule
depth / width of \ac{MADE} network & \multicolumn{3}{c}{\num{1} hidden layer / $4 \numMaterialParams$ neurons}\\
\midrule

\textbf{Wasserstein-1 minimization} & & &\\
\midrule
\multicolumn{1}{l}{samples for estimating expectations} & \multicolumn{3}{c}{32}\\
\midrule
\multicolumn{1}{l}{\begin{tabular}{@{}l@{}} discretization points $\numFuncPointsTesti$ per \\ stress-deformation function\end{tabular}} & \multicolumn{3}{c}{32}\\
\midrule
iterations $\numItersLipschitz$ & \multicolumn{3}{c}{\num{10}}\\
\midrule
\multicolumn{1}{l}{\begin{tabular}{@{}l@{}}iterations $\numItersWasserstein$\\ (discovery / refinement step)\end{tabular}} & \multicolumn{3}{c}{\num{20000} / \num{10000}}\\
\midrule
optimizer for $\lipschitzParams$ & \multicolumn{3}{c}{\begin{tabular}{@{}c@{}} AdamW \cite{loshchilov_AdamW_2018}\\ learning rate: \num{1e-4}\end{tabular}}\\
\midrule
optimizer for $\nfParams$ & \multicolumn{3}{c}{\begin{tabular}{@{}c@{}} RMSprop \cite{tieleman_RMSprop_2012}\\ initial learning rate: \num{5e-4}\\ learning rate decay: \num{0.9999} \end{tabular}}\\
\midrule
Lipschitz-1 penalty coefficient $\lipschitzPenaltyCoefficient$ & \multicolumn{3}{c}{\num{10}}\\
\midrule

\textbf{Sobol' sensitivity analysis} & & &\\
\midrule
considered sensitivity index & \multicolumn{3}{c}{total-order Sobol' index}\\
\midrule
sampling method & \multicolumn{3}{c}{Saltelli sampling method \cite{saltelli_sensitivityIndices_2002,saltelli_sensitivityAnalysis_2010}}\\
\midrule
sampling bounds & \multicolumn{3}{c}{\begin{tabular}{@{}c@{}}minimums and maximums estimated from\\ \num{8192} samples drawn from $\materialParamsDistNF$\end{tabular}}\\
\midrule
threshold & \multicolumn{3}{c}{$\num{1e-4}$}\\
\midrule
samples & \multicolumn{3}{c}{$\num{4096} (\numMaterialParams + 2)$}\\

\bottomrule

\end{tabularx}
}
\end{table}


\section{Sensitivity with respect to NN initialization}\label{sec:appendix_random_initialization}

The framework presented in this manuscript uses \acp{NN} in both the \ac{NF} and as Lipschitz-1 function to minimize the Wasserstein-1 distance. Due to the non-convexity of the optimization problem to be solved, the initialization of these \acp{NN} may affect the optimization result. For this reason, we test five random initializations of the \acp{NN} for each numerical test case. We summarize validation metrics and the structure of the constitutive models discovered in \cref{tab:appendix_initialization_isotropic,tab:appendix_initialization_anisotropic_synthetic,tab:appendix_initialization_anisotropic_experimental} for the Treloar, the synthetic anisotropic and the experimental anisotropic test case, respectively. Please note that we always use the same synthetic noisy data in the synthetic anisotropic case and that data generation is not influenced by the random seed. The results presented in this appendix underline that the initialization slightly affects the structure of the discovered models. However, the discovered model structures differ mainly in a few model terms, and no significant differences in structure and validation metrics were found. In summary, the sensitivity analysis shows that the proposed framework is reasonably robust to the initialization of the \acp{NN} used within this framework.

\begin{table}[!htbp]
\centering
\caption{\textbf{Treloar test case}: Sensitivity of discovered constitutive model with respect to the random initialization of the involved \acp{NN}. This table summarizes the most important results for five different random seeds, such as the validation metrics and the model structure. The results for the random seed \num{2} are considered representative among the five initializations and are discussed in more detail in \cref{subsec:isotropic_data}.}
\label{tab:appendix_initialization_isotropic}

\begin{tabularx}{\textwidth}{
l
>{\centering\arraybackslash}X
>{\centering\arraybackslash}X
>{\centering\arraybackslash}X
>{\centering\arraybackslash}X
>{\centering\arraybackslash}X
}
\toprule

\textbf{random seed}            &   0           &   1           &   2           &   3           &   4           \\

\midrule

\textbf{\ac{GP} posterior}      &               &               &               &               &               \\
total $\estimatedCoverage [\%]$ & \num{96.23}   & \num{96.23}   & \num{96.23}   & \num{96.23}   & \num{96.23}   \\

\midrule

\textbf{discovered model}       &               &               &               &               &               \\
total $\estimatedCoverage [\%]$ & \num{86.79}   & \num{94.34}   & \num{92.45}   & \num{92.45}   & \num{94.34}   \\
total $\rSquare$                & \num{0.9964}  & \num{0.9963}  & \num{0.9964}  & \num{0.9964}  & \num{0.9958}  \\
total $\rmse$                   & \num{0.0947}  & \num{0.0955}  & \num{0.0942}  & \num{0.0942}  & \num{0.1026}  \\

\midrule

\textbf{model structure}        &               &               &               &               &               \\
$\materialParamLinear^{(1, 0)} (\invariantOne - 3)$             
                                & \checkmark    & \checkmark    & \checkmark    & \checkmark    & \checkmark    \\
$\materialParamLinear^{(3, 0)} (\invariantOne - 3)^{3}$       
                                & \checkmark    & \checkmark    & \checkmark    & \checkmark    & \checkmark    \\
$\materialParamLinear^{(0, 1)} (\invariantTwo - 3)$             
                                & \checkmark    &               & \checkmark    &               & \checkmark    \\
$\materialParamLinear^{(-1)} (\principalStretch{1}^{-1} + \principalStretch{2}^{-1} + \principalStretch{3}^{-1} - 3)$ 
                                & \checkmark    & \checkmark    & \checkmark    & \checkmark    & \checkmark    \\
$\materialParamLinear^{(1)} (\principalStretch{1}^{1} + \principalStretch{2}^{1} + \principalStretch{3}^{1} - 3)$ 
                                & \checkmark    & \checkmark    & \checkmark    & \checkmark    & \checkmark    \\

\bottomrule

\end{tabularx}
\end{table}

\begin{table}[!htbp]
\centering
\footnotesize
\caption{\textbf{Synthetic anisotropic test case}: Sensitivity of discovered constitutive model with respect to the random initialization of the involved \acp{NN}. This table summarizes the most important results for five different random seeds, such as the validation metrics and the model structure. The results for the random seed \num{0} are considered representative among the five initializations and are discussed in more detail in \cref{subsubsec:anisotropic_synthetic_data}. For comparison, the structure of the four-term model \cite{martonova_modelDiscoveryCardiacTissue_2025} used to generate the synthetic dataset is also shown.}
\label{tab:appendix_initialization_anisotropic_synthetic}

\begin{tabularx}{\textwidth}{
l
>{\centering\arraybackslash}X
>{\centering\arraybackslash}X
>{\centering\arraybackslash}X
>{\centering\arraybackslash}X
>{\centering\arraybackslash}X
c
}
\toprule

\textbf{random seed}            &   0           &   1           &   2           &   3           &   4           & \multicolumn{1}{c}{\begin{tabular}{@{}c@{}}four-term \\ model \cite{martonova_modelDiscoveryCardiacTissue_2025}\end{tabular}} \\

\midrule

\textbf{\ac{GP} posterior}      &               &               &               &               &               &               \\
total $\coverage [\%]$ & \num{89.77}   & \num{89.77}   & \num{89.77}   & \num{89.77}   & \num{89.77}   & -             \\

\midrule

\textbf{discovered model}       &               &               &               &               &               &               \\
total $\coverage [\%]$ & \num{96.59}   & \num{98.86}   & \num{97.73}   & \num{95.45}   & \num{96.02}   & -             \\
total $\rSquare$                & \num{0.9997}  & \num{0.9993}  & \num{0.9997}  & \num{0.9996}  & \num{0.9997}  & -             \\
total $\rmse$                   & \num{0.0207}  & \num{0.0345}  & \num{0.0231}  & \num{0.0239}  & \num{0.0211}  & -             \\

\midrule

\textbf{model structure}        &               &               &               &               &               &               \\
$\materialParamLinear^{(2,1)} \bigr[ \invariantOne - 3 \bigl]$
                                &               &               & \checkmark    & \checkmark    & \checkmark    &               \\
$\materialParamLinear^{(2,3)} \bigr[ \invariantOne - 3 \bigl]^{2}$
                                & \checkmark    & \checkmark    & \checkmark    & \checkmark    & \checkmark    &               \\
$\materialParamLinear^{(2,4)} \Bigl( \operatorname{exp}\bigl( \materialParamNonlinearSingle^{(1,4)} \bigr[ \invariantOne - 3 \bigl]^{2} \bigr) - 1 \Bigr)$
                                &               &               & \checkmark    &               &               &               \\
$\materialParamLinear^{(2,7)} \bigr[ \invariantTwo - 3 \bigl]^{2}$
                                & \checkmark    & \checkmark    & \checkmark    & \checkmark    & \checkmark    & \checkmark    \\
$\materialParamLinear^{(2,8)} \Bigl( \operatorname{exp}\bigl( \materialParamNonlinearSingle^{(1,8)} \bigr[ \invariantTwo - 3 \bigl]^{2} \bigr) - 1 \Bigr)$
                                & \checkmark   & \checkmark     & \checkmark    & \checkmark    & \checkmark    &               \\
$\materialParamLinear^{(2,12)} \Bigl( \operatorname{exp}\bigl( \materialParamNonlinearSingle^{(1,12)} \bigl[ \invariantFourFBar - 1 \bigr]^{2} \bigr) - 1 \Bigr)$
                                & \checkmark    & \checkmark    & \checkmark    & \checkmark    & \checkmark    & \checkmark    \\
$\materialParamLinear^{(2,19)} \bigr[ \invariantFourNBar - 1 \bigl]^{2}$
                                & \checkmark    & \checkmark    & \checkmark    & \checkmark    & \checkmark    &               \\
$\materialParamLinear^{(2,20)} \Bigl( \operatorname{exp}\bigl( \materialParamNonlinearSingle^{(1,20)} \bigl[ \invariantFourNBar - 1\bigr]^{2} \bigr) - 1 \Bigr)$
                                & \checkmark    & \checkmark    & \checkmark    & \checkmark    & \checkmark    & \checkmark    \\
$\materialParamLinear^{(2,24)} \Bigl( \operatorname{exp}\bigl( \materialParamNonlinearSingle^{(1,24)} \bigl[ \invariantEightFS \bigr]^{2} \bigr) - 1 \Bigr)$
                                & \checkmark    & \checkmark    & \checkmark    & \checkmark    & \checkmark    & \checkmark    \\
$\materialParamLinear^{(2,27)} \bigr[ \invariantEightFN \bigl]^{2}$
                                & \checkmark    &               &               &               &               \\
$\materialParamLinear^{(2,28)} \Bigl( \operatorname{exp}\bigl( \materialParamNonlinearSingle^{(1,28)} \bigl[ \invariantEightFN \bigr]^{2} \bigr) - 1 \Bigr)$
                                &               & \checkmark    & \checkmark    & \checkmark    & \checkmark    &               \\
\bottomrule

\end{tabularx}
\end{table}

\begin{table}[!htbp]
\centering
\footnotesize
\caption{\textbf{Experimental anisotropic test case}: Sensitivity of discovered constitutive model with respect to the random initialization of the involved \acp{NN}. This table summarizes the most important results for five different random seeds, such as the validation metrics and the model structure. The results for the random seed \num{0} are considered representative among the five initializations and are discussed in more detail in \cref{subsubsec:anisotropic_experimental_data}. For comparison, the structure of the four-term model previously discovered in \cite{martonova_modelDiscoveryCardiacTissue_2025} using a deterministic model discover approach is also shown.}
\label{tab:appendix_initialization_anisotropic_experimental}

\begin{tabularx}{\textwidth}{
l
>{\centering\arraybackslash}X
>{\centering\arraybackslash}X
>{\centering\arraybackslash}X
>{\centering\arraybackslash}X
>{\centering\arraybackslash}X
c
}
\toprule

\textbf{random seed}            &   0           &   1           &   2           &   3           &   4           & \multicolumn{1}{c}{\begin{tabular}{@{}c@{}}four-term \\ model \cite{martonova_modelDiscoveryCardiacTissue_2025}\end{tabular}} \\

\midrule

\textbf{\ac{GP} posterior}      &               &               &               &               &               &               \\
total $\estimatedCoverage [\%]$ & \num{93.75}   & \num{93.75}   & \num{93.75}   & \num{93.75}   & \num{93.75}   & -             \\

\midrule

\textbf{discovered model}       &  & & & &\\
total $\estimatedCoverage [\%]$ & \num{34.66}   & \num{34.66}   & \num{35.23}   & \num{35.23}   & \num{34.09}   & -             \\
total $\rSquare$                & \num{0.9352}  & \num{0.9355}  & \num{0.9358}  & \num{0.9358}  & \num{0.9355}  & \num{0.9239}  \\
total $\rmse$                   & \num{0.3441}  & \num{0.3434}  & \num{0.3426}  & \num{0.3425}  & \num{0.3434}  & \num{0.3729}  \\

\midrule

                                &               &               &               &               &               &               \\
$\materialParamLinear^{(2,5)} \bigr[ \invariantTwo - 3 \bigl]$
                                & \checkmark    & \checkmark    & \checkmark    & \checkmark    & \checkmark    &               \\
$\materialParamLinear^{(2,6)} \Bigl( \operatorname{exp}\bigl( \materialParamNonlinearSingle^{(1,6)} \bigl[ \invariantTwo - 3 \bigr] \bigr) - 1 \Bigr)$
                                & \checkmark    & \checkmark    & \checkmark    & \checkmark    & \checkmark    &               \\
$\materialParamLinear^{(2,7)} \bigr[ \invariantTwo - 3 \bigl]^{2}$
                                & \checkmark    & \checkmark    & \checkmark    & \checkmark    & \checkmark    & \checkmark    \\
$\materialParamLinear^{(2,12)} \Bigl( \operatorname{exp}\bigl( \materialParamNonlinearSingle^{(1,12)} \bigl[ \invariantFourFBar - 1 \bigr]^{2} \bigr) - 1 \Bigr)$
                                & \checkmark    & \checkmark    & \checkmark    & \checkmark    & \checkmark    & \checkmark    \\
$\materialParamLinear^{(2,16)} \Bigl( \operatorname{exp}\bigl( \materialParamNonlinearSingle^{(1,16)} \bigl[ \invariantFourSBar - 1 \bigr]^{2} \bigr) - 1 \Bigr)$
                                & \checkmark    & \checkmark    & \checkmark    & \checkmark    & \checkmark    &               \\
$\materialParamLinear^{(2,20)} \Bigl( \operatorname{exp}\bigl( \materialParamNonlinearSingle^{(1,20)} \bigl[ \invariantFourNBar - 1\bigr]^{2} \bigr) - 1 \Bigr)$
                                & \checkmark    & \checkmark    & \checkmark    & \checkmark    & \checkmark    & \checkmark    \\
$\materialParamLinear^{(2,23)} \bigr[ \invariantEightFS \bigl]^{2}$
                                & \checkmark    & \checkmark    & \checkmark    & \checkmark    & \checkmark    &               \\
$\materialParamLinear^{(2,24)} \Bigl( \operatorname{exp}\bigl( \materialParamNonlinearSingle^{(1,24)} \bigl[ \invariantEightFS \bigr]^{2} \bigr) - 1 \Bigr)$
                                &               &               &               &               &               & \checkmark    \\

\bottomrule

\end{tabularx}
\end{table}


\section{Complementary results}
This appendix provides complementary results for the synthetic and experimental anisotropic numerical test cases.

\subsection{Synthetic anisotropic test case}

\begin{figure}[!hp]
    \centering
    \includegraphics[width=\linewidth]{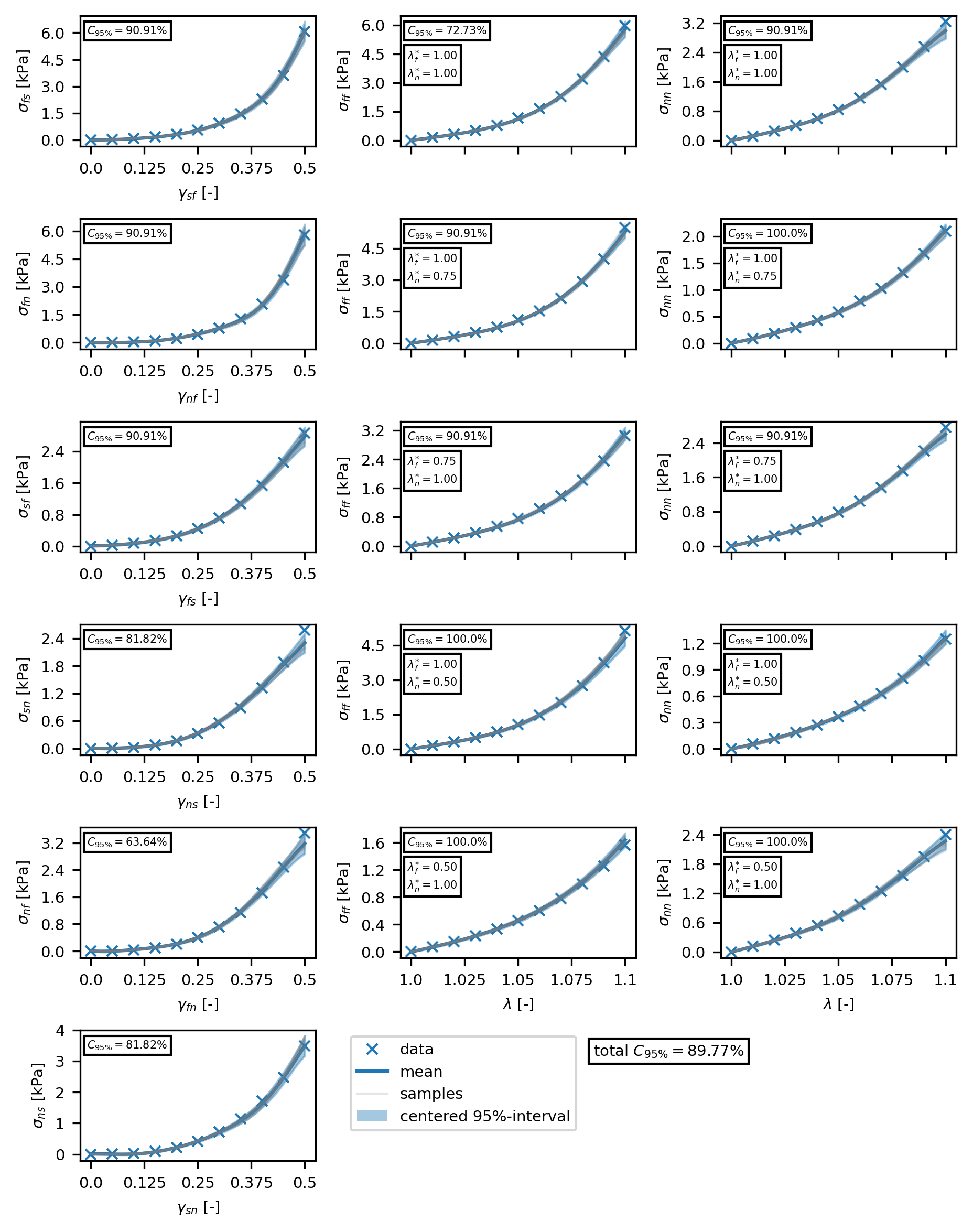}       
    \caption{\textbf{Synthetic anisotropic test case}: \Ac{GP} posterior. The illustrations show the \ac{GP} posterior mean, the centered \SI{95}{\percent}-intervals, some random stress-deformation function samples and the estimated coverages for the six \ac{SS} and five \ac{BT} tests as well as the total estimated coverage. The \ac{GP} posterior is used for data augmentation in the subsequent steps of the proposed framework.}
    \label{fig:gp_anisotropic_synthetic}
\end{figure}

In \cref{fig:gp_anisotropic_synthetic}, we show the distribution of stress-deformation functions given by the \ac{GP} posterior. For the discussion of the results for the synthetic anisotropic test case, we refer to \cref{subsubsec:anisotropic_synthetic_data}.

\subsection{Experimental anisotropic test case}

\begin{figure}[!hp]
    \centering
    \includegraphics[width=\linewidth]{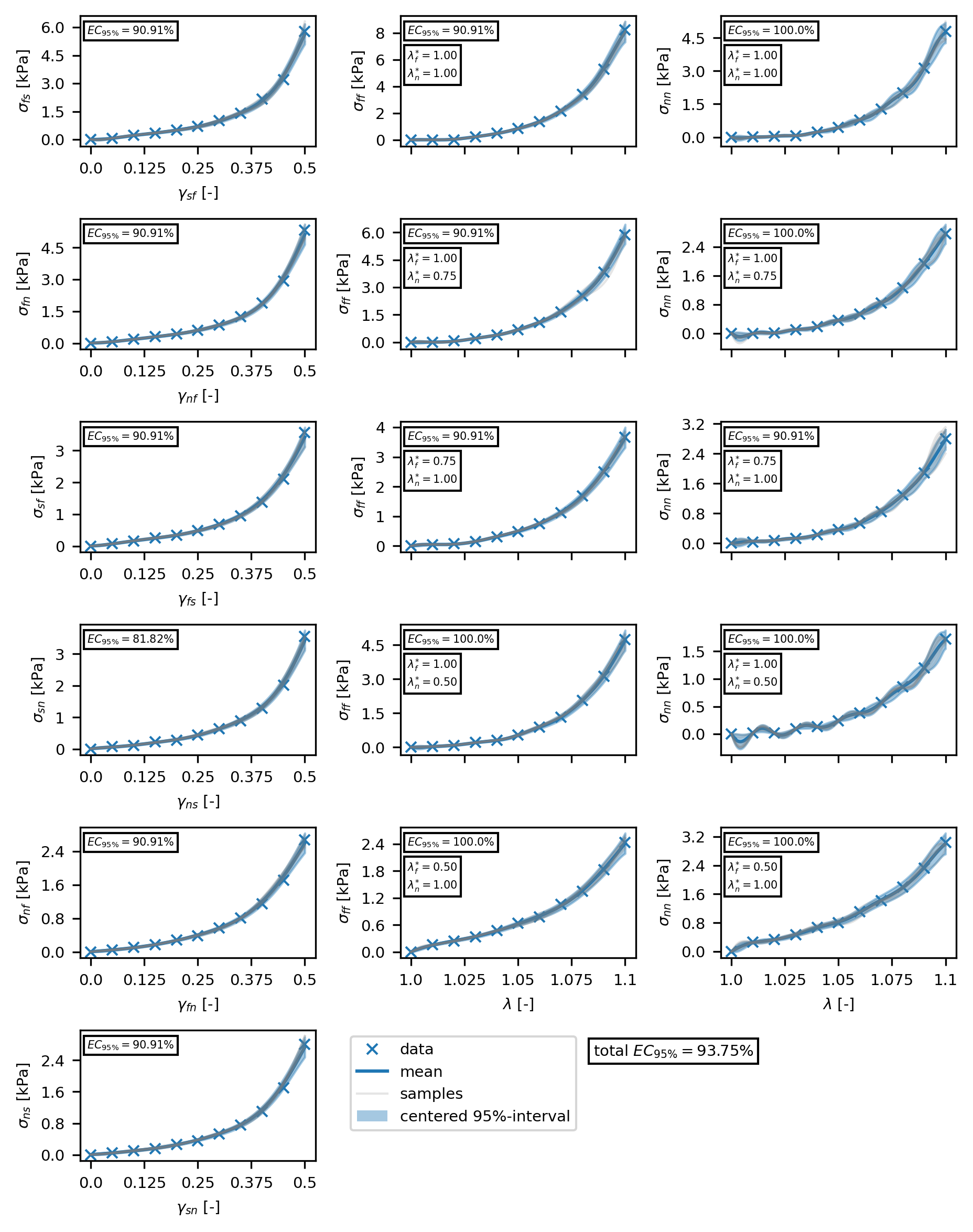}       
    \caption{\textbf{Experimental anisotropic test case}: \Ac{GP} posterior. The illustrations show the \ac{GP} posterior mean, the centered \SI{95}{\percent}-intervals, some random stress-deformation function samples and the estimated coverages for the six \ac{SS} and five \ac{BT} tests as well as the total estimated coverage. The \ac{GP} posterior is used for data augmentation in the subsequent steps of the proposed framework.}
    \label{fig:gp_anisotropic_experimental}
\end{figure}

In \cref{fig:gp_anisotropic_experimental}, we show the distribution of stress-deformation functions given by the \ac{GP} posterior. For the discussion of the results for the experimental anisotropic test case, we refer to \cref{subsubsec:anisotropic_experimental_data}.

\clearpage

\bibliographystyle{elsarticle-num} 
\bibliography{literature_R1}

@inproceedings{paszke_PyTorch_2019,
    title = {{PyTorch: An Imperative Style, High-Performance Deep Learning Library}},
    booktitle = {Advances in Neural Information Processing Systems},
    author={
        Adam Paszke and 
        Sam Gross and 
        Francisco Massa and 
        Adam Lerer and 
        James Bradbury and 
        Gregory Chanan and 
        Trevor Killeen and 
        Zeming Lin and 
        Natalia Gimelshein and 
        Luca Antiga and 
        Alban Desmaison and 
        Andreas Köpf and 
        Edward Yang and 
        Zach DeVito and 
        Martin Raison and 
        Alykhan Tejani and 
        Sasank Chilamkurthy and 
        Benoit Steiner and 
        Lu Fang and 
        Junjie Bai and 
        Soumith Chintala
    },
    editor = {
        H. Wallach and 
        H. Larochelle and 
        A. Beygelzimer and 
        F. d\textquotesingle Alch\'{e}-Buc and 
        E. Fox and 
        R. Garnett
    },
    year = {2019},
    pages = {1-12},
    volume = {32},
    location = {Vancouver, Canada},
    publisher = {Curran Associates, Inc.},
    address = {Red Hook, NY, USA},
    note = {Software available from pytorch.org},
    url = {https://proceedings.neurips.cc/paper_files/paper/2019/file/bdbca288fee7f92f2bfa9f7012727740-Paper.pdf},
}

@article{roemer_modelCalibrationInSolidMechanics_2025,
    title = {
        {Reduced and All-At-Once Approaches for Model Calibration and Discovery in Computational Solid Mechanics}
    },    
    author = {
        Römer, Ulrich and 
        Hartmann, Stefan and 
        Tröger, Jendrik-Alexander and 
        Anton, David and 
        Wessels, Henning and 
        Flaschel, Moritz and 
        De Lorenzis, Laura
    },
    year = {2025},
    journal = {Applied Mechanics Reviews},
    volume = {77},
    number = {4},
    pages = {040801},
    doi = {https://doi.org/10.1115/1.4066118},
}

@book{holzapfel_nonlinearSolidMechanics_2000,
    title = {Nonlinear Solid Mechanics: A Continuum Approach for Engineering},    
    author = {Holzapfel, G. A.},
    year = {2000},
    edition = {1},
    publisher = {Wiley},
    address = {Chichester},
    pagetotal = {480},
    isbn = {978-0-471-82319-3},
}

@article{flaschel_unsupervisedDiscoveryEUCLID_2021,
    title = {
        {Unsupervised discovery of interpretable hyperelastic constitutive laws}
    },
    author = {
        Flaschel, Moritz and 
        Kumar, Siddhant and 
        De Lorenzis, Laura
    }, 
    year = {2021}, 
    journal = {Computer Methods in Applied Mechanics and Engineering},
    volume = {381},
    pages = {113852},
    doi = {https://doi.org/10.1016/j.cma.2021.113852},
}

@book{gelman_bayesianDataAnalysis_2013,
    title = {Bayesian Data Analysis},    
    author = {
        Gelman, Andrew and
        Carlin, John B. and
        Stern, Hal S. and 
        Dunson, David B. and 
        Vehtari, Aki and 
        Rubin, Donald B. 
    },
    year = {2013},
    edition = {3},
    series = {Texts in Statistical Science},
    publisher = {Chapman and Hall/CRC},
    address = {New York},
    pagetotal = {675},
    doi = {https://doi.org/10.1201/b16018},
    isbn = {978-1-4398-9820-8},
}

@book{rasmussen_gaussianProcesses_2005,
    title = {Gaussian Processes for Machine Learning},
    author = {
        Rasmussen, Carl Edward and 
        Williams, Christopher K. I.
    },
    year = {2005},
    edition = {1},
    publisher = {MIT Press},
    address = {Cambridge},
    pagetotal={266},
    isbn = {9780262256834},
    doi = {https://doi.org/10.7551/mitpress/3206.001.0001},
}

@article{wang_normalizingFlowAdaptiveSurrogate_2022,
    title = {
        {Variational inference with NoFAS: Normalizing flow with adaptive surrogate for computationally expensive models}
    },
    author = {
        Wang, Yu and 
        Liu, Fang and 
        Schiavazzi, Daniele E.}, 
    year = {2022},
    journal = {Journal of Computational Physics},
    volume = {467},
    pages = {111454},
    doi = {https://doi.org/10.1016/j.jcp.2022.111454},
}

@article{papamakarios_normalizingFlows_2021, 
    title = {
        {Normalizing Flows for Probabilistic Modeling and Inference}
    },
    author = {
        Papamakarios, George and 
        Nalisnick, Eric and 
        Rezende, Danilo Jimenez and 
        Mohamed, Shakir and 
        Lakshminarayanan, Balaji
    }, 
    year = {2021}, 
    journal = {Journal of Machine Learning Research}, 
    volume = {22}, 
    number = {1}, 
    pages = {2617-2680}, 
    url = {https://dl.acm.org/doi/abs/10.5555/3546258.3546315}
}

@article{tran_functionalPriorForBNN_2022, 
    title = {
        {All You Need is a Good Functional Prior for Bayesian Deep Learning}
    },     
    author = {
        Tran, Ba-Hien and 
        Rossi, Simone and 
        Milios, Dimitrios and 
        Filippone, Maurizio
    }, 
    year = {2022}, 
    journal = {Journal of Machine Learning Research}, 
    volume = {23}, 
    number = {1}, 
    pages = {3210-3265},
    url = {https://dl.acm.org/doi/abs/10.5555/3586589.3586663}
}

@article{hartmann_identifiabilityMaterialParameters_2018,
    title = {
        {Identifiability of material parameters in solid mechanics}
    },
    author = {
        Hartmann, Stefan and 
        Gilbert, Rose Rogin
    },
    year = {2018},
    journal = {Archive of Applied Mechanics},
    volume = {88},
    number = {1},
    pages = {3-26},
    doi = {https://doi.org/10.1007/s00419-017-1259-4},
}

@article{tabak_nonparametricDensityEstimationAlgorithms_2013,
    title = {
        {A Family of Nonparametric Density Estimation Algorithms}
    },
    author = {
        Tabak, E. G. and 
        Turner, Cristina V.
    },
    year = {2013},
    journal = {Communications on Pure and Applied Mathematics},
    volume = {66},
    number = {2},
    pages = {145-164},
    doi = {https://doi.org/10.1002/cpa.21423},
}

@inproceedings{kingma_inverseAutoregressiveFlow_2016, 
    title = {
        {Improved variational inference with inverse autoregressive flow}
    },     
    booktitle = {Advances in Neural Information Processing Systems},     
    author = {
        Kingma, Diederik P. and 
        Salimans, Tim and 
        Jozefowicz, Rafal and 
        Chen, Xi and 
        Sutskever, Ilya and 
        Welling, Max
    }, 
    editor = {
        Lee, Daniel D. and
        Sugiyama, Masashi and
        Von Luxburg, Ulrike and
        Guyon, Isabelle
        Garnett, Roman and
    },
    year = {2016}, 
    volume = {29},
    pages = {4743–4751}, 
    location = {Barcelona, Spain}, 
    publisher = {Curran Associates Inc.}, 
    address = {Red Hook, NY, USA}, 
    url = {https://proceedings.neurips.cc/paper_files/paper/2016/file/ddeebdeefdb7e7e7a697e1c3e3d8ef54-Paper.pdf},
}

@inproceedings{rezende_normalizingFlows_2015,  
    title =  {{Variational Inference with Normalizing Flows}},  
    booktitle =  {Proceedings of the 32nd International Conference on Machine Learning},      
    author = {
        Rezende, Danilo and 
        Mohamed, Shakir
    },  
    editor = {
        Bach, Francis and 
        Blei, David
    }, 
    year = {2015},
    pages = {1530-1538},       
    volume = {37},  
    series = {Proceedings of Machine Learning Research},  
    address = {Lille, France},  
    publisher = {PMLR},  
    url = {https://proceedings.mlr.press/v37/rezende15.html},  
}

@book{villani_optimalTransport_2009,
    title = {Optimal Transport: Old and New},
    author = {Villani, Cédric},
    year = {2009},
    edition = {1},
    series = {Grundlehren der mathematischen Wissenschaften},
    publisher = {Springer},
    address = {Berlin, Heidelberg},
    pagetotal = {998},
    doi = {https://doi.org/10.1007/978-3-540-71050-9},
    isbn = {978-3-540-71050-9},
}

@book{rudin_realAndComplexAnalysis_1987,
    title = {Real And Complex Analysis},
    author = {Rudin, Walter},
    year = {1987},
    edition = {3},
    series = {Mathematics Series},
    publisher = {McGraw-Hill, Inc.},
    address = {New York},
    pagetotal = {433},
    isbn = {0-07-100276-6},
}

@book{bogachev_measureTheory_2006,
    title = {Measure Theory},
    author = {Bogachev, Vladimir I.},
    year = {2007},
    edition = {1},
    publisher = {Springer},
    address = {Berlin},
    pagetotal = {1092},
    isbn = {978-3-540-34514-5},
    doi = {https://doi.org/10.1007/978-3-540-34514-5}
}

@inproceedings{gardner_gpytorch_2018, 
    title = {
        {
        {GPyTorch: Blackbox Matrix-Matrix Gaussian Process Inference with GPU Acceleration}
    }
    }, 
    booktitle = {Advances in Neural Information Processing Systems},     
    author = {
        Gardner, Jacob R. and 
        Pleiss, Geoff and 
        Bindel, David and 
        Weinberger, Kilian Q. and 
        Wilson, Andrew Gordon
    }, 
    editor = {
        Bengio, Samy and
        Wallach, Hanna M. and
        Larochelle, Hugo and
        Grauman, Kristen and
        Cesa-Bianchi, Nicolò and
        Garnett, R.
    },
    year = {2018}, 
    volume = {31},
    pages = {7587–7597}, 
    location = {Montreal, Canada}, 
    publisher = {Curran Associates Inc.}, 
    address = {Red Hook, NY, USA},
    url = {https://proceedings.neurips.cc/paper_files/paper/2018/file/27e8e17134dd7083b050476733207ea1-Paper.pdf},
}

@article{stimper_normflows_2023, 
    title = {{normflows: A PyTorch Package for Normalizing Flows}},  
    author = {
        Stimper, Vincent and 
        Liu, David and 
        Campbell, Andrew and 
        Berenz, Vincent and 
        Ryll, Lukas and 
        Schölkopf, Bernhard and 
        Hernández-Lobato, José Miguel 
    }, 
    year = {2023},
    journal = {Journal of Open Source Software}, 
    volume = {8},
    number = {86}, 
    pages = {5361}, 
    doi = {https://doi.org/10.21105/joss.05361}, 
}

@article{linka_CANNs_2021,
    title = {
        {Constitutive artificial neural networks: A fast and general approach to predictive data-driven constitutive modeling by deep learning}
    },
    author = {
        Linka, Kevin and 
        Hillgärtner, Markus and 
        Abdolazizi, Kian P. and 
        Aydin, Roland C. and 
        Itskov, Mikhail and 
        Cyron, Christian J.
    },
    year = {2021},
    journal = {Journal of Computational Physics},
    volume = {429},
    pages = {110010},
    doi = {https://doi.org/10.1016/j.jcp.2020.110010},
}

@article{bogachev_triangularTransformationsOfMeasures_2005,
    title = {{Triangular transformations of measures}},
    author = {
        Bogachev, V. I. and 
        Kolesnikov, A. V. and 
        Medvedev, K. V.
    },
    year = {2005},
    journal = {Sbornik: Mathematics},
    volume = {196},
    number = {3},
    pages = {309–335},
    doi = {https://doi.org/10.1070/SM2005v196n03ABEH000882},
}

@inproceedings{germain_MADE_2015,  
    title =  {
        {MADE: Masked Autoencoder for Distribution Estimation}
    },  
    booktitle = {Proceedings of the 32nd International Conference on Machine Learning},    
    author =  {
        Germain, Mathieu and 
        Gregor, Karol and 
        Murray, Iain and 
        Larochelle, Hugo
    },  
    editor =  {
        Bach, Francis and 
        Blei, David
    },    
    year =  {2015},  
    pages =  {881--889}, 
    volume =  {37},  
    series =  {Proceedings of Machine Learning Research},  
    address =  {Lille, France},   
    publisher = {PMLR}, 
    url =  {https://proceedings.mlr.press/v37/germain15.html}, 
}

@inproceedings{papamakarios_maskedAutoregressiveFlow_2017,
    title = {{Masked Autoregressive Flow for Density Estimation}},    
    booktitle = {Advances in Neural Information Processing Systems},
    author = {
        Papamakarios, George and 
        Pavlakou, Theo and 
        Murray, Iain
    },
    editor = {
        Guyon, I. and 
        Von Luxburg, U. and 
        Bengio, S. and 
        Wallach, H. and 
        Fergus, R.
        Vishwanathan, S. and 
        Garnett, R.
    },
    year = {2017},
    volume = {30},
    pages = {1-10},
    location = {Long Beach, California, USA},
    publisher = {Curran Associates, Inc.},
    address = {Red Hook, NY, USA},
    url = {https://proceedings.neurips.cc/paper_files/paper/2017/file/6c1da886822c67822bcf3679d04369fa-Paper.pdf},
}

@inproceedings{dinh_realNVP_2017,
    title={{Density estimation using Real NVP}},
    booktitle={International Conference on Learning Representations},
    author={
        Dinh, Laurent and 
        Sohl-Dickstein, Jascha and 
        Bengio, Samy
    },
    year={2017},
    pages = {1-32},
    location = {Toulon, France},
    publisher = {OpenReview.net},
    url={https://openreview.net/forum?id=HkpbnH9lx}
}

@inproceedings{chen_residualFLows_2019,
    title = {{Residual Flows for Invertible Generative Modeling}},
    booktitle = {Advances in Neural Information Processing Systems},
    author = {
        Chen, Ricky T. Q. and 
        Behrmann, Jens and 
        Duvenaud, David K. and 
        Jacobsen, Joern-Henrik
    },
    editor = {
        Wallach, H. and 
        Larochelle, H. and 
        Beygelzimer, A. and 
        d'Alché-Buc, F. and 
        Fox, E. and 
        Garnett, R.
    },
    year = {2019},
    pages = {1-11},
    volume = {32},
    location = {Vancouver, Canada},
    publisher = {Curran Associates, Inc.},
    address = {Red Hook, NY, USA},
    url = {https://proceedings.neurips.cc/paper_files/paper/2019/file/5d0d5594d24f0f955548f0fc0ff83d10-Paper.pdf},
}

@inproceedings{arjovsky_wassersteinGANs_2017,
    title = {{Wasserstein Generative Adversarial Networks}},
    booktitle = {Proceedings of the 34th International Conference on Machine Learning},
    author = {
        Arjovsky, Martin and 
        Chintala, Soumith and 
        Bottou, Léon
    },
    editor = {
        Precup, Doina and 
        Teh, Yee Whye
    },
    year = {2017},
    pages = {214-223},
    series = {Proceedings of Machine Learning Research},
    volume = {70},
    publisher = {PMLR},
    address = {Sydney, Australia},
    url =  {https://proceedings.mlr.press/v70/arjovsky17a.html}
}

@inproceedings{goodfellow_GANs_2014,
    title = {{Generative Adversarial Nets}},    
    booktitle = {Advances in Neural Information Processing Systems},
    author = {
        Goodfellow, Ian J. and 
        Pouget-Abadie, Jean and 
        Mirza, Mehdi and 
        Xu, Bing and 
        Warde-Farley, David and 
        Ozair, Sherjil and 
        Courville, Aaron and 
        Bengio, Yoshua
    },
    editor = {
        Ghahramani, Z. and 
        Welling, M. and 
        Cortes, C. and 
        Lawrence, N. and 
        Weinberger K. Q.
    },
    year = {2014},
    pages = {1-9},
    volume = {27},
    location = {Montreal, Canada},
    publisher = {Curran Associates, Inc.},
    address = {Red Hook, NY, USA},
    url = {https://proceedings.neurips.cc/paper_files/paper/2014/file/f033ed80deb0234979a61f95710dbe25-Paper.pdf},   
}

@inproceedings{gulrajani_improvedWassersteinGANs_2017,
    title = {{Improved Training of Wasserstein GANs}},     
    booktitle = {Advances in Neural Information Processing Systems},
    author = {
        Gulrajani, Ishaan and 
        Ahmed, Faruk and 
        Arjovsky, Martin and 
        Dumoulin, Vincent and 
        Courville, Aaron C.
    },
    editor = {
        Guyon, I. and 
        Von Luxburg, U. and 
        Bengio, S. and 
        Wallach, H. and 
        Fergus, R. and 
        Vishwanathan, S. and 
        Garnett, R.
    },
    year = {2017},
    pages = {1-11},
    volume = {30},
    location = {Long Beach, California, USA},
    publisher = {Curran Associates, Inc.},
    address = {Red Hook, NY, USA},
    url = {https://proceedings.neurips.cc/paper_files/paper/2017/file/892c3b1c6dccd52936e27cbd0ff683d6-Paper.pdf},     
}

@misc{tieleman_RMSprop_2012,
    title = {
        {Lecture 6.5-rmsprop: Divide the Gradient by a Running Average of Its Recent Magnitude}
    },    
    author = {
        Tieleman, T. and 
        Hinton, G.
    },
    note = {{COURSERA: Neural Networks for Machine Learning, 4, 26-31}},
    year = {2012},
    howpublished = {online}
}

@inproceedings{miyato_spectralNorm_2018,
    title={{Spectral Normalization for Generative Adversarial Networks}},
    booktitle={International Conference on Learning Representations},
    author={
        Miyato, Takeru and 
        Kataoka, Toshiki and 
        Koyama, Masanori and 
        Yoshida, Yuichi
    },
    year={2018},
    pages = {1-26},
    location = {Vancouver, Canada},
    publisher = {OpenReview.net},
    url={https://openreview.net/forum?id=B1QRgziT-},
}

@inproceedings{loshchilov_AdamW_2018,
    title={{Decoupled Weight Decay Regularization}},
    booktitle={International Conference on Learning Representations},
    author={
        Loshchilov, Ilya and 
        Hutter, Frank
    },
    year={2019},
    pages = {1-8},
    location = {New Orleans, USA},
    publisher = {OpenReview.net},
    url={https://openreview.net/forum?id=Bkg6RiCqY7},
}

@article{sobol_globalSensitivityIndices_2001,
    title = {
        {Global sensitivity indices for nonlinear mathematical models and their Monte Carlo estimates}
    },
    author = {Sobol', I. M.},
    year = {2001},
    journal = {Mathematics and Computers in Simulation},
    volume = {55},
    number = {1},
    pages = {271-280},
    doi = {https://doi.org/10.1016/S0378-4754(00)00270-6},
}

@article{saltelli_sensitivityIndices_2002,
    title = {
        {Making best use of model evaluations to compute sensitivity indices}
    },
    author = {Saltelli, Andrea},
    year = {2002},
    journal = {Computer Physics Communications},
    volume = {145},
    number = {2},
    pages = {280-297},
    doi = {https://doi.org/10.1016/S0010-4655(02)00280-1},
}

@article{saltelli_sensitivityAnalysis_2010,
    title = {
        {Variance based sensitivity analysis of model output. Design and estimator for the total sensitivity index}
    },
    author = {
        Saltelli, Andrea and 
        Annoni, Paola and 
        Azzini, Ivano and 
        Campolongo, Francesca and 
        Ratto, Marco and 
        Tarantola, Stefano
    },
    year = {2010},
    journal = {Computer Physics Communications},
    volume = {181},
    number = {2},
    pages = {259-270},
    doi = {https://doi.org/10.1016/j.cpc.2009.09.018},
}

@article{herman_SALib_2017,
    title = {
        {SALib: An open-source Python library for Sensitivity Analysis}
    },
    author = {
        Herman, Jon and 
        Usher, Will 
    },
    year  = {2017},
    journal = {Journal of Open Source Software},
    volume = {2},
    number = {9},
    pages = {97},
    publisher = {The Open Journal},
    doi = {https://doi.org/10.21105/joss.00097},  
}

@article{iwanaga_SALib2_2022,
    title = {
        {Toward SALib 2.0: {Advancing} the accessibility and interpretability of global sensitivity analyses}
    },
    author = {
        Iwanaga, Takuya and 
        Usher, William and 
        Herman, Jonathan
    },
    year = {2022},
    journal = {Socio-Environmental Systems Modelling},
    volume = {4},
    pages = {18155},
    doi = {https://doi.org/10.18174/sesmo.18155},
}

@article{treloar_vulcanisedRubberData_1944,
    title={
        {Stress-strain data for vulcanised rubber under various types of deformation}
    },
    author={Treloar, L. R. G.},
    year={1944},
    journal={Transactions of the Faraday Society},
    volume={40},
    pages={59-70},
    doi = {http://dx.doi.org/10.1039/TF9444000059},
    publisher  ={The Royal Society of Chemistry},
}

@article{ricker_systematicFittingHyperelasticity_2023,
    title={
        {Systematic Fitting and Comparison of Hyperelastic Continuum Models for Elastomers}
    },
    author={
        Ricker, Alexander and 
        Wriggers, Peter
    },
    year={2023},
    journal={Archives of Computational Methods in Engineering},
    volume={30},
    number={3},
    pages={2257-2288},
    doi={https://doi.org/10.1007/s11831-022-09865-x}
}

@article{flaschel_generalizedStandardMaterialModelsEUCLID_2023,
    title={
        {Automated discovery of generalized standard material models with EUCLID}
    },
    author={
        Flaschel, Moritz and 
        Kumar, Siddhant and 
        De Lorenzis, Laura
    },
    year={2023},
    journal={Computer Methods in Applied Mechanics and Engineering},
    volume={405},
    pages={115867},
    doi={https://doi.org/10.1016/j.cma.2022.115867}
}

@article{flaschel_brainEUCLID_2023,
    title={
        {Automated discovery of interpretable hyperelastic material models for human brain tissue with EUCLID}
    },
    author={
        Flaschel, Moritz and 
        Yu, Huitian and 
        Reiter, Nina and 
        Hinrichsen, Jan and 
        Budday, Silvia and 
        Steinmann, Paul and 
        Kumar, Siddhant and 
        De Lorenzis, Laura
    },
    year={2023},
    journal={Journal of the Mechanics and Physics of Solids},
    volume={180},
    pages={105404},
    doi={https://doi.org/10.1016/j.jmps.2023.105404}
}

@article{steinmann_hyperelasticModelsTreloar_2012,
    title={
        {Hyperelastic models for rubber-like materials: consistent tangent operators and suitability for Treloar’s data}
    },
    author={
        Steinmann, Paul and 
        Hossain, Mokarram and 
        Possart, Gunnar
    },
    year={2012},
    journal={Archive of Applied Mechanics},
    volume={82},
    pages={1183-1217},
    doi={https://doi.org/10.1007/s00419-012-0610-z},
}

@article{sommer_humanVentricularMyocardium_2015,
    title = {
        {Biomechanical properties and microstructure of human ventricular myocardium}
    },
    author = {
        Sommer, Gerhard and 
        Schriefl, Andreas J. and 
        Andrä, Michaela and 
        Sacherer, Michael and 
        Viertler, Christian and 
        Wolinski, Heimo and 
        Holzapfel, Gerhard A. 
    },        
    year = {2015},
    journal = {Acta Biomaterialia},
    volume = {24},
    pages = {172-192},
    doi = {https://doi.org/10.1016/j.actbio.2015.06.031},
}

@article{martonova_modelDiscoveryCardiacTissue_2025,
    title = {
        {Automated model discovery for human cardiac tissue: Discovering the best model and parameters}
    },
    author = {
        Martonová, Denisa and 
        Peirlinck, Mathias and 
        Linka, Kevin and 
        Holzapfel, Gerhard A. and 
        Leyendecker, Sigrid and 
        Kuhl, Ellen
    },
    year = {2024},
    journal = {Computer Methods in Applied Mechanics and Engineering},
    volume = {428},
    pages = {117078},
    doi = {https://doi.org/10.1016/j.cma.2024.117078},
}

@article{ogden_OgdenModel_1972,
    title = {
        {Large deformation isotropic elasticity – on the correlation of theory and experiment for incompressible rubberlike solids}
    },    
    author = {Ogden, Raymond William },
    year = {1972},
    journal = {Proceedings of the Royal Society of London. A. Mathematical and Physical Sciences},
    volume = {326},
    number = {1567},
    pages = {565-584},
    doi = {https://doi.org/10.1098/rspa.1972.0026},
}

@article{rivlin_MooneyRivlinModel_1947,
    title = {{Torsion of a Rubber Cylinder}},    
    author = {Rivlin, R. S.},
    year = {1947},
    journal = {Journal of Applied Physics},
    volume = {18},
    number = {5},
    pages = {444-449},
    doi = {https://doi.org/10.1063/1.1697674},
}

@article{fuhg_extremeSparsificationModelDiscovery_2024,
    title = {
        {Extreme sparsification of physics-augmented neural networks for interpretable model discovery in mechanics}
    },
    author = {
        Fuhg, Jan Niklas and 
        Jones, Reese Edward and 
        Bouklas, Nikolaos 
    },
    year = {2024},
    journal = {Computer Methods in Applied Mechanics and Engineering},
    volume = {426},
    pages = {116973},
    doi = {https://doi.org/10.1016/j.cma.2024.116973},
}

@article{linka_newCANNs_2023,
    title = {
        {A new family of Constitutive Artificial Neural Networks towards automated model discovery}
    },
    author = {
        Linka, Kevin and 
        Kuhl, Ellen
    },
    year = {2023},
    journal = {Computer Methods in Applied Mechanics and Engineering},
    volume = {403},
    pages = {115731},
    doi = {https://doi.org/10.1016/j.cma.2022.115731},
}

@article{mcculloch_LPRegularizationModelDiscovery_2024,
    title = {
        {On sparse regression, Lp-regularization, and automated model discovery}
    },    
    author = {
        McCulloch, Jeremy A. and 
        St. Pierre, Skyler R. and 
        Linka, Kevin and 
        Kuhl, Ellen
    },
    year = {2024},
    journal = {International Journal for Numerical Methods in Engineering},
    volume = {125},
    number = {14},
    pages = {e7481},
    doi = {https://doi.org/10.1002/nme.7481},
}

@article{joshi_BayesianEUCLID_2022,
    title = {
        {Bayesian-EUCLID: Discovering hyperelastic material laws with uncertainties}
    },
    author = {
        Joshi, Akshay and 
        Thakolkaran, Prakash and 
        Zheng, Yiwen and 
        Escande, Maxime and 
        Flaschel, Moritz and 
        De Lorenzis, Laura and 
        Kumar, Siddhant
    },
    year = {2022},
    journal = {Computer Methods in Applied Mechanics and Engineering},
    volume = {398},
    pages = {115225},
    doi = {https://doi.org/10.1016/j.cma.2022.115225},
}

@article{linka_BayesianCANNs_2025,
    title = {
        {Discovering uncertainty: Bayesian constitutive artificial neural networks}
    },
    author = {
        Linka, Kevin and 
        Holzapfel, Gerhard A. and 
        Kuhl, Ellen
    },
    year = {2025},
    journal = {Computer Methods in Applied Mechanics and Engineering},
    volume = {433},
    pages = {117517},
    doi = {https://doi.org/10.1016/j.cma.2024.117517},
}

@article{mcculloch_GaussianCANNsCorrelatedWeights_2025,
    title = {
        {Discovering uncertainty: Gaussian constitutive neural networks with correlated weights}
    },
    author = {
        McCulloch, Jeremy A. and
        Kuhl, Ellen
    },
    year = {2025},
    journal = {Computational Mechanics},
    doi = {https://doi.org/10.1007/s00466-025-02660-y},
}

@book{ogden_nonLinearElasticDeformations_1984,
    title = {Non-Linear Elastic Deformations},   
    author = {Ogden, R. W.},
    year = {1984},
    publisher = {Ellis Horwood},
    address = {Chichester},
    pagetotal = {532},
    isbn = {9780853122739},
}

@book{gurtin_mechanicsThermodynamicsOfContinua_2010, 
    title={The Mechanics and Thermodynamics of Continua},    
    author={
        Gurtin, Morton E. and 
        Fried, Eliot and 
        Anand, Lallit
    },
    year={2010},
    publisher={Cambridge University Press}, 
    address={Cambridge},  
    pagetotal={718},
    isbn={9780521405980},
    doi={https://doi.org/10.1017/CBO9780511762956}    
}

@article{flaschel_discoveringPlasticityModels_2022,
    title = {{Discovering plasticity models without stress data}},    
    author = {
        Flaschel, Moritz and
        Kumar, Siddhant and
        De Lorenzis, Laura
    },
    year = {2022},
    journal = {npj Computational Materials},
    volume = {8},
    number = {91},
    doi = {https://doi.org/10.1038/s41524-022-00752-4},
}

@article{marino_linearViscoelasticityEUCLID_2023,
    title = {
        {Automated identification of linear viscoelastic constitutive laws with EUCLID}
    },
    author = {
        Marino, Enzo  and 
        Flaschel, Moritz and 
        Kumar, Siddhant and 
        De Lorenzis, Laura
    },
    year = {2023},
    journal = {Mechanics of Materials},
    volume = {181},
    pages = {104643},
    doi = {https://doi.org/10.1016/j.mechmat.2023.104643},
}

@article{fuhg_reviewDataDrivenConstitutiveLaws_2025,
    title = {
        {A Review on Data-Driven Constitutive Laws for Solids}
    },
    author = {
        Fuhg, Jan N. and
        Anantha Padmanabha, Govinda and 
        Bouklas, Nikolaos and 
        Bahmani, Bahador and 
        Sun, WaiChing and 
        Vlassis, Nikolaos N. and
        Flaschel, Moritz and 
        Carrara, Pietro and
        De Lorenzis, Laura 
    },
    year = {2025},
    journal = {Archives of Computational Methods in Engineering},
    volume = {32},
    pages = {1841-1883},
    doi = {https://doi.org/10.1007/s11831-024-10196-2},
}

@article{klein_polyconvexAnisotropicHyperelasticityWithNN_2022,
    title = {
        {Polyconvex anisotropic hyperelasticity with neural networks}
    },
    author = {
        Klein, Dominik K. and 
        Fernández, Mauricio and 
        Martin, Robert J. and 
        Neff, Patrizio and 
        Weeger, Oliver
    },
    year = {2022},
    journal = {Journal of the Mechanics and Physics of Solids},
    volume = {159},
    pages = {104703},
    doi = {https://doi.org/10.1016/j.jmps.2021.104703},
}

@article{linden_NNHyperelasticityEnforcingPhysics_2023,
    title = {
        {Neural networks meet hyperelasticity: A guide to enforcing physics}
    },
    author = {
        Linden, Lennart and 
        Klein, Dominik K. and 
        Kalina, Karl A. and 
        Brummund, Jörg and 
        Weeger, Oliver and 
        Kästner, Markus
    },
    year = {2023},
    journal = {Journal of the Mechanics and Physics of Solids},
    volume = {179},
    pages = {105363},
    doi = {https://doi.org/10.1016/j.jmps.2023.105363},
}

@article{fuhg_hyperelasticAnisotropyTensorBasedNN_2022,
    title = {
        {Learning hyperelastic anisotropy from data via a tensor basis neural network}
    },
    author = {
        Fuhg, J. N. and 
        Bouklas, N. and 
        Jones, R. E.
    },
    year = {2022},
    journal = {Journal of the Mechanics and Physics of Solids},
    volume = {168},
    pages = {105022},
    doi = {https://doi.org/10.1016/j.jmps.2022.105022},
}

@article{kalina_NNMeetsAnisotropicHyperelasticity_2025,
    title = {
        {Neural networks meet anisotropic hyperelasticity: A framework based on generalized structure tensors and isotropic tensor functions}
    },
    author = {
        Kalina, Karl A. and 
        Brummund, Jörg and 
        Sun, WaiChing and 
        Kästner, Markus
    },
    year = {2025},    
    journal = {Computer Methods in Applied Mechanics and Engineering},
    volume = {437},
    pages = {117725},
    doi = {https://doi.org/10.1016/j.cma.2024.117725},
}

@article{linka_brainCANNs_2023,
    title = {
        {Automated model discovery for human brain using Constitutive Artificial Neural Networks}
    },
    author = {
        Linka, Kevin  and 
        St. Pierre, Sarah R. and 
        Kuhl, Ellen 
    },    
    year = {2023},
    journal = {Acta Biomaterialia},
    volume = {160},
    pages = {134-151},
    doi = {https://doi.org/10.1016/j.actbio.2023.01.055},   
}

@article{peirlinck_universalMaterialSubroutineHyperelasticity_2024,
    title = {
        {On automated model discovery and a universal material subroutine for hyperelastic materials}
    },
    author = {
        Peirlinck, Mathias and 
        Linka, Kevin and 
        Hurtado, Juan A. and 
        Kuhl, Ellen
    },
    year = {2024},
    journal = {Computer Methods in Applied Mechanics and Engineering},
    volume = {418},
    pages = {116534},
    doi = {https://doi.org/10.1016/j.cma.2023.116534},
}

@article{linka_skinCANNs_2023,
    title = {
        {Automated model discovery for skin: Discovering the best model, data, and experiment}
    },
    author = {
        Linka, Kevin and 
        Buganza Tepole, Adrian and 
        Holzapfel, Gerhard A. and 
        Kuhl, Ellen
    },
    year = {2023},
    journal = {Computer Methods in Applied Mechanics and Engineering},
    volume = {410},
    pages = {116007},
    doi = {https://doi.org/10.1016/j.cma.2023.116007},
}

@article{wollner_BayesianFrameworkUncertaintyEstimationCalibration_2025,
    title = {
        {A reparameterization-invariant Bayesian framework for uncertainty estimation and calibration of simple materials}
    },
    author = {
        Wollner, Maximilian P. and 
        Rolf-Pissarczyk, Malte and
        Holzapfel, Gerhard A.
    },    
    year = {2025},
    journal = {Computational Mechanics},
    doi = {https://doi.org/10.1007/s00466-024-02573-2},   
}

@article{tac_physicsInformedProbabilisticDiffusionFields_2025,
    title = {
        {Generative hyperelasticity with physics-informed probabilistic diffusion fields}
    },
    author = {
        Taç, Vahidullah and
        Rausch, Manuel K. and
        Bilionis, Ilias and 
        Costabal, Francisco Sahli and 
        Tepole, Adrian Buganza
    },
    year = {2025},
    journal = {Engineering with Computers},
    volume = {41},
    pages = {51–69},
    doi = {https://doi.org/10.1007/s00366-024-01984-2},
}

@article{aggarwal_strainEnergyDensityAsGP_2023,
    title = {
        {Strain energy density as a Gaussian process and its utilization in stochastic finite element analysis: Application to planar soft tissues}
    },
    author = {
        Aggarwal, Ankush and 
        Jensen, Bjørn Sand and 
        Pant, Sanjay and 
        Lee, Chung-Hao
    },
    year = {2023},
    journal = {Computer Methods in Applied Mechanics and Engineering},
    volume = {404},
    pages = {115812},
    doi = {https://doi.org/10.1016/j.cma.2022.115812},
}

@article{thakolkaran_inputConvexKANs_2025,
    title = {
        {Can KAN CANs? Input-convex Kolmogorov-Arnold Networks (KANs) as hyperelastic constitutive artificial neural networks (CANs)}
    },
    author = {
        Thakolkaran, Prakash and 
        Guo, Yaqi and 
        Saini, Shivam and 
        Peirlinck, Mathias and 
        Alheit, Benjamin and 
        Kumar, Siddhant
    },
    year = {2025},
    journal = {Computer Methods in Applied Mechanics and Engineering},
    volume = {443},
    pages = {118089},
    doi = {https://doi.org/10.1016/j.cma.2025.118089},
}

@article{abdolazizi_CKANs_2025,
    title={
        {Constitutive Kolmogorov-Arnold Networks (CKANs): Combining accuracy and interpretability in data-driven material modeling}
    }, 
    author={
        Abdolazizi, Kian P. and 
        Aydin, Roland C. and 
        Cyron, Christian J. and
        Linka, Kevin
    },
    year={2025},
    journal={Journal of the Mechanics and Physics of Solids},
    volume = {203},
    pages = {106212},
    doi={https://doi.org/10.1016/j.jmps.2025.106212}
}

@inproceedings{ustyuzhaninov_monotonicGPFlows_2020,
    title = {{Monotonic Gaussian Process Flows}},
    booktitle = {Proceedings of the Twenty Third International Conference on Artificial Intelligence and Statistics},
    author = {
        Ustyuzhaninov, Ivan and
        Kazlauskaite, Ieva and 
        Ek, Carl Henrik and
        Campbell, Neill
    },
    editor = {
        Chiappa, Silvia and
        Calandra, Roberto
    },
    year = {2020},
    pages = {3057-3067},
    series = {Proceedings of Machine Learning Research},
    volume = {108},
    publisher = {PMLR},
    address = {online},
    url =  {https://proceedings.mlr.press/v108/ustyuzhaninov20a.html}
}

@article{tacke_constitutiveScientificGenerativeAgent_2025,
    title = {
        {Constitutive scientific generative agent (CSGA): Leveraging large language models for automated constitutive model discovery}
    },
    author = {
        Tacke, Marius and
        Busch, Matthias and
        Bali, Kartik and
        Abdolazizi, Kian and
        Linka, Kevin and
        Cyron, Christian and
        Aydin, Roland 
    },
    year = {2025},
    journal = {Machine Learning for Computational Science and Engineering},
    volume = {1},
    number = {23},
    doi = {https://doi.org/10.1007/s44379-025-00022-2},
}

@article{roemer_GPBasedCalibrationBiomechanicalModels_2022,
    title = {
        {Surrogate-based Bayesian calibration of biomechanical models with isotropic material behavior}
    },
    author = {
        Römer, Ulrich and 
        Liu, Jintian and 
        Böl, Markus
    },
    year = {2022},
    journal = {International Journal for Numerical Methods in Biomedical Engineering},
    volume = {38},
    number = {4},
    pages = {e3575},
    doi = {https://doi.org/10.1002/cnm.3575},
}

@article{tepole_polyconvexPANNs_2025,
    title = {
        {Polyconvex physics-augmented neural network constitutive models in principal stretches}
    },
    author = {
        Tepole, Adrian Buganza and 
        Jadoon, Asghar Arshad and 
        Rausch, Manuel and 
        Fuhg, Jan Niklas
    },
    year = {2025},
    journal = {International Journal of Solids and Structures},
    volume = {320},
    pages = {113469},
    doi = {https://doi.org/10.1016/j.ijsolstr.2025.113469},
}

@article{thakolkaran_NNEUCLID_2022,
    title = {
        {NN-EUCLID: Deep-learning hyperelasticity without stress data}
    },
    author = {
        Thakolkaran, Prakash and 
        Joshi, Akshay and 
        Zheng, Yiwen and 
        Flaschel, Moritz and 
        De Lorenzis, Laura and 
        Kumar, Siddhant
    },
    year = {2022},
    journal = {Journal of the Mechanics and Physics of Solids},
    volume = {169},
    pages = {105076},
    doi = {https://doi.org/10.1016/j.jmps.2022.105076}
}

@article{martonova_discoveringDispersion_2025,
    title={
        {Discovering dispersion: How robust is automated model discovery for human myocardial tissue?}
    },
    author={
        Martonová, Denisa and 
        Leyendecker, Sigrid and 
        Holzapfel, Gerhard A. and 
        Kuhl, Ellen
    },
    journal={Biomechanics and Modeling in Mechanobiology},
    year={2025},
    doi = {https://doi.org/10.1007/s10237-025-02005-x}
}

@article{narouie_statFEMPolynomialChaos_2025,
    title = {
        {Mechanical state estimation with a Polynomial-Chaos-Based Statistical Finite Element Method}
    },    
    author = {
        Narouie, Vahab and 
        Wessels, Henning and 
        Cirak, Fehmi and 
        Römer, Ulrich
    },
    year = {2025},
    journal = {Computer Methods in Applied Mechanics and Engineering},
    volume = {441},
    pages = {117970},
    doi = {https://doi.org/10.1016/j.cma.2025.117970},
}

@article{melev_noFreeBayesianUQ_2025,
    title = {
        {Position: There Is No Free Bayesian Uncertainty Quantification}
    },
    author = {
        Melev, Ivan and 
        Kauermann, Goeran
    },
    year = {2025},
    journal = {arXiv Preprint},
    doi = {https://doi.org/10.48550/arXiv.2506.03670}
}

@article{hirsh_sparsifyingPriorsModelDiscovery_2022,
    title = {
        {Sparsifying priors for Bayesian uncertainty quantification in model discovery}
    },
    author = {
        Hirsh, Seth M. and 
        Barajas-Solano, David A. and 
        Kutz, J. Nathan
    },
    year = {2022},    
    journal = {Royal Society Open Science},
    volume = {9},
    number = {2},
    pages = {211823},
    doi = {https://doi.org/10.1098/rsos.211823}
}

@article{linden_BayesianParameterEstimation_2022,
    title = {
        {Bayesian parameter estimation for dynamical models in systems biology}
    },    
    author = {
        Linden, Nathaniel J. and 
        Kramer, Boris and 
        Rangamani, Padmini
    },
    year = {2022},
    journal = {PLOS Computational Biology},
    volume = {18},
    number = {10},
    pages = {1-48},
    doi = {https://doi.org/10.1371/journal.pcbi.1010651},
}

@article{urrea_automatedModelDiscovery_2025,
    title={
        {Automated constitutive model discovery by pairing sparse regression algorithms with model selection criteria}
    }, 
    author={
        Urrea-Quintero, Jorge-Humberto and 
        Anton, David and
        De Lorenzis, Laura and
        Wessels, Henning
    },
    year={2026},
    journal={Computer Methods in Applied Mechanics and Engineering},
    volume = {449},
    pages = {118551},
    doi = {https://doi.org/10.1016/j.cma.2025.118551}
}

@article{schmidt_distillingNaturalLaws_2009,
    title = {
        {Distilling Free-Form Natural Laws from Experimental Data}
    },
    author = {
        Schmidt, Michael and 
        Lipson, Hod
    },
    year = {2009},
    journal = {Science},
    volume = {324},
    number = {5923},
    pages = {81-85},
    doi = {https://doi.org/10.1126/science.1165893},
}

@misc{anton_codeUQInModelDiscovery_2025,
    title = {
        {Code for the publication: Uncertainty quantification in model discovery by distilling interpretable material constitutive models from Gaussian process posteriors}
    },    
    author = {Anton, David},
    year = {2025},
    publisher = {Zenodo},
    doi = {https://doi.org/10.5281/zenodo.17442182},
    note = {Code available from \url{https://github.com/david-anton/UQInModelDiscovery}},
}

@article{lemoine_beyondNoninformativePriors_2019,
    title = {
        {Moving beyond noninformative priors: why and how to choose weakly informative priors in Bayesian analyses}
    },
    author = {Lemoine, Nathan P.},
    year = {2019},
    journal = {Oikos},
    volume = {128},
    number = {7},
    pages = {912-928},
    doi = {https://doi.org/10.1111/oik.05985},
}

@book{treloar_physicsRubberElasticity_2005,
    title = {{The Physics of Rubber Elasticity}},    
    author = {Treloar, L. R. G.},
    year = {2005},
    edition = {3},
    series = {Oxford Classic Texts in the Physical Sciences},
    publisher = {Oxford University Press},
    address = {Oxford},
    doi = {https://doi.org/10.1093/oso/9780198570271.001.0001},
    isbn = {978-0-19-857027-1},
}

@article{arruda_constitutiveModelRubber_1993,
    title = {
        {A three-dimensional constitutive model for the large stretch behavior of rubber elastic materials}
    },    
    author = {
        Arruda, E. M. and 
        Boyce, M. C.
    },
    year = {1993},
    journal = {Journal of the Mechanics and Physics of Solids},
    volume = {41},
    number = {2},
    pages = {389-412},
    doi = {https://doi.org/10.1016/0022-5096(93)90013-6}
}

@article{tosaka_vulcanizedNaturalRubber_2010,
    title = {
        {Molecular orientation and stress relaxation during strain-induced crystallization of vulcanized natural rubber}
    },    
    author = {
        Tosaka, Masatoshi and
        Kohjiya, Shinzo and
        Ikeda, Yuko and
        Toki, Shigeyuki and
        Hsiao, Benjamin S.
    },
    year = {2010},
    journal = {Polymer Journal},
    volume = {42},
    number = {6},
    pages = {474-481},
    doi = {https://doi.org/10.1038/pj.2010.22}
}

@article{laita_myocardiumModelingMultimodalDeformations_2025,
    title = {
        {On the myocardium modeling under multimodal deformations: a comparison between costa's, Holzapfel and Ogden's formulations}
    },    
    author = {
        Laita, N. and
        Martínez, M. A. and
        Doblaré, M. and
        Peña, E.
    },
    year = {2025},
    journal = {Meccanica},
    volume = {60},
    number = {8},
    pages = {2291–2324},
    doi = {https://doi.org/10.1007/s11012-025-01959-7}
}

@article{holzapfel_constitutiveModellingMyocardium_2009,
    title = {
        {Constitutive modelling of passive myocardium: a structurally based framework for material characterization}
    },    
    author = {
        Holzapfel, Gerhard A. and
        Ogden, Ray W.
    },
    year = {2009},
    journal = {Philosophical Transactions of the Royal Society A: Mathematical, Physical and Engineering Sciences},
    volume = {367},
    number = {1902},
    pages = {3445-3475},
    doi = {https://doi.org/10.1098/rsta.2009.0091}
}

@article{owen_modellingCardiovascularSystem_2018,
    title = {
        {Structural modelling of the cardiovascular system}
    },    
    author = {
        Owen, Benjamin and
        Bojdo, Nicholas and
        Jivkov, Andrey and
        Keavney, Bernard and
        Revell, Alistair
    },
    year = {2018},
    journal = {Biomechanics and Modeling in Mechanobiology},
    volume = {17},
    number = {5},
    pages = {1217-1242},
    doi = {https://doi.org/10.1007/s10237-018-1024-9}
}

@article{dokos_shearPropertiesMyocardium_2002,
    title = {
        {Shear properties of passive ventricular myocardium}
    },    
    author = {
        Dokos, Socrates and 
        Smaill, Bruce H. and 
        Young, Alistair A. and 
        LeGrice, Ian J.
    },
    year = {2002},
    journal = {American Journal of Physiology-Heart and Circulatory Physiology},
    volume = {283},
    number = {6},
    pages = {H2650-H2659},
    doi = {https://doi.org/10.1152/ajpheart.00111.2002}
}

@article{vaverka_modificationModelMyocardium_2025,
    title = {
        {A modification of Holzapfel–Ogden hyperelastic model of myocardium better describing its passive mechanical behavior}
    },    
    author = {
        Vaverka, J. and 
        Burša, J.
    },
    year = {2025},
    journal = {European Journal of Mechanics - A/Solids},
    volume = {111},
    pages = {105586},
    doi = {https://doi.org/10.1016/j.euromechsol.2025.105586}
}

@article{brown_cardiacMechanicsModeling_2025,
    title={
        {Cardiac mechanics modeling: recent developments and current challenges}
    },
    author={
        Brown, Aaron L. and 
        Liu, Ju and
        Ennis, Daniel B. and 
        Marsden, Alison L.
    },
    year={2025},
    journal={arXiv Preprint},
    doi = {https://doi.org/10.48550/arXiv.2509.07971}
}

\end{document}